\documentclass[twoside]{article}

\usepackage[utf8]{inputenc} 
\usepackage[T1]{fontenc}    
\usepackage{hyperref}       
\usepackage{url}            
\usepackage{booktabs}       
\usepackage[makeroom]{cancel}

\usepackage{amsmath}
\usepackage{amsfonts}	
\usepackage{amssymb} 
\usepackage{mathtools}	
\usepackage{amsthm}	
\usepackage{nicefrac}       
\usepackage{microtype}      
\usepackage{dsfont} 
\usepackage{mathrsfs}
\usepackage{xparse}

\usepackage{graphicx}
\usepackage{caption}
\usepackage{subcaption}

\usepackage[svgnames,table]{xcolor}
\usepackage[tableposition=above]{caption}
\usepackage{pifont}

\ExplSyntaxOn
\NewDocumentCommand{\rvect}{m}
 {
  \seq_set_split:Nnn \l_tmpa_seq { , } { #1 }
  \begin{bmatrix}
  \seq_use:Nn \l_tmpa_seq { & }
  \end{bmatrix}
 }
\ExplSyntaxOff

\newcommand{\ie}{i.e., }

\newcommand{\eg}{e.g., }

\newcommand{\RR}{I\!\!R}

\newcommand{\p}[1]{p{\left( #1 \right)}}

\newcommand{\q}[1]{q{\left( #1 \right)}}

\newcommand{\cp}[2]{p{\left( #1 \mid #2 \right)}}
\newcommand{\cq}[2]{q{\left( #1 \mid #2 \right)}}

\newcommand{\eValue}[2]{\mathbb{E}_{#1}\left\{ #2 \right\}}
\newcommand{\Ex}[2]{\mathbb{E}_{#1}\left\{ #2 \right\}}

\newcommand{\kl}[2]{\mathrm{KL}\left[#1 \| #2\right]}
\newcommand{\N}[1]{\mathcal{N}\left( #1\right)}

\newcommand{\tr}[1]{\mathrm{tr}\left\{ #1 \right\}} 
\newcommand{\diag}[1]{\mathrm{diag}\left\{ #1 \right\}}

\newcommand{\bigO}[1]{\mathcal{O}\left( #1 \right)}

\newcommand{\gp}{\mathcal{GP}}
\newcommand{\GP}[1]{\mathcal{GP}\left( #1\right)}
\newcommand{\fb}{\mathbf{f}}

\newcommand{\kb}{\mathbf{k}}
\newcommand{\Kb}{\mathbf{K}}

\newcommand{\Sigmab}{\mathbf{\Sigma}}
\newcommand{\mb}{\mathbf{m}}
\newcommand{\Qb}{\mathbf{Q}}

\newcommand{\mbtilde}[1]{\widetilde{\mathbf{m}}_{#1}}
\newcommand{\Kbtilde}[2]{\widetilde{\mathbf{K}}_{#1,#2}}

\newcommand{\mbstar}[1]{\mathbf{m}_{#1}^*}
\newcommand{\Kbstar}[2]{\mathbf{K}_{#1,#2}^*}

\newcommand{\XbZ}{\mathbf{X}_{M}}
\newcommand{\fbZ}{\mathbf{f}_{M}}
\newcommand{\ybC}{\mathbf{y}_{M}}
\newcommand{\XbC}{\mathbf{X}_{M}}
\newcommand{\fbC}{\mathbf{f}_{M}}
\newcommand{\betabC}{\ensuremath{\boldsymbol{\beta}_{M}}}
\newcommand{\SigmabetaC}{\Sigmab_{\betabC}}

\newcommand{\SigmabCC}{\Sigmab_{\fbC,\fbC}}
\newcommand{\KbXX}{\mathbf{K}_{NN}}
\newcommand{\KbZZ}{\mathbf{K}_{MM}}
\newcommand{\KbCC}{\mathbf{K}_{MM}}
\newcommand{\KbXZ}{\mathbf{K}_{NM}}
\newcommand{\KbZX}{\mathbf{K}_{MN}}
\newcommand{\KbXC}{\mathbf{K}_{NM}}
\newcommand{\KbCX}{\mathbf{K}_{MN}}
\newcommand{\KbSX}{\mathbf{k}_{*N}}
\newcommand{\KbXS}{\mathbf{k}_{N*}}

\newcommand{\Ab}{\mathbf{A}}

\newcommand{\Loss}{\mathcal{L}}

\newcommand{\xb}{\mathbf{x}}
\newcommand{\xbstar}{\mathbf{x}^\star}
\newcommand{\Xb}{\mathbf{X}}

\newcommand{\Zb}{\mathbf{Z}}

\newcommand{\yb}{\mathbf{y}}

\newcommand{\wb}{\mathbf{w}}

\newcommand{\Sb}{\mathbf{S}}
\newcommand{\zerob}{\mathbf{0}}

\newcommand{\Ib}{\mathbf{I}}
\newcommand{\ub}{\mathbf{u}}

\newcommand{\hilbert}{\mathcal{H}}
\DeclareMathOperator{\xspace}{\mathcal{X}}
\DeclareMathOperator{\yspace}{\mathcal{Y}}

\newcommand{\argmin}{\mathop{\mathrm{argmin}}}

\newcommand{\map}[1]{\Phi\left( #1 \right)}
\newcommand{\Dist}[1]{{\rm Dist}\left( #1\right)}
\newcommand{\expp}[1]{\exp\left\{ #1 \right\}}

\newcommand*\diff{\mathop{}\!\mathrm{d}}

\newcommand{\red}[1]{\textcolor{red}{#1}}
\newcommand{\green}[1]{\textcolor{ForestGreen}{#1}}
\newcommand{\blue}[1]{\textcolor{blue}{#1}}
\definecolor{darkgreen}{rgb}{0,0.5,0} 

\usepackage{comment}

\usepackage[round]{natbib}

\usepackage[accepted]{aistats2024}
\begin{document}

%

%

\twocolumn[

\aistatstitle{Accurate and Scalable Stochastic Gaussian Process Regression \\
via Learnable Coreset-based Variational Inference
}

\aistatsauthor{
\href{mailto:<mk4139j@columbia.edu>?Subject=About CVGP}{Mert Ketenci}
\And \href{mailto:<ajp2120@cumc.columbia.edu>?Subject=About CVGP}{Adler Perotte}
\And \href{mailto:<ne60@cumc.columbia.edu>?Subject=About CVGP}{Noémie Elhadad}
\And \href{mailto:<iurteaga@bcamath.org>?Subject=CVGP}{I\~nigo Urteaga}
}

\aistatsaddress{
Columbia University$^\dag$ \\New York, USA
\And Columbia University$^\ddag$ \\New York, USA
\And Columbia University$^{\dag\ddag}$\\New York, USA \\
\And BCAM \& Ikerbasque$^\star$\\Bilbao, Spain} ]

\begin{abstract}
We introduce a novel stochastic variational inference method for Gaussian process ($\gp$) regression, by deriving a posterior over a learnable set of coresets:
\ie over pseudo-input/output, weighted pairs.
Unlike former free-form variational families for stochastic inference,
our coreset-based variational $\gp$ (CVGP\footnote{Code is available at \url{github.com/ketencimert/cvgp}.}) is defined in terms of the $\gp$ prior and the (weighted) data likelihood.
This formulation naturally incorporates inductive biases of the prior,
and ensures its kernel and likelihood dependencies are shared with the posterior.
We derive a variational lower-bound on the log-marginal likelihood by marginalizing over the latent $\gp$ coreset variables,
and show that CVGP's lower-bound is amenable to stochastic optimization.
CVGP reduces the dimensionality of the variational parameter search space to linear $\bigO{M}$ complexity, while ensuring numerical stability at $\bigO{M^3}$ time complexity and $\bigO{M^2}$ space complexity.
Evaluations on real-world and simulated regression problems demonstrate that CVGP achieves superior inference and predictive performance
than state-of-the-art, stochastic sparse $\gp$ approximation methods.
\end{abstract}

\section{INTRODUCTION}
\label{sec:introduction}
Training $\gp$s efficiently with large datasets has been a long-standing challenge,
as exact inference complexities grow $O(N^3)$ in time and $O(N^2)$ in space requirements.

Successful state-of-the-art (SOTA) methods to scale $\gp$s
---a detailed review can be found in~\citep{liu2020gaussian}---
are based on sparse and low-rank approximations~\citep{williams2000using, snelson2005sparse, quinonero2005unifying},
often using inducing random variables ~\citep{naish2007generalized, titsias2009variational, hensman2013gaussian,wilson2015kernel}.

Amongst these techniques,
variational learning of inducing variables by~\citet{titsias2009variational} allows for time and space complexities of $O(NM^2)$ and $O(NM)$,
with clear benefits when inducing point size $M$ is small, \ie $M \leq N$.
However, in real-world applications, $N$ can be in the order of millions, making model learning impractical.
More recently, \citet{hensman2013gaussian} introduced stochastic variational inference for Gaussian processes (SVGP), which reduces the time and space complexities to \(\mathcal{O}(M^3)\) and \(\mathcal{O}(M^2)\), respectively. This method has become the standard for training \(\mathcal{GP}\) models on large datasets. However, SVGP’s scalability comes at a cost: it requires learning additional \(\mathcal{O}(M^2)\) parameters, resulting in an optimization problem that scales quadratically with the number of inducing points.

In this work,
we propose a coreset-based variational $\gp$ (CVGP) technique
that is amenable to stochastic optimization (\ie scalable to big datasets)
at reduced $\bigO{M}$ parameter complexity (see Table~\ref{table:bigo}),
with accurate inference and predictive performance in a wide range of real-datasets (see results in Section~\ref{sec:experiments_iurteaga}).
We take inspiration from the optimal variational posterior of~\citet{titsias2009variational},
and ensure that CVGP's variational family also
(1) obeys $\gp$s' prior-conditional structure,
and (2) maintains the $\gp$ prior's dependencies in its posterior,
via Bayesian coreset principles~\citep{huggins2016coresets, zhang2021bayesian}.

We design and learn a variational distribution over a probabilistic model,
defined through a subset of learnable pseudo-data points and a weighted likelihood function,
in line with existing Black-Box Bayesian coreset frameworks \citep{manousakas2020bayesian, manousakas2022black}.

CVGP's coreset-based variational $\gp$ posterior,
learnable via stochastic maximization of a tighter lower-bound of the data log-marginal likelihood, 
enables not only a more accurate approximation to the $\gp$ posterior of interest,
but a more efficient optimization process.

In summary, our contribution is a novel, coreset-based stochastic variational $\gp$ inference algorithm that:
\begin{enumerate}
    \item 
    Finds a coreset-based, sparse variational posterior to faithfully approximate the true $\gp$ posterior,
    enabling up- and down-weighting the influence of pseudo-points during learning (Section \ref{ssec:exp_coresets} and Appendix \ref{asssec:app_exp_posterior_predictive},\ref{asssec:app_exp_coreset_weights});

    \item 
    Maximizes a lower-bound over the marginal log-likelihood that is amenable to efficient stochastic optimization (Section \ref{ssec:cvtgp_lowerbound});
    
    \item 
    Provides a numerically stable algorithm requiring only $\bigO{M}$ parameters to be learned, at computational and memory complexities of $\bigO{M^3}$ and $\bigO{M^2}$ (Table~\ref{table:bigo}); 
    
    \item 
    Outperforms SOTA stochastic variational $\gp$ inference alternatives on real-world datasets (Section \ref{sec:experiments_iurteaga});
    CVGP not only provides improved predictive performance (Section \ref{ssec:exp_predictive}),
    but achieves a tighter lower variational bound than alternatives (Section \ref{ssec:exp_inference}).
\end{enumerate}

\section{BACKGROUND}
\label{sec:background}
We introduce the notation and foundations of $\gp$ regression in Section~\ref{ssec:background_gp},
and describe sparse approximations for scalable $\gp$ inference in Section~\ref{ssec:background_sparsegp}.
We review the variational inducing point foundational 
work of~\citet{titsias2009variational} and~\citet{hensman2013gaussian}, in Sections~\ref{ssec:background_titsias} and~\ref{ssec:background_svgp}, respectively.
These are SOTA $\gp$ algorithms that will serve as competitive baselines for the experiments in Section~\ref{sec:experiments_iurteaga}.

\subsection{GAUSSIAN PROCESS REGRESSION}
\label{ssec:background_gp}
A (univariate) $\gp$ is a non-parametric prior over functions from input domain $\xb \in \xspace$ into scalar space $y \in \yspace$,
denoted as $f(\xb ;\Theta) \sim \GP{m(\xb ; \theta_m), k(\xb, \xb ; \theta_k)}$.
A $\gp$ is specified by its
mean, $m(\xb ; \theta_m): \xspace \rightarrow \RR$,
and covariance (kernel), $k(\xb, \xb ; \theta_k) : \xspace \times \xspace \rightarrow \RR$, functions
with parameters $\theta_m$ and $\theta_k$ that are jointly referred to as $\gp$ hyperparameters $\Theta = \{\theta_m, \theta_k \}$.

The function $f$ is a mapping from $\xspace$ to the real numbers and we may equivalently write $f \in \RR^{\xspace}$,
viewing functions as (infinite-dimensional) vectors with elements indexed by members of $\xspace$.

Using vector notation, we define  $\fb=f(\Xb)$ as the vector containing the $\gp$ prior values at a collection of points $\Xb =\{ \xb_i \}_{i=1}^N$.
The $\gp$ prior evaluated at any subset of points $\Xb$ follows a multivariate Gaussian distribution
$\p{\fb ; \Theta } \sim \N{\fb \mid \zerob, \KbXX}$,
with $\KbXX = (k(\xb_i, \xb_j))_{1\leq i,j \leq N}$.
In the reminder, we assume zero mean $\gp$ priors without loss of generality,
and suppress explicit dependence on input points $\Xb$ to avoid notation clutter.

In $\gp$ regression with observations subject to Gaussian noise,
\ie $y = f(\xb;\Theta) + \epsilon$, $\epsilon \sim \N{\epsilon | 0, \sigma^2}$,
the data marginal likelihood is given by

\begin{align}
    \hspace*{-2.0ex} \p{\yb} \hspace*{-0.5ex} = \hspace*{-1.0ex} \int_{\fb} \p{\yb|\fb; \sigma} \p{\fb ; \Theta } \diff{\fb}  \hspace*{-0.5ex}=
    \hspace*{-0.5ex} \N{\yb | \zerob, \sigma^2 \Ib+\KbXX} . \hspace*{-2ex}
    \label{eq:data_marginal_analytical} 
\end{align}

Given some observed data $\Xb$, the posterior over the $\gp$ function at any input $\xbstar$, $f^\star = f(\xbstar)$, is Gaussian and computable in closed form, \ie

\begin{align}
& \cp{f^\star}{\xb^\star, \yb} = \N{f \mid m_{f^\star|\xb^\star}, k_{f^\star|\xb^\star} } , \; \text{ with} \label{eq:gp_posterior} \\
& \hspace*{2.0ex} m_{f^\star|\xbstar} = \KbSX \left( \sigma^2 \Ib+\KbXX\right)^{-1} \yb \;, \nonumber \\
& \hspace*{2.0ex} k_{f^\star|\xb^\star} = k(\xbstar, \xbstar) - \KbSX\footnotemark \left( \sigma^2 \Ib+\KbXX\right)^{-1} \KbXS \;. \nonumber
\end{align}

\footnotetext{$\KbSX$ is the $N$-dimensional row vector of kernel function values between a new input $\xbstar$ and observed data $\Xb$.}

\subsection{SPARSE GAUSSIAN PROCESS REGRESSION}

\label{ssec:background_sparsegp}
Even though posterior statistics in Equation~\eqref{eq:gp_posterior} are analytically tractable,
they raise computational challenges for big data, 
as they require computation of the inverse of $N \times N$ matrices
with, in general, $\bigO{N^3}$ time and $\bigO{N^2}$ space complexity.

An overview of sparse approximations to reduce such computational burden for $\gp$ regression can be found in~\citep[Chapter 8]{rasmussen2006gaussian},
with a unifying view presented in~\citep{quinonero2005unifying},
summarized below.
The innovation in sparse $\gp$s is to design \textit{approximate} posteriors
over $\gp$ function values $\fbZ=f(\XbZ)$ at a subset of $M$ inducing inputs $\XbZ$.
\footnote{
    The classical sparse $\gp$ literature~\citep{quinonero2005unifying, titsias2009variational} refers to inducing points with $\Zb$,
    and use $\ub=f(\Zb)$ for their corresponding $\gp$ values:
    we use a different notation here, for the sake of a clear and unified exposition of methods, easing the comparison across them.
}.

\citet{quinonero2005unifying} presented the Fully Independent Training Conditional (FITC) technique,
as a unifying framework for many of the sparse $\gp$ formulations that had previously been presented, \eg~\citep{csato2002sparse, smola2000,snelson2005sparse}.
FITC, which was later connected to methods that approximate the $\gp$ posterior via Expectation Propagation~\citep{snelson2008flexible,yuan2012,bui2017},
uses ---unlike previous methods~\citep{csato2002sparse, seeger2003fast}--- 
the marginal likelihood to jointly learn the hyperparameters and the inducing points~\citep{snelson2005sparse}.
This relaxes the constraint of having the inducing points limited to a subset of the dataset,
and turns a discrete inducing point selection problem into a continuous optimization one.

Careful inspection of these sparse methodologies and, in particular, FITC~\citep{quinonero2005unifying, bauer2016}
pointed out several limitations related to their tendency to overestimate marginal likelihood,
which motivated \citet{titsias2009variational} to propose a variational formulation for sparse $\gp$ regression.

\subsection{VARIATIONAL SPARSE GAUSSIAN PROCESS}
\label{ssec:background_titsias}

\citet{titsias2009variational} revisited sparse $\gp$ inference and pose it as a variational optimization problem on jointly learning $M$ inducing inputs $\XbZ$
(and $\gp$ hyperparameters $\Theta$),
by maximizing a lower-bound of the log-marginal likelihood:

\begin{align}
   \hspace*{-3ex} 
   \log \p{\yb} \geq \Loss_{SparseGP} = \Ex{\q{\fb,\fbZ}}{\log\frac{\p{\yb,\fb,\fbZ}}{\q{\fb,\fbZ}}} , \hspace*{-3ex}
    \label{eq:dataloglikelihood_lowerbound_titsias}
\end{align}

which is equivalent to minimizing the Kullback–Leibler (KL) divergence
between the variational family $q \in \mathcal{Q}$ and the $\gp$ posterior, \ie $\kl{\q{\fb,\fbZ}}{\p{\fb, \fbZ | \yb}}$.

\citet{titsias2009variational} showed that, for a factorization of the variational family of the $\q{\fb,\fbZ}=\cp{\fb}{\fbZ}\q{\fbZ}$ form,
one can marginalize over the $\gp$ inducing variables $\fbZ=f(\XbZ)$,
to derive the following analytical lower-bound

\begin{align}
    &\Loss_{SparseGP} = \hspace*{-0.5ex} \log \mathcal{N}\hspace*{-0.5ex}
    \left(
        \yb \mid
        \zerob, \sigma^2 \Ib+ \KbXZ \KbZZ^{-1} \KbZX
    \right) \nonumber\\
    &\qquad -\frac{1}{2 \sigma^2}\tr{\KbXX -   \KbXZ \KbZZ^{-1}\KbZX} ,
\label{eq:dataloglikelihood_lowerbound_titsias_analytical}
\end{align}

which can be computed in $\bigO{NM^2}$ time and $\bigO{NM}$ space complexity.
Equation~\eqref{eq:dataloglikelihood_lowerbound_titsias_analytical} is the result of integrating out the \emph{optimal} Gaussian variational posterior $q^*(\fbZ)$
---available in closed form for a set of inducing points $\XbZ$, 
and expressed in terms of the prior modeling choices of 
kernel and likelihood noise, only used implicitly in inference.

\subsection{STOCHASTIC VARIATIONAL GAUSSIAN PROCESS}
\label{ssec:background_svgp}

\citet{hensman2013gaussian} showed that \citet{titsias2009variational}'s evidence lower-bound (ELBO) can be revisited, and made amenable for stochastic variational inference for $\gp$s (SVGP),
by avoiding direct marginalization over inducing variables $\XbZ$ and re-organizing the ELBO as

\begin{align}
       \log p(\yb) \geq \Loss_{SVGP} &= \Ex{q(\fbZ)}{\Ex{q(\fb|\fbZ)}{\log p(\yb|\fb)}} \nonumber \\
       & \qquad - \kl{q(\fbZ)}{p(\fbZ)} \;.
\end{align}

SVGP proceeds by defining a free-form variational family $q(\fbZ) = \N{\fbZ \mid \mb, \Sb}$
and analytically computing the revised ELBO:

\begin{align}
    \Loss_{SVGP} &= 
    \log \N{\yb \mid \KbXZ\KbZZ^{-1}\mb, \sigma^2 \Ib} \nonumber \\
    & \hspace{2ex} - \frac{1}{2 \sigma^2} \tr{\KbXX - \KbXZ\KbZZ^{-1}\KbZX } \nonumber \nonumber \\
    & \hspace{2ex} - \frac{1}{2 \sigma^2} \tr{\KbXZ \KbZZ^{-1}\Sb \KbZZ^{-1} \KbZX} \nonumber \\
    & \hspace{2ex} - \kl{\q{\fbZ}}{\p{\fbZ}} \label{eq:dataloglikelihood_lowerbound_SVGP}  \;. 
\end{align}

Equation~\eqref{eq:dataloglikelihood_lowerbound_SVGP} now allows for data subsampling and is amenable to stochastic optimization
for learning the free variational parameters $\{\mb, \Sb\}$ in $q(\fbZ) = \N{\fbZ \mid \mb, \Sb}$ of order $\bigO{2M + M^2}$,
where an unbiased estimate of the SVGP loss can be computed with $\bigO{M^3}$ time- and $\bigO{M^2}$ space-complexity.

The optimum of Equation~\eqref{eq:dataloglikelihood_lowerbound_SVGP} matches that of Equation~\eqref{eq:dataloglikelihood_lowerbound_titsias_analytical}, 
yet the latter directly leverages the optimal variational distribution $q^*(\fbZ$),
while the former resorts to stochastic optimization of its free-form,
variational $\bigO{M^2}$ parameters to find it.
Namely, SparseGP operates by maximizing a tight
---based on the optimal $q^*(\fbZ)$)--- lower-bound,
with the disadvantage of not being able to use stochastic optimization. 

Our goal here is to leverage the best of each world
and to design a variational posterior that incorporates the dependencies set by the prior $\gp$ model (\ie the kernel and the likelihood noise)
for approximate $\gp$ inference that is amenable to stochastic optimization.

\section{CORESET BASED VARIATIONAL POSTERIOR GAUSSIAN PROCESS (CVGP)}
\label{sec:method}

We use Bayesian coreset principles to derive an sparse approximation to the true $\gp$ posterior that is learnable via \textit{stochastic} variational inference.

Bayesian coresets search for samples from a smaller data subset that can,
via weighted likelihoods, approximate otherwise hard to compute posterior distributions~\citep{huggins2016coresets, campbell2018bayesian, campbell2019automated, j-jubran2019}.
From an optimization perspective, Bayesian coresets can also be understood as a set of \textit{learnable} (observed or unobserved) points selected to minimize some divergence to a distribution of interest~\citep{ manousakas2020bayesian, manousakas2022black}.

Inspired by such framework,
we posit a coreset-based, variational posterior distribution for $\gp$s (CVGP):
\ie we learn a small subset of \emph{pseudo-inputs} $\XbC=\{\xb_1, \cdots, \xb_M\}$,
and \emph{pseudo-observations} $\ybC=\{y_1, \cdots, y_M\}$,
that if reweighted appropriately with parameters $\betabC=(\beta_{1}, \cdots, \beta_M)$,
approximate the $\gp$ posterior accurately.
Contrary to standard Bayesian coreset methodology,
the coreset tuple $\{\XbC, \ybC \}$ is composed by \emph{learnable pseudo-points} in the input-output data space
---not restricted to the observed empirical data. 
For accurate approximation of the posterior, 
and inspired by ~\citet{titsias2009variational}'s optimal solution,
we ensure that CVGP's posterior obeys the $\gp$ prior-conditional and it's inductive biases (see Section~\ref{ssec:cvtgp_coreset_posterior}).
We learn the CVGP posterior by formulating a variational lower-bound objective that is amenable to its stochastic maximization (see Section~\ref{ssec:cvtgp_lowerbound}).

\subsection{THE CORESET-BASED GAUSSIAN PROCESS POSTERIOR}

\label{ssec:cvtgp_coreset_posterior}
CVGP's key novelty is a coreset-based distribution 
designed to incorporate the $\gp$'s prior model characterization into the CVGP posterior.

We formulate a coreset-based distribution $\q{\fbC}$
over $\gp$ variables $\fbC=f(\XbC)$ at pseudo-inputs $\XbC =\{\xb_m\}_{m=1}^M$
and associated pseudo-observations $\ybC=\{y_m\}_{m=1}^M$,

\begin{align}
	&\cq{\fbC}{\XbC, \ybC, \betabC} \nonumber = \frac {\cq{\ybC}{ \fbC, \betabC} \cp{\fbC}{\XbC}}{\cp{\ybC}{\XbC, \betabC} } \nonumber \\
	&\qquad = \frac {\left(\prod_{m=1}^M p(y_{m} | f_{m})^{\beta_{m}} \right) p(\fbC|\XbC)}{p(\ybC | \XbC, \betabC) } \;, 
	\label{eq:coreset_gp_posterior_C_f}  
\end{align}

where the data likelihood for each pseudo-observation
$p(y_{m} | f_{m}), m \in \{1, \cdots, M\}$,
is raised to the power of learnable parameters $\betabC=(\beta_{1}, \cdots, \beta_M)$.

The CVGP posterior is a tempered distribution,
which can be understood as if
a small subset $M \leq N$ of pseudo-input/output pairs $\{\Xb_m, y_m\}$
are each drawn $\beta_m \geq 0$ times.

For a Gaussian observation likelihood,\footnote{
    Derivation of closed-form, coreset-based posteriors for non-Gaussian likelihoods is part of future investigations.
}
we derive in Appendix Section~\ref{assec:cvtgp_qf_analytical}
the closed-form multivariate Gaussian distribution 
of CVGP's posterior over $\gp$ function variables $\fbC$,
given coreset triplet $\{\XbC, \ybC, \betabC\}$:

\begin{align}
&\cq{\fbC}{\XbC, \ybC, \betabC} = \N{\fbC | \mb_{\fbC|\ybC}, \Kb_{\fbC|\ybC}}, \hspace*{-5ex}
\label{eq:coreset_gp_posterior_fC_analytical}
\\
\small
& \quad \mb_{\fbC|\ybC} = \KbCC(\KbCC + \Sigmab_{\betabC})^{-1} \ybC , \nonumber \\
& \quad \Kb_{\fbC|\ybC} = \KbCC - \KbCC(\KbCC + \Sigmab_{\betabC})^{-1}\KbCC , \nonumber 
\end{align}

where $\Sigmab_{\betabC}= \sigma^2 \cdot \diag{\betabC^{-1}}$.
With this coreset-based distribution $q(\fbC),$\footnote{
The interested reader can find the complementary weight-space derivations in Appendix Section~\ref{assec:cvtgp_weightspace}.
}
we now accommodate the $\gp$ prior's conditional dependency,
$\q{\fb, \fbC} = \cp{\fb}{\fbC} \q{\fbC}$, 
where 

\begin{align}
    \cp{\fb}{\fbC} &= \N{\fb | \mb_{\fb|\fbC}, \Kb_{\fb|\fbC}} \;, \text{ with} \hspace*{-5ex} \label{eq:gp_cond_coreset}\\
    \mb_{\fb|\fbC} &= \KbXC\KbCC^{-1} \fbC , \nonumber \\
    \Kb_{\fb|\fbC} &= \KbXX  - \KbXC\KbCC^{-1} \KbCX , \nonumber
\end{align}

and compute the variational posterior of interest,
\ie $\cq{\fb}{\XbC, \ybC, \betabC}$ over $\gp$ function values $\fb=f(\Xb)$,
by marginalizing the coreset-based, tempered posterior of Equation~\eqref{eq:coreset_gp_posterior_fC_analytical}
from the joint distribution $\q{\fb, \fbC}$. The resulting CVGP coreset-based variational posterior is

\begin{align}
    &\cq{\fb}{\XbC, \ybC, \betabC} = \N{\fb | \mb_{\fb|\ybC}, \Kb_{\fb|\ybC}}\;, 
    \label{eq:coreset_gp_posterior_f_analytical} \\
\small
    & \hspace{0.5ex} \mb_{\fb|\ybC} = \KbXC\left( \KbCC + \Sigmab_{\betabC} \right)^{-1} \ybC \;, \nonumber\\
    & \hspace{0.5ex} \Kb_{\fb|\ybC} = \KbXX  - \KbXC \left(\KbCC + \Sigmab_{\betabC}\right)^{-1} \KbCX . \nonumber 
\end{align}

If one were to follow standard Bayesian coreset procedures,
we would directly aim to learn the coresets
that best approximate $\cq{\fb}{\XbC, \ybC, \betabC}$ to the true posterior
---which requires computing the $\gp$ posterior in Equation~\eqref{eq:gp_posterior} of $\bigO{N^3}$ complexity \citep{manousakas2022black}.
On the contrary, we learn the coreset triplet $\{\XbC, \ybC, \betabC\}$
using a variational objective that aims to minimize the divergence between such two distributions,
at reduced computational cost, and in a form amenable to its stochastic minimization.

\subsection{CVGP'S VARIATIONAL LOWER-BOUND}
\label{ssec:cvtgp_lowerbound}

We denote with $\q{\cdot}$ a generic variational family of distributions over a $\gp$.
Whenever $\q{\fb} \neq \cp{\fb}{\yb}$, we can lower-bound the log-marginal distribution,
incurring on a gap determined by the Kullback–Leibler (KL) divergence between the variational distribution 
and the true $\gp$ posterior:
\begin{align}
    \log\p{\yb } \geq \Loss & = \Ex{\q{\fb}}{\log \cp{\yb}{\fb}} \nonumber - \kl{\q{\fb}}{\p{\fb}} \;.
\end{align}
Hence, maximizing the loss $\Loss$ is equivalent to minimizing 
the KL divergence between the variational family $q(\fb)$ and the true posterior $p(\fb|\yb)$,
\ie $\Delta(\fb) = \kl{\q{\fb}}{\cp{\fb}{\yb}}$.

In CVGP, we use the coreset-based posteriors of Equations~\eqref{eq:coreset_gp_posterior_fC_analytical} and~\eqref{eq:coreset_gp_posterior_f_analytical}
to maximize the lower-bound, \ie

\begin{align}
\Loss_{CVGP} &= \Ex{q(\fb|\ybC, \XbC, \betabC)}{\log p(\yb|\fb) } \nonumber\\
& \quad - \kl{q(\fbC|\ybC, \XbC, \betabC)}{p(\fbC)} \;,
\label{eq:loss_coreset_posterior_gp}
\end{align}

which has the following analytical solution:
\begin{align}
&\Loss_{CVGP}\hspace{-0.2ex}=\hspace{-0.2ex}\log \N{\yb | \mb_{\fb|\ybC},\hspace{-0.1ex}\sigma^2 \Ib}\hspace{-0.2ex}-\hspace{-0.2ex}\frac{1}{2 \sigma^2}\tr{\Kb_{\fb|\ybC}} \nonumber \\
&+ \frac {1}{2} \left[ \tr{\Ab \KbCC} \right.  - \ybC^\top
        \Ab \KbCC \Ab \ybC  + \left. \ln \left|\Ab \Sigmab_{\betabC}\right| \right] \:,
\label{eq:loss_coreset_posterior_gp_analytical}
\end{align}
where $\Ab = \left(\KbCC + \Sigmab_{\betabC}\right)^{-1}$
---details of the derivation are provided in Appendix Section~\ref{asec:cvtgp_derivation}.

\paragraph{$\Loss_{CVGP}$: optimality at reduced complexity and increased numerical stability.}
Maximization of the variational lower-bound in Equation~\eqref{eq:loss_coreset_posterior_gp_analytical}
to learn CVGP's coreset-based posterior in Equation\eqref{eq:coreset_gp_posterior_f_analytical}
results in the following desirable properties:
\begin{enumerate}
    \item The maximum of Equation~\eqref{eq:loss_coreset_posterior_gp_analytical} is identical to the loss in Equation~\eqref{eq:dataloglikelihood_lowerbound_titsias_analytical} derived by~\citet{titsias2009variational}:
    \ie SparseGP and CVGP \emph{have the same optimum} 
    ---see proofs in Appendix Section~\ref{assec:cvgp_lower_bound_optimum}.
    
    \item The lower-bound in Equation~\eqref{eq:loss_coreset_posterior_gp_analytical}
    is amenable to data-subsampling.
    Due to the uncorrelated Gaussian likelihood term and properties of the trace,
    we can apply \emph{stochastic optimization} for its maximization, computing unbiased loss estimates with a single data sample.
    
    \item The algorithmic complexity of CVGP, for coreset size $M$, 
    is $\bigO{M^3}$ in computational time and $\bigO{M^2}$ in space complexity.
    Importantly, CVGP's parameter complexity is of \emph{reduced $\bigO{M}$ order}, as it only requires learning coreset triplets $(\XbC, \ybC, \betabC)$, each of size $M$
    ---see Table~\ref{table:bigo} for a full comparison.
    
    \item CVGP's posterior and lower-bound inherently provide a numerically stable stochastic algorithm, as all matrix inverse operations in Equation\eqref{eq:loss_coreset_posterior_gp_analytical}
    involve sums of a diagonal matrix defined by
    coreset weights ($\Sigmab_{\betabC}$)
    and positive definite matrix $\KbCC$\footnote{
    In theory, $\KbCC$ is positive definite and invertible.
    However, in practice, numerical issues can cause instability.
},
    \ie $\Ab = \left(\KbCC + \Sigmab_{\betabC}\right)^{-1}$.
\end{enumerate}

\begin{table}[!ht]
    \rowcolors{5}{}{gray!10}
    \begin{tabular}{*4c}
        \toprule
        & \multicolumn{3}{c}{Complexities} \\
        \cmidrule(lr){2-4}
        Algorithm & Time  & Space & Parameter\\    
        \midrule
        SparseGP &  $\;$$\;$$\;$$ {\bigO{NM^2}}$  &  $\;$$\;$$\;$${\bigO{NM^2}}$    & $\boldsymbol{{\bigO{M}}}$\\
        SVGP  &     $\boldsymbol{{\bigO{M^3}}}$   & $\boldsymbol{{\bigO{M^2}}}$ &      $\;$$\;$${\bigO{M^2}}$ \\
        \textbf{CVGP} &  $\boldsymbol{{\bigO{M^3}}}$      & $\boldsymbol{{\bigO{M^2}}}$ &  $\boldsymbol{{\bigO{M}}}$ \\
        \bottomrule
    \end{tabular}
    \centering
   \caption{{Computational analysis of CVGP and sparse variational $\gp$ alternatives:
    time and space complexities for obtaining an unbiased estimate of
    their objectives.
    Desirable complexities are highlighted in \textbf{bold}.
    CVGP enjoys same time and space complexity as SVGP,
    at a reduced variational parameter dimensionality.
    \label{table:bigo}
    }}
\end{table}
\subsection{A COMPARISON TO ALTERNATIVES}
\label{ssec:cvtgp_discuss}

CVGP is, to the best of our knowledge,
the first variational $\gp$ inference method that leverages
a coreset-based posterior for efficiency and scalability.
It diverges from alternative sparse $\gp$ inference techniques in that
its posterior is based on a coreset triplet $\{\XbC, \ybC, \betabC\}$.
Therefore:
\begin{itemize}
    \item CVGP is not restricted to a sparse selection of observed inputs:
    $\XbC$ is a vector of free parameters, within the data domain,
    but \textbf{not restricted to the empirical data}.
    
    \item CVGP does not learn inducing variables $\Ex{\q{\fbZ}}{\fbZ} = \mb$,
    \ie  posterior $\gp$ mean function values evaluated at inducing points $\XbZ$.
    Instead, it \textbf{learns pseudo-observations $\ybC$} that induce (\ie capture the characteristics of) the observed data (Figure \ref{fig:exp_coresets_histograms}).

    \item CVGP is the only existing $\gp$ method that \textbf{reweights the pseudo-observations} with learnable parameters $\betabC$, for flexibility and explainability of its coreset-based posterior:
    \ie it learns which pseudo-points are important
    for accurate $\gp$ posterior approximation (Figures \ref{fig:exp_coresets_predictive} and \ref{fig:exp_coresets_histograms}).
\end{itemize}

\paragraph{Comparison to non-variational sparse $\gp$s.} 
Selection of $\gp$ inputs from within the training data involves a prohibitive combinatorial optimization
that may require greedy optimization~\citep{csato2002sparse},
based on posterior maximization~\citep{smola2000},
maximum information gain~\citep{seeger2003fast},
matching pursuit~\citep{keerthi2005},
or other techniques~\citep{quinonero2005unifying}.

On the contrary, CVGP leverages \emph{stochastic optimization}
to find a weighted subset of pseudo-points
that efficiently approximate the $\gp$ posterior,
sharing resemblance with the pioneer work of~\citet{snelson2005sparse}.
To circumvent overestimation of the marginal likelihood and under-estimation of the noise variance as reported by~\citet{titsias2009variational,bauer2016},
CVGP resorts to variational inference.
Hence, CVGP shares the variational formulation of \citet{titsias2009variational} and \citet{hensman2013gaussian},
yet is distinct in several important aspects.

\paragraph{Comparison to variational sparse $\gp$s.}
CVGP aligns with \citet{titsias2009variational} in the use of a variational lower-bound on the marginal log-likelihood that leverages the $\gp$ prior’s conditional dependency, \ie $\q{\fb, \fbC} = \cp{\fb}{\fbC} \q{\fbC}$, and analytically marginalizes $\q{\fbC}$. In contrast, SVGP does not marginalize this distribution and devises a different lower-bound for stochastic optimization. As a result, SparseGP and CVGP posteriors directly incorporate the $\gp$ prior’s inductive biases and the likelihood model. The main difference is in the choice of $\q{\fbC}$:

\begin{itemize}

    \item SparseGP derives the optimum distribution at inputs $\XbZ$ over function values $\fbZ$, given observed data $\yb$:
    
    \small{
    \begin{align}
	& q^\star(\fbZ)=\N{\fbZ; \mbstar{\fbZ}, \Kbstar{\fbZ}{\fbZ}} \label{eq:sparseGP_optimal_q} \;, \text{with } \\
    & \hspace{-2ex} \begin{cases}
        \mbstar{\fbZ} = \KbZZ \red{\left(\sigma^2 \KbZZ + \KbZX \KbXZ\right)^{-1}} \green{\KbZX \yb}\\ 
        \Kbstar{\fbZ}{\fbZ} = \KbZZ  \blue{\left( \KbZZ + \frac{1}{\sigma^2} \KbZX \KbXZ\right)^{-1}} \KbZZ
    \end{cases} \nonumber 
    \end{align}
    }\normalsize
    
    \item CVGP defines a learnable distribution $q(\XbC)$ with free coreset parameter triplet $\{\XbC, \ybC, \betabC\}:$
    
    \small{
    \begin{align}
	& \q{\fbC}=\N{\fbC; \mb_{\fbC|\ybC}, \Kb_{\fbC|\ybC}}  \label{eq:cvtgp_q}\;,\text{with} \\
        & \hspace{-2ex} \begin{cases}
		 \mb_{\fbC|\ybC} = \KbCC \red{\left( \KbCC + \SigmabetaC \right)^{-1}} \green{\ybC} \\
		\Kb_{\fbC|\ybC} = 
       \KbCC \blue{\left[\KbCC^{-1} -  \left( \KbCC + \SigmabetaC \right)^{-1} \right]} \KbCC  \nonumber\\
	\end{cases}
    \end{align}
    
    }
\end{itemize}

The building blocks of CVGP's coreset based posterior
are analogous to SparseGP's optimal posterior:
CVGP's learned pseudo-observations \green{$\ybC$}
can be viewed as a weighted combination of observed datapoints,
\ie the \green{$\KbZX \yb$} term in SparsedGP's posterior mean.
CVGP pseudo-observations \green{$\ybC$} are modulated
by the \red{$\left( \KbCC + \SigmabetaC \right)^{-1}$} term in its posterior mean;
in SparseGP, the \red{$\left(\sigma^2 \KbZZ + \KbZX \KbXZ\right)^{-1}$} term
similarly weights the transformed observations \green{$\KbZX \yb$}.
In both posterior distributions,
these terms in red are responsible for balancing
the prior inductive biases with the information provided by observed data:
\ie the posterior means interpolate between the prior and observations.

A similar dependency between the prior and the information provided by data
is observed in the posterior covariances:
\ie the blue terms in both posteriors adapt the prior covariance to account for the uncertainty reduction due to observations.
In CVGP, this balance is adjusted through the learnable matrix \blue{$\SigmabetaC$},
whereas in SparseGP, it is determined by the fixed dependency \blue{$\frac{1}{\sigma^2}\KbZX\KbXZ$} set by the prior covariance and the likelihood noise.

Notably, as shown in Appendix Section~\ref{assec:cvgp_lower_bound_optimum},
when CVGP matrix $\SigmabetaC$ matches the appropriate weighting,
the optimum of SparseGP and CVGP's loss-functions are identical.
Hence, the learned solutions match
with $\ybC = \sigma^{-2}\SigmabetaC^* \yb$
and $\SigmabetaC^* = \sigma^{2}\KbCC \left(\KbCX \KbXC\right)^{-1} \KbCC$,
recovering ~\citet{titsias2009variational}'s optimal solution.
We empirically showcase CVGP's ability to quickly and
\textbf{efficiently close the gap to ExactGP's marginal log-likelihood}
in Section~\ref{ssec:exp_inference}.
Contrary to SparseGP,
CVGP's loss in Equation~\eqref{eq:loss_coreset_posterior_gp_analytical} 
is amenable to stochastic optimization,
making sparse $\gp$ regression scalable at reduced complexity.

CVGP matches SVGP’s scalability \citep{hensman2013gaussian},
yet offers two key advantages: parameter complexity of order $\bigO{M}$ and a distinct optimization landscape.
These arise from $q(\fbC)$: whereas SVGP’s free-form family $q(\fbZ) = \N{\fbZ \mid \mb, \Sb}$ requires $\bigO{M^2}$ parameters and yields statistics ($\mathbf{m}, \mathbf{S}$) not directly tied to the model or data likelihood,
CVGP’s posterior leverages the model’s inductive biases,
acting as a \textbf{natural interpolation between the $\gp$ prior and the data likelihood}\footnote{
We analyze CVGP’s noise prior to posterior adaptation as a function of observation noise
in Appendix~\ref{assec:app_exp_noisy}.
}.
These structural differences produce distinct loss landscapes, with SVGP’s higher-dimensional optimization often struggling to converge, as shown in Section~\ref{ssec:exp_inference}.

\subsection{CVGP AS BAYESIAN CORESET LEARNING}

\label{ssec:cvtgp_inference}
CVGP enables a complementary, Bayesian coreset learning-based view of sparse $\gp$ inference.

Methodologically, CVGP maximizes the loss in Equation~\eqref{eq:loss_coreset_posterior_gp_analytical} for $\gp$ posterior inference;
\ie it maximizes the variational lower-bound $\Loss_{CVGP}$
with respect to CVGP parameters $\{\XbC, \ybC, \betabC\}$,
encouraging approximations that minimize the gap to the true $\gp$ posterior.
We note that,
$\Loss_{CVGP} \rightarrow \Loss = \log \p{\yb} \;$
implies $\Delta_{CVGP}=\kl{\cq{\fb,\fbC}{\XbC, \ybC, \betabC}}{\cp{\fb,\fbC}{\yb}} \rightarrow 0$.
Hence,
CVGP learns coresets that minimize the distance between
its variational distribution
and the true $\gp$ posterior.
To do so, it finds ---indirectly, yet efficiently--- a sparse representation of the data (\ie the coreset triplet)
that captures as much information as the posterior of interest,
measured by the KL divergence between the true and CVGP's posterior.


Initial estimates of the coreset triplet $\{\XbC, \ybC, \betabC\}$
can be selected randomly or using k-means.
We evaluate CVGP's robustness to coreset initialization
in Appendix \ref{asssec:app_exp_robustness} and \ref{asec:qual_study}. 
Importantly, CVGP's learning procedure enables an
\textbf{automatic relevance determination of pseudo-points} $\{\XbC, \ybC\}$
via adaptation of their $\betabC$ values:
\ie CVGP has the inherent flexibility to
up- or down-weight (``ignore'') the pseudo-points that are
deemed (or not) important to describe the observed data
---see experiments in Section~\ref{ssec:exp_coresets}.

Finally, we note that CVGP's coreset-based variational posteriors,
when derived from the function-space and weight-space views of $\gp$s
---see Appendix Section~\ref{asec:cvtgp_derivation} for both derivations---
provide complementary posterior insights.
Namely, $\cq{\fb}{\XbC, \ybC, \betabC}$,
illustrates which learned coreset tuples $\{\XbC, \ybC\}$ weighted by $\betabC$,
help describe the $\gp$ posterior best
---we provide examples in Figure~\ref{fig:exp_coresets_predictive}.
The latter, $\cq{\wb}{\XbC, \ybC, \betabC}$, enables interpretability over 
which input features regress $\yb$ from $\Xb$ more accurately,
via inspection of the posterior weights $\wb$, for finite-dimensional feature vectors.

\section{EXPERIMENTS}
\label{sec:experiments_iurteaga}
\begin{figure*}[!th]
    \centering
    \includegraphics[width=\textwidth]{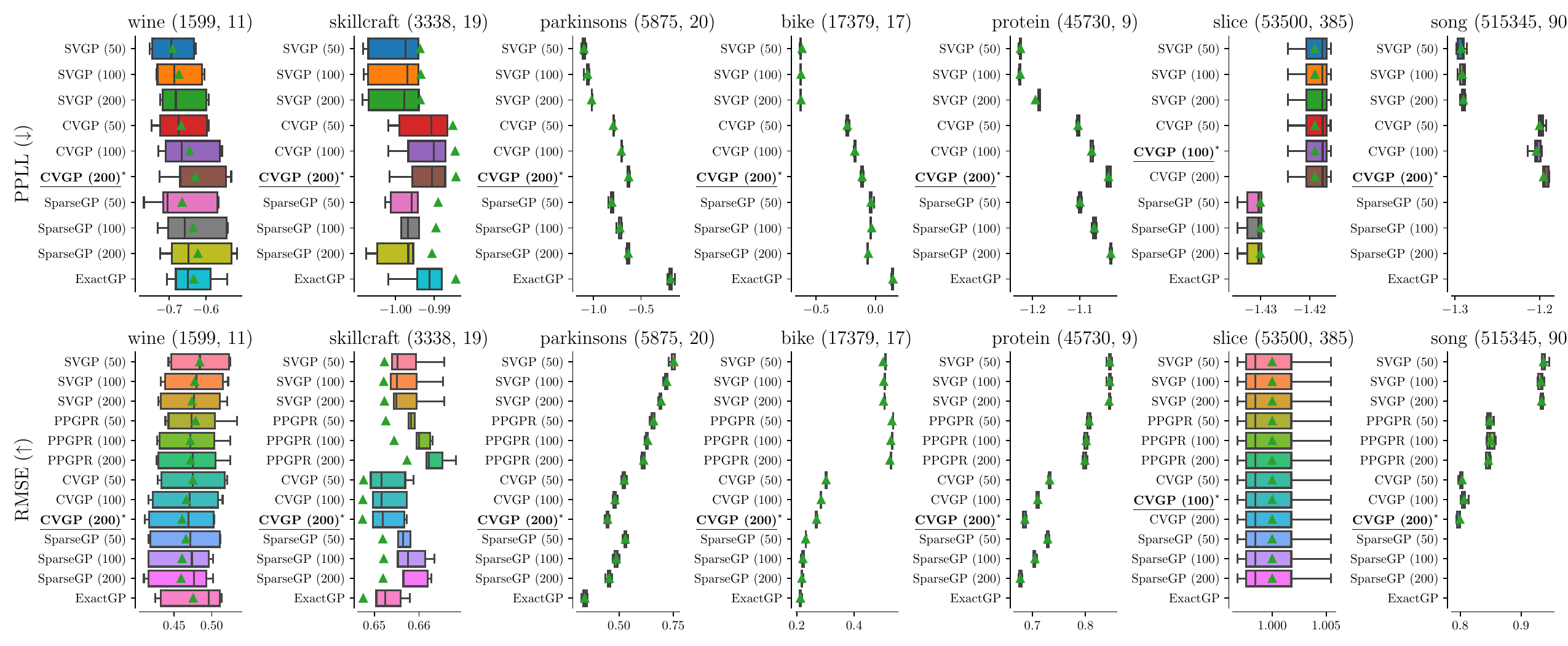}
    \caption{{
        Box-and-whisker diagrams of predictive metrics (RMSE and PPLL) on real datasets.
        The titles denote the dataset and in parenthesis, its size and feature-dimensionality.
        Arrows indicate the desirable metric direction: higher PPLL, lower RMSE. CVGP outperforms SVGP and PPGPR, and is on-par with SparseGP,
        with as few as $50$ coresets. The best performing \emph{stochastic gradient} model mean statistic (\textcolor{darkgreen}{\scalebox{1}{$\blacktriangle$}}) is $\textbf{\underline{emphasized}}^\star$.
        SparseGP and ExactGP results are omitted for the largest datasets due to computational complexities.
        }}
    \label{fig:exp_predictive_real_all}
\end{figure*}

We demonstrate CVGP's superior predictive performance in real-datasets in Section~\ref{ssec:exp_predictive},
before delving into its inference advantages in Section~\ref{ssec:exp_inference}.
We showcase the quality and explainability of the learned CVGP posteriors in Section~\ref{ssec:exp_coresets}.

\subsection{EXPERIMENTAL SETUP}
\label{ssec:exp_setup}

We compare CVGP against benchmark $\gp$ alternatives described in Section~\ref{sec:background}:
ExactGP~\citep{rasmussen2006gaussian},
SparseGP~\citep{titsias2009variational},
and SVGP~\citet{hensman2013gaussian}.
We also incorporate Parametric Gaussian Process Regressors (PPGPR) by~\citet{jankowiak2020parametric}
as a strong predictive baseline.
We implement CVGP using Pytorch and GPyTorch libraries,
and use benchmark GPytorch implementations \citep{gardner2018gpytorch} for the baselines.

We use a zero-mean $\gp$ prior with a Radial basis kernel function (RBF) in all experiments.
We evaluate different coreset (CVGP) and inducing point sizes $M$ for sparse $\gp$ baselines (SparseGP, SVGP and PPGPR),
all initialized with k-means~\citep{hartigan1979algorithm},
enforcing best validation RMSE performance as early stopping criteria.
We employ 5-fold cross-validation to compute and report each technique's
predictive root-mean-squared error (RMSE)
and posterior predictive log-likelihood (PPLL),
over held out test splits.

All details for the reproducibility of the experiments are provided in Appendix Section~\ref{assec:reproducibility}.
CVGP predictive and inference experiments of Sections~\ref{ssec:exp_predictive} and ~\ref{ssec:exp_inference}
are based on real-world regression datasets from the UCI machine learning repository data~\citep{asuncion2007uci}.
We use simulated datasets to showcase learned predictive posteriors in Section~\ref{ssec:exp_coresets},
with all dataset details described in Appendix Section~\ref{assec:datasets}.

\subsection{PREDICTIVE PERFORMANCE}
\label{ssec:exp_predictive}

We assess the predictive performance of all sparse $\gp$ methods for different real-datasets,
and illustrate the performance of ExactGP
---when computationally possible--- as the optimal benchmark,
in Figure~\ref{fig:exp_predictive_real_all}.

CVGP outperforms \emph{stochastic} sparse $\gp$ alternatives (SVGP and PPGPR) consistently,
with performance on par with SparseGP across all predictive metrics
---we inspect the learning and inference gaps between methods in Section~\ref{ssec:exp_inference} and Appendix \ref{asec:gap_results}.

Although CVGP, SVGP and SparseGP share the same theoretical optimum,
empirical predictive performance in Figure~\ref{fig:exp_predictive_real_all} showcases that
SVGP rarely reaches the desirable performance of SparseGP,
while CVGP's is consistently similar to SparseGP
---recall that SparseGP does not allow for stochastic optimization, while CVGP does.

CVGP's performance improves with increaset coreset size and ---with as little as 50 coresets---
consistently outperforms alternative stochastic methods,
even when these baselines use \emph{4-times} more inducing points, \ie SVGP (200) and PPGPR (200).

 \begin{figure}[!ht]
     \includegraphics[width=0.45\textwidth]{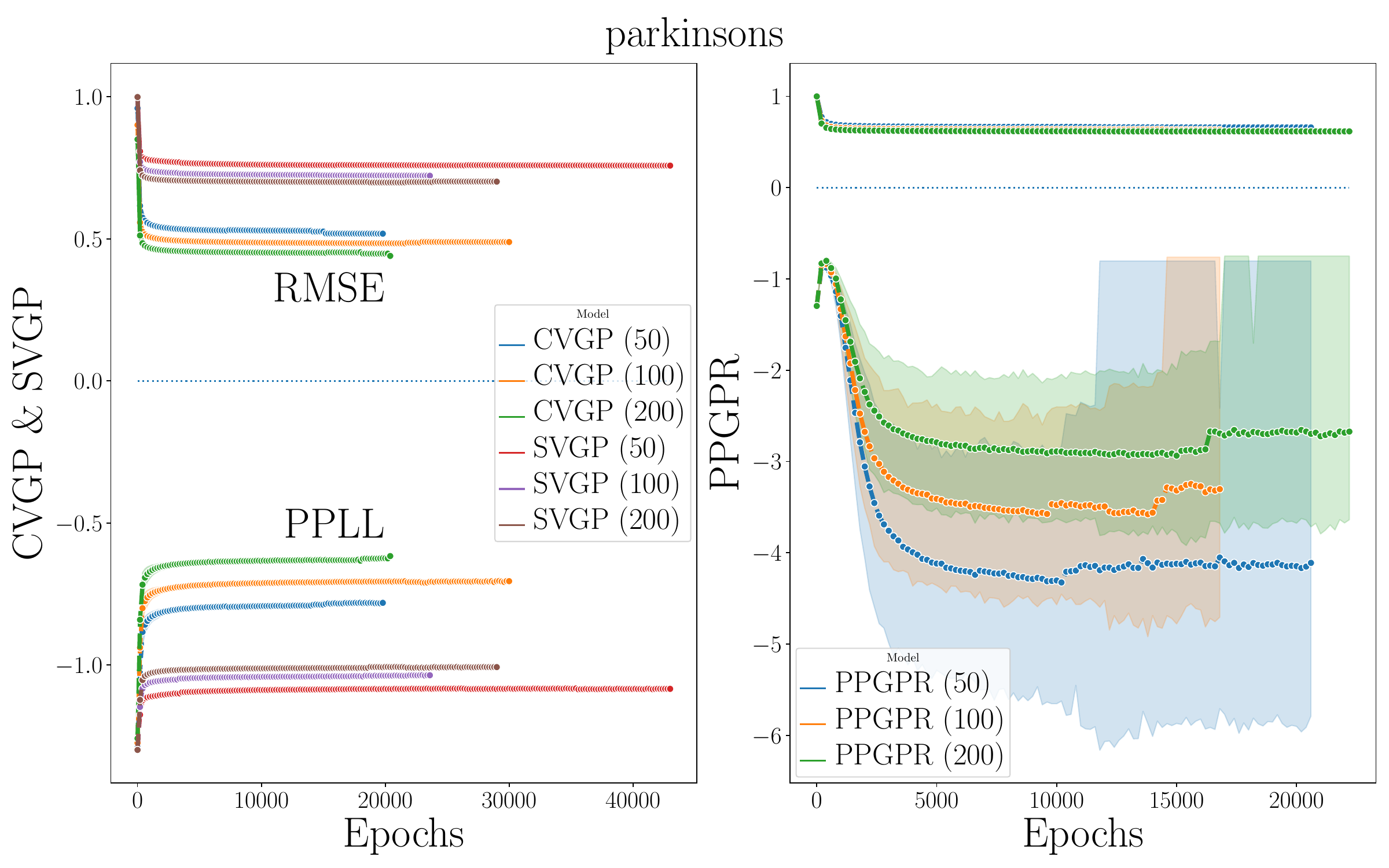}
    \caption{{
         Evolution of RMSE and PPLL across training epochs. CVGP and SVGP's RMSE and PPLL consistently decrease with training epochs. Even though PPGPR's RMSE improves over epochs, its PPLL deteriorates -- indicating some form of overfitting.
         }
     }
     \label{fig:exp_predictive_training_parkinsons}
     \vspace{-3ex}
\end{figure}

CVGP's predictive performance is also better than PPGPR,
an approximate $\gp$ algorithm specifically designed for predictive performance.
We showcase in
Figure~\ref{fig:exp_predictive_training_parkinsons} and Appendix \ref{asec:epoch_results}
the evolution of RMSE and PPLL across training,
where training of models does not stop until there are no RMSE improvements.
Notice that, while CVGP metrics improve consistently over training,
the RMSE for PPGPR improves, while its PPLL deteriorates over training epochs.
\footnote{
Due to the large negative PPLL values of PPGPR,
we have not reported them in Figure \ref{fig:exp_predictive_real_all}. 
}

We demonstrate CVGP's predictive performance robustness to initialization in Figure~\ref{fig:exp_predictive_cvgp_randomcvgp_real} in Appendix Section~\ref{asssec:app_exp_robustness}.
We notice k-means and randomly initialized CVGPs' performance
to be similar across metrics and datasets,
which is likely due to the coreset-based posteriors' flexibility
to up- and down-weight pseudo-input/output pairs via $\betabC$, \emph{a property other methods do not pose}.

\subsection{INFERENCE PERFORMANCE}
\label{ssec:exp_inference}

\begin{figure*}[!h]
    \centering
\includegraphics[width=\textwidth]{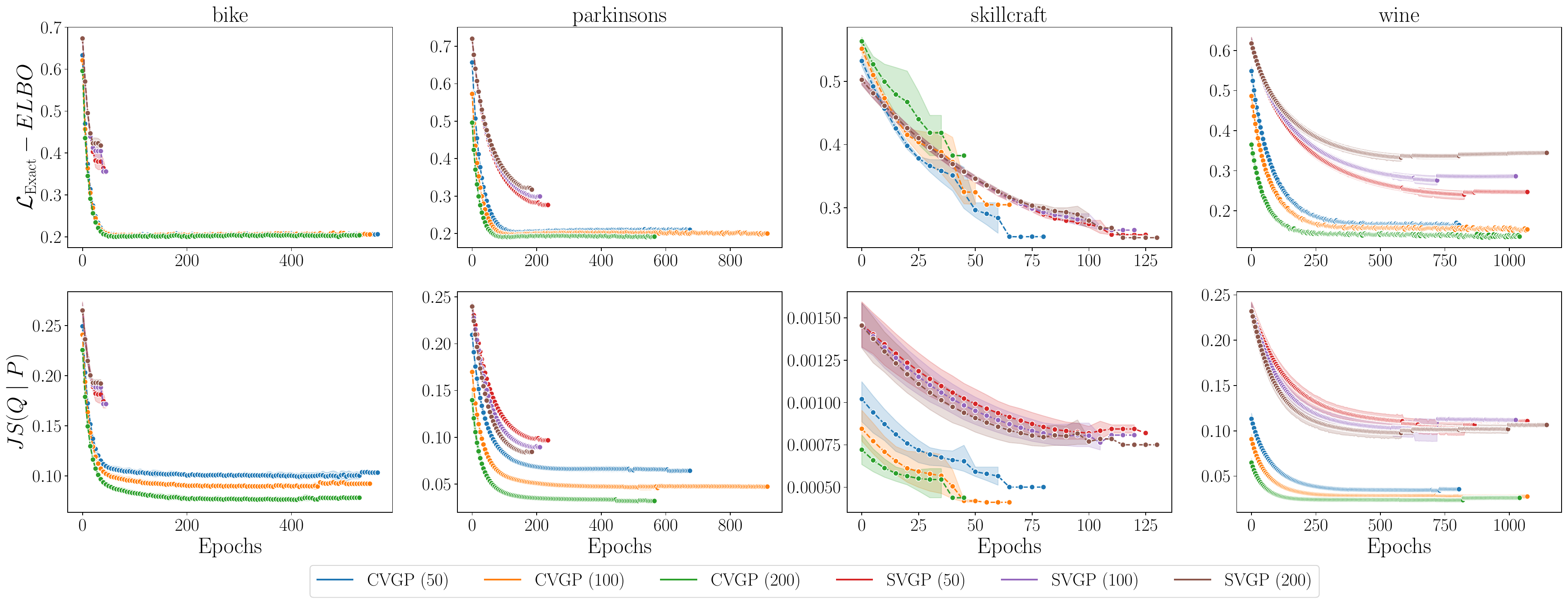}
        \caption{\small{
        Learning and inference gaps for sparse $\gp$ methods over training,
        as measured by
        (top-row) the difference between the log-marginal of ExactGP and the variational bound for SVGP and CVGP; and
        (bottom-row) the Jensen-Shannon divergence between the exact posterior predictive and each method's approximate posterior predictive.
        CVGP provides a better approximation to the exact $\gp$ posterior,
        consistently optimizing a tighter lower-bound. We showcase a more detailed figure including PPGPR in Appendix \ref{asec:gap_results}.}}
        \label{fig:exp_inference_gap}
\end{figure*}

We investigate why CVGP approximates the $\gp$ posterior predictive distributions more accurately,
by studying the relationship between the variational lower-bounds ($\mathcal{L}$)
of sparse $\gp$ alternatives and the true $\gp$ marginal log-likelihood in Equation~\eqref{eq:data_marginal_analytical}:
\ie the difference between the log-marginal of ExactGP and the variational loss optimized by SVGP and CVGP. 
We also show in Figure~\ref{fig:exp_inference_gap}
the inference gap of these methods while in training, over held-out datasets,
using the Jensen-Shannon divergence between the exact posterior predictive distribution 
$\cp{\fb^\star}{\xb^\star, \yb}=\int \cp{\fb^\star}{\xb^\star, \fb}\cp{\fb}{\yb} \diff{\fb}$ in Equation~\eqref{eq:gp_posterior},
and each method's approximate posterior predictive $\cq{\fb^\star}{\xb^\star}=\int \cp{\fb^\star}{\xb^\star, \fb_M}\q{\fb_M} \diff{\fb_M}$.
We employ fixed, equal $\gp$ prior hyperparameters for all models.

Results in Figure~\ref{fig:exp_inference_gap} demonstrate how
CVGP closes the learning gap to ExactGP better. On the contrary, SVGP offers a looser bound even if, in theory, both loss functions have the same optimum.
Moreover, smaller divergence from CVGP's posterior to that of ExactGP
suggests that CVGP better approximates the $\gp$ posterior of interest,
at only $\bigO{M}$ parameter, $\bigO{M^3}$ time and $\bigO{M^2}$ space complexities.
This improvement is attained with as little as 50 coresets
---performance not reached by SVGP even with 200 inducing points.

We argue that this performance gap
is the result of the distinct optimization landscapes
of the former compared to the latter,
induced by the lower-dimensionality of CVGP's optimization problem
and the explicit inductive biases present in CVGP's posterior:
($i$) ability to interpolate easily between prior and posterior
(see Appendix~\ref{assec:app_exp_noisy}), and
($ii$) ability to learn informative pseudo-points
---further investigated below and Appendix \ref{asec:qual_study}.

\subsection{CVGP AS BAYESIAN CORESET LEARNING}
\label{ssec:exp_coresets}

\begin{figure*}[!h]
    \includegraphics[width=\textwidth]{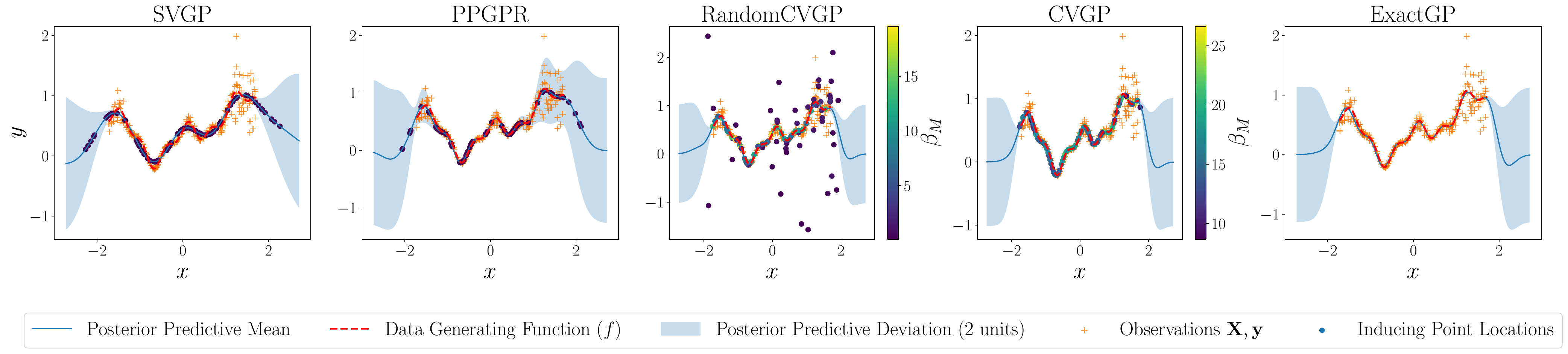}
    \caption{{
    True data generating function ({\red{\textbf{---}}}), posterior predictive mean ({\textbf{\blue{---}}}), and
    2-unit credible intervals (shaded)
    for the stochastic $\gp$ methods.
    We indicate  the inducing variables learned by SVGP and PPGPR, and for CVGP, the learned coreset pseudo-points $\{\XbC, \ybC\}$, with each pseudo-point's color intensity weighted by the learned $\beta_m$ on the right hand-side bars.
    Notice CVGP's high-quality posterior, most similar to that of ExactGP, which serves as gold standard. All methods revert to the prior for ranges where \emph{no data} has been observed.
    }}
    \label{fig:exp_coresets_predictive}
\end{figure*}
We illustrate posterior predictive distributions learned by stochastic sparse $\gp$ methods in Figure~\ref{fig:exp_coresets_predictive},
for a 1-dimensional synthetic dataset.

We observe CVGP's approximate posterior to be closest to the exact predictive posterior,
both in predicted mean and uncertainty quantification.
On the contrary, SVGP and PPGPR encounter difficulties in accurately modeling
the function of interest and their uncertainty in the $x \in (0,2)$ range:
SVGP computes a \emph{low-uncertainty}, \emph{smooth} posterior predictive mean, while PPGPR captures the mean but overestimates uncertainty for $x \in (1,2)$. CVGP, regardless of initialization, better handles this noisy region, matching ExactGP’s mean and uncertainty by learning coreset triplets ${\XbC, \ybC, \betabC}$ with up-weighted $\ybC$ that mitigate posterior $\gp$ uncertainty overestimation.

\begin{figure}[!h]
    \includegraphics[width=0.45\textwidth]{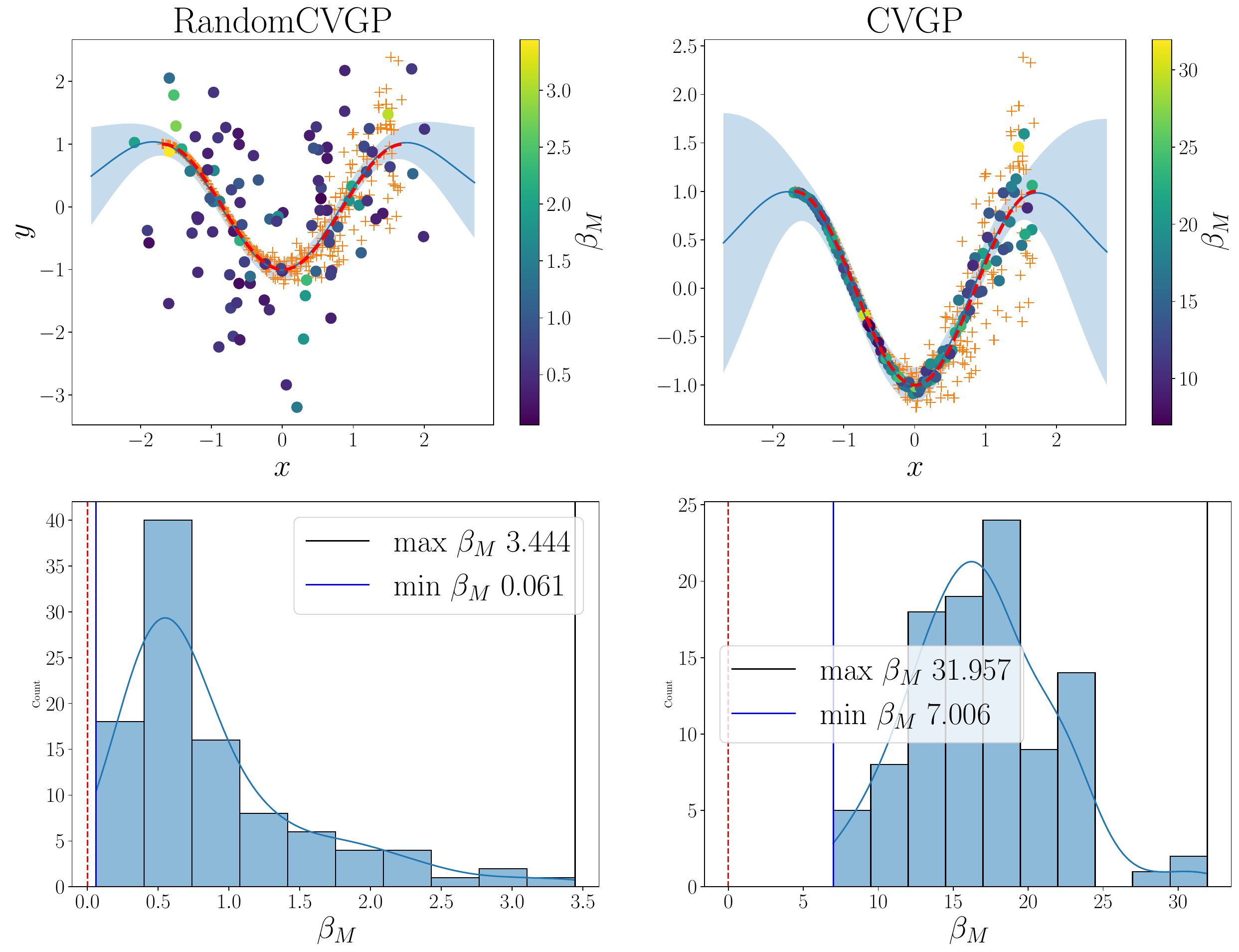}
    \caption{{ Histogram of learned coreset weights for CVGP with random (RandomCVGP) and k-means initialization. Legends are same as Figure \ref{fig:exp_coresets_predictive}'s. CVGP down-weights uninformative data, yielding many $\betabC\approx0$ for RandomCVGP (removing unhelpful points from the approximate posterior).
    \emph{Unlike other inducing-point methods
    ---which must learn good locations---
    CVGP can eliminate (down-weight) points that do not converge to plausible values.}}
    }
    \label{fig:exp_coresets_histograms}
\end{figure}

The input locations  $\XbC$ learned by all sparse $\gp$ methods spread across the range of observed data $\Xb$.
While inducing-points methods SVGP and PPGPR learn
$\{\XbZ, \Ex{q}{\fbZ}=\mb_M\}$ pairs,
CVGP learns pseudo-points $\{\XbC, \ybC\}$ with pseudo-observations $\ybC$ in the observation space $\mathcal{Y}$.
Hence, CVGP can learn pseudo-observations $\ybC$ that are correlated
with observed data $\yb$.
Notice how, in Figure \ref{fig:exp_coresets_histograms},
CVGP's posterior is based on coreset pseudo-outputs that are 
far from the $\gp$ latent values $\fb$ in the $x \in (1, 2)$ range,
which are up-weighted (\ie green colored dots),
where the observations are subject to heteroskedasticity.

Figure~\ref{fig:exp_coresets_histograms} shows CVGP’s learned histograms of $\betabC$,
where we compare CVGP with k-means and random initializations (RandomCVGP),
illustrating its ability to up- and down-weight pseudo-input/output pairs.
RandomCVGP drives many $\beta_m$ to 0 for uninformative data regions,
effectively ignoring those pseudo-points, while up-weighting more informative ones
---recall that, in coreset-based posteriors, $\beta_m \geq 0$ corresponds to drawing $\beta_m$ samples for each pseudo-point $\{\Xb_m, \yb_m\}$, improving posterior efficiency.

We argue that it is this coreset-based distribution
that enables CVGP's efficient and accurate approximation of $\gp$ posteriors at a lower parameter complexity:
\ie better predictive posterior, based on fewer pseudo-points $M$.

Additional benefits of CVGP coreset-based posteriors,
namely posterior explainability
and compact, informative representations of datasets
are illustrated in Appendix Section~\ref{asssec:app_exp_coresets}.

\section{CONCLUSION}
\label{sec:conclusions}
We introduced CVGP,
the first scalable $\gp$ inference method that
leverages a coreset-based, variational posterior
for accurate and scalable $\gp$ inference.

CVGP enables stochastic optimization of
its variational lower-bound to the $\gp$'s marginal log-likelihood,
after marginalization of latent $\gp$ variables,
at reduced $\bigO{M}$ parameter complexity
---with $\bigO{M^3}$ time- and $\bigO{M^2}$ space-requirements.
Experimental results demonstrate that CVGP provides improved inference and predictive capabilities,
outperforming stochastic variational inference-based alternatives.

CVGP provides a high-quality $\gp$ posterior approximation that
effectively interpolates between the $\gp$ prior and the data likelihood,
and is learned via an efficient lower-dimensional stochastic optimization problem
that results in CVGP achieving a tighter lower-bound than stochastic variational alternatives.

Overall, CVGP's coreset-based posterior
accurately approximates the true $\gp$ posterior,
providing a sparse and explainable representation of the $\gp$ posterior,
with added flexibility to adjust (and discard) pseudo-input/output pairs.

Building upon CVGP's formulation for $\gp$ regression,
we envision follow-up work
with other data-likelihoods (\eg $\gp$-based classification),
and methods leveraging the data-compression benefits of CVGP's coreset-based posterior.

\section{ACKNOWLEDGEMENTS} 

Mert Ketenci acknowledges
this research is supported by NHLBI award R01HL148248.

I\~{n}igo Urteaga acknowledges
this research is supported by ``la Caixa'' foundation fellowship LCF/BQ/PI22/11910028,
and also by the Basque Government through the BERC 2022-2025 program
and by the Ministry of Science and Innovation: BCAM Severo Ochoa accreditation
CEX2021-001142-S / MICIN / AEI / 10.13039/501100011033.

\bibliography{main}
\bibliographystyle{plainnat}

\clearpage

\appendix
\onecolumn


\section{CVGP DERIVATION DETAILS}
\label{asec:cvtgp_derivation}

We derive CVGP's coreset-based posterior
and the log-marginal likelihood's variational lower-bound,
first from the function-space view of $\gp$s in Section~\ref{assec:cvtgp_functionspace},
and then from the complementary weight-space view in Section~\ref{assec:cvtgp_weightspace}.
Independently of the route taken, the attained variational lower-bounds are identical, 
yet the weight- and function-space coreset-based variational posteriors
enable complementary understanding of CVGP's inference and posterior explanations.

\vspace{-1ex}
\subsection{FUNCTION-SPACE DERIVATION OF CVGP}
\label{assec:cvtgp_functionspace}

We derive below,
under the assumption of standard Gaussian, uncorrelated observation noise,
\ie $\yb = \fb + \epsilon\;, \epsilon \sim \N{\epsilon\mid \zerob, \sigma^2 \Ib_N}$,
the coreset-based tempered posterior over $\gp$ coreset function values $q(\fbC)$.
The derivations are equivalent for non-zero mean and/or correlated noise functions.

\subsubsection{CVGP's Coreset-based Posterior}
\label{assec:cvtgp_qf_analytical}

To be able to accurately approximate the full $\gp$ posterior with a coreset $\{\XbC, \ybC \}$,
we propose to weight with $\beta_c \geq 0$ \footnote{
In practice, we ensure positive $\beta_m$ using the $\mathrm{softplus}(.)$ function.
}
the likelihood of each psuedo-point when computing their corresponding coreset $\gp$ value $\fbC$, \ie 

\begin{align}
	\cq{\fbC}{\XbC, \ybC, \betabC} &= \frac {\cq{\ybC}{\fbC, \betabC} \cp{\fbC}{\XbC}}{\cq{\ybC}{\XbC, \betabC} } 
		= \frac {\left(\prod_{m=1}^M \cp{y_{m}}{f_{m}}^{\beta_{m}} \right) \cp{\fbC}{\XbC}}{\cq{\ybC }{\XbC, \betabC} } \; .
\end{align}

We start by deriving a closed form expression for the $\betabC$-weighted likelihood function $q(\ybC | \fbC, \betabC)$,
by considering each coreset pair $\{x_{m},y_{m}\}$, for $m = 1, \cdots, M,$ independently,

\begin{align}
	\cq{y_m}{f_m, \beta_m} &= \cp{y_{m}}{f_{m}}^{\beta_{m}} = \N{y_{m} \mid f_{m}, \sigma^2}^{\beta_{m}} \\
	&= \left(\frac{1}{\sqrt{2 \pi \sigma^2}} e^{- \frac{1}{2} (y_{m} - f_{m}) \sigma^{-2} (y_{m} - f_{m})}\right)^{\beta_{m}} \\
	&= \left(\frac{1}{\sqrt{2 \pi \sigma^2}}\right)^{\beta_{m}} e^{- \frac{1}{2} (y_{m} - f_{m}) (\beta_{m}^{-1}\sigma^2)^{-1} (y_{m} - f_{m})} \\
	&= \left(\frac{1}{\sqrt{2 \pi \sigma^2}}\right)^{\beta_{m}}
	\left(\frac{\sqrt{2 \pi \beta_{m}^{-1}\sigma^2}}{\sqrt{2 \pi \beta_{m}^{-1}\sigma^2}}\right) \exp\left\{- \frac{1}{2} (y_{m} - f_{m}) (\beta_{m}^{-1}\sigma^2)^{-1} (y_{m} - f_{m})\right\} \\
        &= \frac{\sqrt{2 \pi \beta_{m}^{-1}\sigma^2}}{\left(\sqrt{2 \pi \sigma^2}\right)^{\beta_{m}}}
	\left(\frac{1}{\sqrt{2 \pi \beta_{m}^{-1}\sigma^2}}\right) \exp\left\{- \frac{1}{2} (y_{m} - f_{m}) (\beta_{m}^{-1}\sigma^2)^{-1} (y_{m} - f_{m})\right\} \\
        &= \frac{\sqrt{2 \pi \beta_{m}^{-1}\sigma^2}}{\left(\sqrt{2 \pi \sigma^2}\right)^{\beta_{m}}} \cdot \N{y_m|f_m, \beta_{m}^{-1}\sigma^2} \\
        &= Q_c \cdot \N{y_m|f_m, \beta_{m}^{-1}\sigma^2} \;, 
            \text{ with } Q_c = \frac{\sqrt{2 \pi \beta_{m}^{-1}\sigma^2}}{\left(\sqrt{2 \pi \sigma^2}\right)^{\beta_{m}}} \;.
\end{align}

We write the joint over the full coreset pseudo-observations as a product over each likelihood term:
\begin{align}
	&q(\ybC | \fbC, \betabC) = \prod_{m=1}^M p(y_{m} | f_{m})^{\beta_{m}} \\
	& \qquad = \prod_{m=1}^M\left(\frac{1}{\sqrt{2 \pi \sigma^2}}\right)^{\beta_{m}}
	\left(\frac{\sqrt{2 \pi \beta_{m}^{-1}\sigma^2}}{\sqrt{2 \pi \beta_{m}^{-1}\sigma^2}}\right) \exp\left\{- \frac{1}{2} (y_{m} - f_{m}) (\beta_{m}^{-1}\sigma^2)^{-1} (y_{m} - f_{m})\right\} \\
        & \qquad = \prod_{m=1}^M Q_m \cdot \N{y_m|f_m, \beta_{m}^{-1}\sigma^2} \\
        & \qquad = Q_M \cdot \N{\ybC \mid \fbC, \Sigma_{\betabC}} \;, \text{ with }
        \begin{cases}
            Q_M = \prod_{m=1}^M\frac{\sqrt{2 \pi \beta_{m}^{-1}\sigma^2}}{\left(\sqrt{2 \pi \sigma^2}\right)^{\beta_{m}}}  \\
            \Sigma_{\betabC} = \sigma^2 \cdot \diag{\betabC^{-1}} 
        \end{cases} \;.
        \label{eq:coreset_likelihood}
\end{align}

We derive the marginalized pseudo-observation coreset distribution
\begin{align}
	q(\ybC|\XbC, \betabC) &= \int_{\fbC} q(\ybC, \fbC | \XbC, \betabC) \diff{\fbC} = \int_{\fbC} q(\ybC | \fbC, \betabC) p(\fbC|\XbC) \diff{\fbC} \\
        & = \int_{\fbC} Q_M \cdot \N{\ybC \mid \fbC, \Sigma_{\betabC}} \cdot \N{\fbC \mid \zerob, \KbCC}  \diff{\fbC} \\
        & = Q_M \int_{\fbC} \N{\ybC \mid \fbC, \Sigma_{\betabC}} \cdot \N{\fbC \mid \zerob, \KbCC}  \diff{\fbC} \\
        & = Q_M \cdot \N{\ybC \mid \zerob, \KbCC + \Sigma_{\betabC}} \;.
	\label{eq:coreset_marginal_likelihood}
\end{align}

We leverage the above distributions to derive the coreset-based, tempered $\gp$ posterior
\begin{align}
	& q(\fbC|\XbC, \ybC, \betabC) = \frac {q(\ybC | \fbC, \betabC) q(\fbC|\XbC)}{q(\ybC | \XbC, \betabC) } 	\\
	& \qquad = \frac{
		Q_M \cdot \N{\ybC \mid \fbC, \Sigma_{\betabC}} 
		 \N{\fbC \mid \zerob, \KbCC}
	}{
		Q_M \cdot \N{\ybC \mid \zerob, \KbCC + \Sigma_{\betabC}} 
	} \\
        & \qquad = \frac{
		\N{\ybC \mid \fbC, \Sigma_{\betabC}} 
		 \N{\fbC \mid \zerob, \KbCC}
	}{
		\N{\ybC \mid \zerob, \KbCC + \Sigma_{\betabC}} 
	} \\
        &\qquad = \N{\fbC \mid \mb_{\fbC|\ybC}, \Kb_{\fbC|\ybC}} \;, \text{ with} \begin{cases}
		\mb_{\fbC|\ybC} = \Kb_{\fbC|\ybC} \left(
		\Sigma_{\betabC}^{-1} \ybC 
		\right) \\
		\Kb_{\fbC|\ybC} = \left( 
		\KbCC^{-1} + \Sigma_{\betabC}^{-1}
		\right)^{-1}
	\end{cases}  \hspace*{-2ex} .
	\label{eq:coreset_posterior_fc}
\end{align}

The sufficient statistics of the coreset-based, tempered distibution above can be rewritten as
\begin{align}
    q(\fbC) = q(\fbC|\XbC, \ybC, \betabC) &= \N{\fbC \mid \mb_{\fbC|\ybC}, \Kb_{\fbC|\ybC}} \;, \\
    &\text{ with}
     \begin{cases}
		\mb_{\fbC|\ybC} = \KbCC \left( \KbCC + \Sigma_{\betabC} \right)^{-1} \ybC \\
		\Kb_{\fbC|\ybC} = \KbCC - \KbCC \left( \KbCC + \Sigma_{\betabC} \right)^{-1} \KbCC  \\
                \qquad \qquad \text{by \href{https://en.wikipedia.org/wiki/Woodbury_matrix_identity}{Woodbury matrix identity}}  \; .
	\end{cases} 
\end{align}

\paragraph{The coreset-based posterior over $\gp$ function values.}
We now compute the posterior over $\gp$ values for any given data point $\Xb$,
by marginalizing the $\gp$'s prior-conditional over the coreset-based distribution, \ie

\begin{align}
    q(\fb|\XbC, \ybC) &= \int_{\fbC} p(\fb|\fbC) q(\fbC | \XbC, \ybC, \betabC) \diff{\fbC} \;.
\end{align}

The above is analytically solvable due to all the distributions being Gaussian:
\begin{align}
	q(\fbC|\XbC, \ybC, \betabC) &= \N{\fbC \mid \mb_{\fbC|\ybC}, \Kb_{\fbC|\ybC}} \;, \\
        & \text{ with} \begin{cases}
		\mb_{\fbC|\ybC} = \Kb_{\fbC|\ybC} \left( \Sigma_{\betabC}^{-1} \ybC \right) \\
		\Kb_{\fbC|\ybC} = \KbCC - \KbCC \left(\KbCC + \Sigma_{\betabC}\right)^{-1} \KbCC 
	\end{cases} \hspace*{-2ex},
	\\
	p(\fb|\fbC) &= \N{\fb \mid \mb_{\fb|\fbC}, \Kb_{\fb|\fbC}} \;, \\
        & \text{ with} \begin{cases}
		\mb_{\fb|\fbC} = \KbXC\KbCC^{-1} \fbC \\
		\Kb_{\fb|\fbC} = \KbXX - \KbXC\KbCC^{-1} \KbCX
	\end{cases} \hspace*{-2ex},
	\\
	q(\fb) = q(\fb|\XbC, \ybC, \betabC) &= \N{\fb \mid \mb_{\fb|\ybC}, \Kb_{\fb|\ybC}} \;, \\
        & \text{ with} \begin{cases}
		\mb_{\fb|\ybC} = \KbXC\KbCC^{-1} \mb_{\fbC|\ybC} \\
		\Kb_{\fb|\ybC} = \Kb_{\fb|\fbC} + \KbXC\KbCC^{-1} \Kb_{\fbC|\ybC} \KbCC^{-1} \KbCX 
	\end{cases} \hspace*{-2ex}.
    \label{eq:coreset_posterior_f}
\end{align}

We elaborate on the sufficient statistics of $q(\fb|\XbC, \ybC, \betabC)$.

First, we rewrite the expected value as
\begin{align}
	\mb_{\fb|\ybC} &= \KbXC\KbCC^{-1} \mb_{\fbC|\ybC} \\
	&= \KbXC\KbCC^{-1} \Kb_{\fbC|\ybC} \Sigma_{\betabC}^{-1} \ybC \\
	&= \KbXC\KbCC^{-1} \left( \KbCC^{-1} + \Sigma_{\betabC}^{-1} \right)^{-1} \Sigma_{\betabC}^{-1} \ybC \\
		& \qquad \text{ by using equivalence in Equation~\eqref{eq:equiv_1}} \nonumber \\
	&= \KbXC\left( \KbCC + \Sigma_{\betabC} \right)^{-1} \ybC  \; ,
    \label{eq:coreset_posterior_f_mean}
\end{align} 

where we have made use of the following equivalences,
\begin{align}
    & \KbCC^{-1} \left( \KbCC^{-1} + \Sigma_{\betabC}^{-1} \right)^{-1} \Sigma_{\betabC}^{-1} = \KbCC^{-1} \left( \Sigma_{\betabC} \KbCC^{-1} + \Ib_M \right)^{-1} 
            = \left( \Sigma_{\betabC} + \KbCC \right)^{-1}
        \label{eq:equiv_1} \;, \\
    & \Sigma_{\betabC}^{-1} \left( \KbCC^{-1} + \Sigma_{\betabC}^{-1} \right)^{-1} \KbCC^{-1} = \Sigma_{\betabC}^{-1} \left( \Ib_M + \KbCC\Sigma_{\betabC}^{-1}  \right)^{-1} 
	= \left( \Sigma_{\betabC} + \KbCC \right)^{-1}
		\label{eq:equiv_2} \;.
\end{align}
Second, for the covariance matrix, we write
\begin{align}
	\Kb_{\fb|\ybC} &= \Kb_{\fb|\fbC} + \KbXC\KbCC^{-1} \Kb_{\fbC|\ybC} \KbCC^{-1} \KbCX\\
	&= \KbXX - \KbXC\KbCC^{-1} \KbCX + \KbXC\KbCC^{-1} \left( \KbCC^{-1} + \Sigma_{\betabC}^{-1} \right)^{-1} \KbCC^{-1} \KbCX\\
	& \qquad \qquad \text{ by using the \href{https://en.wikipedia.org/wiki/Woodbury_matrix_identity}{Woodbury matrix identity} for } \left( \KbCC^{-1} + \Sigma_{\betabC}^{-1} \right)^{-1} \nonumber \\
	&= \KbXX - \KbXC\KbCC^{-1} \KbCX + \KbXC\KbCC^{-1} \left( \KbCC - \KbCC \left(\KbCC + \Sigma_{\betabC}\right)^{-1} \KbCC  \right) \KbCC^{-1} \KbCX\\
	&= \KbXX - \KbXC\KbCC^{-1} \KbCX + \KbXC\left( \Ib_M - \left(\KbCC + \Sigma_{\betabC}\right)^{-1} \KbCC  \right) \KbCC^{-1} \KbCX\\	
	&= \KbXX - \KbXC\KbCC^{-1} \KbCX + \KbXC\KbCC^{-1} \KbCX - \KbXC \left(\KbCC + \Sigma_{\betabC}\right)^{-1} \KbCC \KbCC^{-1} \KbCX\\
	&= \KbXX - \KbXC \left(\KbCC + \Sigma_{\betabC}\right)^{-1} \KbCX  \; .
    \label{eq:coreset_posterior_f_cov}
\end{align}

\clearpage
\subsubsection{CVGP's Variational Lower-bound}
\label{assec:cvtgp_qf_lowerbound}
We derive the variational lower-bound by writing everything in terms of sufficient statistics of $\q{\fbC}$:
\begin{align}
	\Loss_{CVGP} & = \eValue{q(\fb)}{\log p(\yb| \fb) }  - \kl{q(\fbC)}{p(\fbC)}  \\
	&= \log \N{\yb | \mb_{\fb|\ybC}, \sigma^2 \Ib_N} - \frac{1}{2 \sigma^{2}} \tr{\Kb_{\fb|\ybC} } \nonumber \\
	& \qquad - \frac{1}{2} \left(
		\tr{\KbCC^{-1}\Kb_{\fbC|\ybC}} - M + \mb_{\fbC|\ybC}^\top \KbCC^{-1} \mb_{\fbC|\ybC} + \log \frac{\left|\KbCC\right|}{\left|\Kb_{\fbC|\ybC}\right|} 
	 	\right) \\
	&= \log \N{\yb | \KbXC \KbCC^{-1} \mb_{\fbC|\ybC} , \sigma^2 \Ib_N} \nonumber \\
	& \qquad - \frac{1}{2 \sigma^{2}} \tr{\KbXX - \KbXC \KbCC^{-1} \KbCX + \KbXC \KbCC^{-1}\Kb_{\fbC|\ybC} \KbCC^{-1} \KbCX} \nonumber \\
	& \qquad - \frac{1}{2} \left(
	\tr{\KbCC^{-1}\Kb_{\fbC|\ybC}} - M + \mb_{\fbC|\ybC}^\top \KbCC^{-1} \mb_{\fbC|\ybC} + \log \frac{\left|\KbCC\right|}{\left|\Kb_{\fbC|\ybC}\right|} 
	\right) \\
	&= \left( -\frac{N}{2} \log (2\pi) - \frac{1}{2} \log\left|\sigma^2 \Ib_N \right| -\frac{1}{2} (\yb-\KbXC \KbCC^{-1} \mb_{\fbC|\ybC})^\top \sigma^{-2} \Ib_N (\yb-\KbXC \KbCC^{-1} \mb_{\fbC|\ybC}) \right) \nonumber \\
	& \qquad - \frac{1}{2 \sigma^{2}} \tr{\KbXX - \KbXC \KbCC^{-1} \KbCX}
		- \frac{1}{2 \sigma^{2}} \tr{\KbXC \KbCC^{-1}\Kb_{\fbC|\ybC} \KbCC^{-1} \KbCX} \nonumber \\
	& \qquad - \frac{1}{2} \left(
	\tr{\KbCC^{-1}\Kb_{\fbC|\ybC}} - M + \mb_{\fbC|\ybC}^\top \KbCC^{-1} \mb_{\fbC|\ybC} + \log \frac{\left|\KbCC\right|}{\left|\Kb_{\fbC|\ybC}\right|} 
	\right) \\
	&= -\frac{N}{2} \log (2\pi) + \frac{M}{2} - \frac{1}{2} \log\left|\sigma^2 \Ib_N \right| - \frac{1}{2} \log \left|\KbCC\right| \nonumber \\
	& -\frac{1}{2} \yb^\top \sigma^{-2} \yb 
		+ \sigma^{-2} \mb_{\fbC|\ybC}^\top \KbCC^{-1}  \KbCX \yb \\
		&-\frac{1}{2} \sigma^{-2} \mb_{\fbC|\ybC}^\top \KbCC^{-1}  \KbCX \KbXC \KbCC^{-1} \mb_{\fbC|\ybC}
		- \frac{1}{2} \mb_{\fbC|\ybC}^\top \KbCC^{-1} \mb_{\fbC|\ybC} \nonumber \\
	& - \frac{1}{2 \sigma^{2}} \tr{\KbXX - \KbXC \KbCC^{-1} \KbCX} \nonumber \\
	& - \frac{1}{2 \sigma^{2}} \tr{ \KbCC^{-1} \KbCX \KbXC \KbCC^{-1}\Kb_{\fbC|\ybC} } 
		- \frac{1}{2} \tr{\KbCC^{-1}\Kb_{\fbC|\ybC}} + \frac{1}{2} \log \left|\Kb_{\fbC|\ybC} \right|\\
	&= -\frac{N}{2} \log (2\pi) + \frac{M}{2} - \frac{1}{2} \log\left|\sigma^2 \Ib_N\right| - \frac{1}{2} \log \KbCC -\frac{1}{2} \yb^\top \sigma^{-2} \yb  \nonumber \\
	& + \sigma^{-2} \mb_{\fbC|\ybC}^\top \KbCC^{-1}  \KbCX \yb 
	-\frac{1}{2} \mb_{\fbC|\ybC}^\top \left(  \KbCC^{-1} + \sigma^{-2}  \KbCC^{-1}  \KbCX \KbXC \KbCC^{-1} \right)\mb_{\fbC|\ybC} \nonumber \\
	& - \frac{1}{2 \sigma^{2}} \tr{\KbXX - \KbXC \KbCC^{-1} \KbCX} \nonumber \\
	& - \frac{1}{2 } \tr{ \left(\KbCC^{-1} + \sigma^{-2}\KbCC^{-1} \KbCX \KbXC \KbCC^{-1} \right)\Kb_{\fbC|\ybC} } 
		+ \frac{1}{2} \log \left|\Kb_{\fbC|\ybC} \right| \\
	&= -\frac{N}{2} \log (2\pi) + \frac{M}{2} - \frac{1}{2} \log\left|\sigma^2 \Ib_N \right| - \frac{1}{2} \log \KbCC -\frac{1}{2} \yb^\top \sigma^{-2} \yb  \nonumber \\
	& + \sigma^{-2} \mb_{\fbC|\ybC}^\top \KbCC^{-1}  \KbCX \yb 
	-\frac{1}{2} \mb_{\fbC|\ybC}^\top \KbCC^{-1} \left(\KbCC^{-1} - \left(\KbCC - \Sigma_{\betabC}\right)^{-1} \right)^{-1} \KbCC^{-1}  \mb_{\fbC|\ybC} \nonumber \\
	& - \frac{1}{2 \sigma^{2}} \tr{\KbXX - \KbXC \KbCC^{-1} \KbCX} \nonumber \\
	& - \frac{1}{2 } \tr{ \KbCC^{-1} \left(\KbCC^{-1} - \left(\KbCC - \Sigma_{\betabC}\right)^{-1} \right)^{-1} \KbCC^{-1}\Kb_{\fbC|\ybC} } 
	+ \frac{1}{2} \log \left|\Kb_{\fbC|\ybC} \right|\\
    &= \log \N{\yb | \mb_{\fb|\ybC}, \sigma^2 \Ib_N} - \frac{1}{2 \sigma^2}\tr{\Kb_{\fb|\ybC} } \nonumber\\
    & - \frac {1}{2} \left[ - \tr{\Ab \KbCC}  + \ybC^\top
        \Ab \KbCC \Ab \ybC  - \ln \left|\Ab\right| - \ln \left|\Sigmab_{\betabC} \right|  \right] \:,
	\label{eq:cvtgp_loss_data_marginal_suffstats}
\end{align}

where $\Ab = \left(\KbCC + \Sigmab_{\betabC}\right)^{-1}$.

\newpage
\clearpage

\clearpage
\subsection{WEIGHT-SPACE DERIVATION OF CVGP}
\label{assec:cvtgp_weightspace}

For completeness and a complementary perspective, we derive CVGP inference from the weight-space view of $\gp$s,
again under the assumption of standard Gaussian, uncorrelated observation noise,
\ie $\yb = \fb + \epsilon\;, \epsilon \sim \N{\epsilon \mid \zerob, \sigma^2 \Ib_N}$.

Recall the weight-space definition of $\gp$s,~\citep{rasmussen2006gaussian}:
\begin{align}
       y_i \mid \wb, \xb_i \sim \N{\yb \mid \Phi(\xb_i)^\top\wb, \sigma^{2}\Ib_N} \; \text{ with } \wb \sim \N{\wb \mid \zerob, \Ib_D} \;, 
\end{align}
where $\map{\cdot}: \xspace \rightarrow \hilbert$ is a feature map with associated kernel $k(\cdot,\cdot): \xspace \times \xspace \rightarrow \RR$ and Hilbert space $\hilbert$.
Namely, a $\gp$ can be viewed as a Bayesian linear regression where the covariates $\xspace$ are embedded into a potentially infinite dimensional Hilbert space $\hilbert$.
An advantage of the weight-space view is that it allows for conditional independence between different data points $\xb$ given $\wb$; the key property we leverage in the following derivations. 

\subsubsection{CVGP's coreset-based tempered-posterior}
\label{assec:cvtgp_qw_analytical}

We aim for a small subset of psuedo-points $\{\XbC, \ybC\}$ that, 
if drawn $\beta_m \geq 0$ times, approximate the true weight posterior:
\begin{align}
    \underbrace{\frac{\p{\wb} \prod_{m=1}^M  \cp{y_c}{\wb, \xb_m}^{\beta_m}}{Z_q} }_{\cq{\wb}{\ybC, \XbC,\betabC}} \approx \underbrace{\frac{\p{\wb} \prod_{i=1}^N  \cp{y_i}{\wb, \xb_i}}{Z_p}}_{\cp{\wb}{ \yb, \Xb}} \;.
\end{align}

Typically, the objective of the coreset problem is to learn vector $\beta^\star = \argmin_{\beta} \Dist{q(\wb), p(\wb)}$,
where $\Dist{.}$ is a distance metric such as the KL divergence~\citep{campbell2019sparse}. 

We derive the coreset-based tempered posterior over $\gp$ weights $\cq{\wb}{\XbC, \ybC, \betabC}$, by noting it is proportional to 
 \begin{align}
    & \expp{-\frac{1}{2} \wb^\top \wb} \prod_{m=1}^M \expp{ -\frac{1}{2} \sigma^{-2} \beta_m  \left(y_c - \map{\xb_m}^\top \wb \right)^2}\\
    &\propto \expp{ -\frac{1}{2} \left( - 2 \sigma^{-2}  \wb^\top \sum_{m=1}^M y_c \beta_m \map{\xb_m}  +   \wb^\top \underbrace{\left(\sigma^{-2} \sum_{m=1}^M \map{\xb_m}\beta_m  \map{\xb_m}^\top + \Ib_D\right)}_{\Sb_{\wb|\ybC}^{-1}}\wb\right)}, 
    \label{eq:coreset_posterior_w_propto}
\end{align}

which is a Gaussian distribution with covariance matrix: 
\begin{align}
    \Sb_{\wb|\ybC} &= \left(\sigma^{-2} \sum_{m=1}^M \map{\xb_m}\beta_m  \map{\xb_m}^\top + \Ib_D\right)^{-1} = \left(\map{\XbC}^\top \Sigma_{\betabC}^{-1}  \map{\XbC} + \Ib_D\right)^{-1} \;,
\end{align}
with $\Sigma_{\betabC} = \sigma^2 \cdot \diag{\betabC^{-1}}$.
Let us define $\Sigma_{\betabC}^{-1} = C^{1/2}C^{1/2}$, 
with $\map{\XbC}^\top C^{1/2} = {\map{\XbC}^\prime}^\top$ and $C^{1/2} \map{\XbC} = {\map{\XbC}^\prime}$.
Then we can write the covariance matrix as
\begin{align}
    \Sb_{\wb|\ybC} &= \left( {\map{\XbC}^\prime}^\top{\map{\XbC}^\prime}+ \Ib_D\right)^{-1}\\
    &= \Ib_D - {\map{\XbC}^\prime}^\top\left( \Ib_D + {\map{\XbC}^\prime}{\map{\XbC}^\prime}^\top\right)^{-1}{\map{\XbC}^\prime}\\
    &=\Ib_D - \map{\XbC}^\top\left( \Sigma_{\betabC} + \KbCC\right)^{-1} \map{\XbC} \;.
\end{align}

We revisit Equation~\eqref{eq:coreset_posterior_w_propto} to identify the mean of the Gaussian distribution as follows,
where we use ${\ybC}^\prime=C^{1/2} \ybC$:
\begin{align}
    \mb_{\wb|\ybC} &= \Sb_{\wb|\ybC} \left( \sigma^{-2} \sum_{m=1}^M y_c \beta_m \map{\xb_m} \right)\\
    &= \Sb_{\wb|\ybC} \left( \map{\XbC}^\top \Sigma_{\betabC}^{-1} \ybC\right)\\
    &= \Sb_{\wb|\ybC} \left( \map{\XbC}^\top C^{1/2}C^{1/2} \ybC\right)\\
    &= \Sb_{\wb|\ybC} \left( {\map{\XbC}^\prime}^\top {\ybC}^\prime\right) \\
    &= \left(\Ib_D - {\map{\XbC}^\prime}^\top\left( \Ib_D + {\map{\XbC}^\prime}{\map{\XbC}^\prime}^\top\right)^{-1}{\map{\XbC}^\prime}\right)\left( {\map{\XbC}^\prime}^\top {\ybC}^\prime\right)\\
    &= \left({\map{\XbC}^\prime}^\top - {\map{\XbC}^\prime}^\top\left( \Ib_D + {\map{\XbC}^\prime}{\map{\XbC}^\prime}^\top\right)^{-1}{\map{\XbC}^\prime} {\map{\XbC}^\prime}^\top \right) {\ybC}^\prime\\
    &= \left({\map{\XbC}^\prime}^\top - {\map{\XbC}^\prime}^\top\left( \left({\map{\XbC}^\prime} {\map{\XbC}^\prime}^\top \right)^{-1} + \Ib_D\right)^{-1}\right) {\ybC}^\prime\\
    &= {\map{\XbC}^\prime}^\top\left(\Ib_D - \left( \left({\map{\XbC}^\prime} {\map{\XbC}^\prime}^\top \right)^{-1} + \Ib_D\right)^{-1}\right) {\ybC}^\prime\\
    &= {\map{\XbC}^\prime}^\top\left(\Ib_D - {\map{\XbC}^\prime} {\map{\XbC}^\prime}^\top \left(\underbrace{\Ib_D + {\map{\XbC}^\prime}{\map{\XbC}^\prime}^\top}_{D}\right)^{-1}\right) {\ybC}^\prime\\
    &={\map{\XbC}^\prime}^\top\left(DD^{-1} - {\map{\XbC}^\prime} {\map{\XbC}^\prime}^\top D^{-1}\right) {\ybC}^\prime\\
    &= {\map{\XbC}^\prime}^\top\left(\bcancel{D} - \bcancel{{\map{\XbC}^\prime} {\map{\XbC}^\prime}^\top} \right) D^{-1} {\ybC}^\prime\\
    &=  {\map{\XbC}^\prime}^\top D^{-1} {\ybC}^\prime\\
    &= \map{\XbC}^\top C^{1/2}\left( \Ib_D + C^{1/2}  \map{\XbC} \map{\XbC}^\top C^{1/2}\right)^{-1} C^{1/2} \ybC\\
    &= \map{\XbC}^\top\left( \Sigma_{\betabC} +  \KbCC \right)^{-1}\ybC \;.
\end{align}

All in all, we have

\begin{align}
q(\wb|\XbC, \ybC, \betabC) &= \N{\wb | \mb_{\wb|\ybC}, \Sb_{\wb|\ybC}} \;, \label{eq:coreset_posterior_wc} 
        \text{ with} \begin{cases}
		\mb_{\wb|\ybC} = \map{\XbC}^\top\left( \Sigma_{\betabC} +  \KbCC \right)^{-1}\ybC \\
		\Sb_{\wb|\ybC} = \Ib_D - \map{\XbC}^\top\left( \Sigma_{\betabC} + \KbCC\right)^{-1} \map{\XbC} 
	\end{cases}  \hspace*{-2ex} .
\end{align}

\subsubsection{CVGP's Weight-space Variational Lower-bound}
\label{assec:cvtgp_qw_lowerbound}

We write the variational lower-bound of the log-marginal likelihood as:
\begin{align}
     \log\cp{\yb}{\Xb} &= \int \q{\wb} \log \left\{ \frac{\cp{\yb, \wb}{\Xb}\q{\wb}}{\cp{\wb}{\Xb, \yb}\q{\wb}}  \right\}\diff{\wb}\\
    &= \Ex{\q{\wb}}{\log \cp{\yb}{\wb, \Xb}} - \kl{\q{\wb}}{\p{\wb}}  + \kl{\q{\wb}}{\cp{\wb}{\Xb, \yb}}\\
    &\geq \underbrace{\Ex{\q{\wb}}{\log \cp{\yb}{\wb, \Xb}} - \kl{\q{\wb}}{\p{\wb}}}_{\mathcal{L}_{CVGP}} \; ,
\end{align}

which is the lower-bound of the weight-space view of CVGP,
where we set $\q{\wb} = \cp{\wb}{\XbC, \ybC, \betabC}$ and derive
\begin{align}
    \Loss_{CVGP} &= \Ex{\cp{\wb}{\XbC, \ybC, \betabC}}{\log \cp{\yb}{\wb, \Xb}} - \kl{\cp{\wb}{\XbC, \ybC, \betabC}}{\p{\wb}}\\
    &= \Ex{\cp{\wb}{\XbC, \ybC, \betabC}}{\sum_{i=1}^N\log \cp{y_i}{\wb, \xb_i}} - \kl{\cp{\wb}{\XbC, \ybC, \betabC}}{\p{\wb}}\\
    &= \sum_{i=1}^N\underbrace{\Ex{\cp{\wb}{\XbC, \ybC, \betabC}}{\log \cp{y_i}{\wb, \xb_i}}}_{\ell_i} - \kl{\cp{\wb}{\XbC, \ybC, \betabC}}{\p{\wb}} .
\end{align}

We compute below the analytical expressions for $\ell_i$ and $\kl{\cp{\yb}{\wb, \Xb}}{\p{\wb}}$.

We start with $\ell_i$:
\begin{align}
    &\ell_i = \int  \cp{\wb}{\XbC, \ybC, \betabC}  \log\frac{1}{\sqrt{2 \pi\sigma^2}} \expp{ - \frac{1}{2\sigma^2}\left(y_i - \map{\xb_i}^\top\wb\right)^2 }  \diff{\wb}\\
    &= -\frac{1}{2} \left( \log 2 \pi\sigma^2  + \sigma^{-2} \int  \cp{\wb}{\XbC, \ybC, \betabC}  \left( y_i^2 - 2 y_i \map{\xb_i}^\top \wb  + \map{\xb_i}^\top \wb\wb^\top \map{\xb_i}\right) \diff{\wb} \right) \;
\end{align}

We need to compute $\int\wb\cp{\wb}{\XbC, \ybC, \betabC}\diff{\wb}$ and $\int\wb\wb^\top\cp{\wb}{\XbC, \ybC, \betabC}\diff{\wb}$.
Note that the former is $\mb_{\wb|\ybC}$
and the latter is $\Sb_{\wb|\ybC} + \mb_{\wb|\ybC}\mb_{\wb|\ybC}^\top$.
Hence,
\begin{align}
    \ell_i =-\frac{1}{2} \left( \log 2 \pi\sigma^2  + \sigma^{-2}\left( y_i^2 - 2 y_i \map{\xb_i}^\top \mb_{\wb|\ybC}  + \map{\xb_i}^\top (\Sb_{\wb|\ybC} + \mb_{\wb|\ybC} \mb_{\wb|\ybC}^\top) \map{\xb_i}\right)\right) \;.
\end{align}

Let us now define
\begin{align}
    m_{f_i|\ybC} &= \map{\xb_i}^\top \mb_{\wb|\ybC}\\
    &= \map{\xb_i}^\top \map{\XbC}^\top\left( \Sigma_{\betabC} +  \KbCC \right)^{-1}\ybC\\
    &= \kb_{iM}\left( \Sigma_{\betabC} +  \KbCC \right)^{-1}\ybC \;,
\end{align}

and

\begin{align}
    k_{f_i|\ybC} &= \map{\xb_i}^\top \Sb_{\wb|\ybC} \map{\xb_i}\\
    &= k_{ii} - \kb_{iM}\left( \Sigma_{\betabC} + \KbCC\right)^{-1} \kb_{\XbC, \xb_i} \;,
\end{align}
where the above relate to the $\gp$ function values via transformation of the weights by the feature vectors, \ie $f_i=f(\xb_i)=\map{\xb_i}^\top \wb$.
Notice how the above expressions match those in Equation~\eqref{eq:coreset_posterior_f_mean} and~\eqref{eq:coreset_posterior_f_cov}.
We can therefore write
\begin{align}
    \ell_i &=-\frac{1}{2} \left( \log 2 \pi\sigma^2  + \sigma^{-2}\left( y_i^2 - 2 y_i m_{f_i|\ybC}  + k_{f_i|\ybC} + {m_{f_i|\ybC}}^2 \right)\right)\\
    &= \log \N{y_i \mid m_{f_i|\ybC}, \sigma^2}\exp\left\{- \frac{1}{2}\sigma^{-2} k_{f_i|\ybC} \right\} \;.
\end{align}

We continue with the KL divergence term,
recalling $\p{\wb}=\N{\wb \mid \zerob, \Ib_D}$,
and write
\begin{align}
    \kl{\cp{\wb}{\XbC, \ybC, \betabC}}{\p{\wb}} &= \frac{1}{2} \left( \mb_{\wb|\ybC}^\top \Ib_D^{-1} \mb_{\wb|\ybC} + \tr {\Ib_D^{-1} \Sb_{\wb|\ybC} } +  \log \mid \Ib_D \mid - \log \mid \Sb_{\wb|\ybC} \mid  - \tr{ \Ib_D } \right)\\
    &= \frac{1}{2} \left( \mb_{\wb|\ybC}^\top \mb_{\wb|\ybC} + \tr { \Sb_{\wb|\ybC} } - \log \mid \Sb_{\wb|\ybC} \mid  - \tr { \Ib_D } \right) \;.
\end{align}

We first compute

\begin{align}
    \mb_{\wb|\ybC}^\top \mb_{\wb|\ybC} &=  \left(\map{\XbC}^\top\left( \Sigma_{\betabC} +  \KbCC \right)^{-1}\ybC\right)^\top \left(\map{\XbC}^\top\left( \Sigma_{\betabC} +  \KbCC \right)^{-1}\ybC\right)\\
    &= \ybC^\top \left( \Sigma_{\betabC} +  \KbCC \right)^{-1} \map{\XbC} \map{\XbC}^\top\left( \Sigma_{\betabC} +  \KbCC \right)^{-1}\ybC\\
    &= \ybC^\top \left( \Sigma_{\betabC} +  \KbCC \right)^{-1} \KbCC\left( \Sigma_{\betabC} +  \KbCC \right)^{-1}\ybC \;,
\end{align}

then,

\begin{align}
     &\tr { \Sb_{\wb|\ybC} } = \tr {\Ib_D - \map{\XbC}^\top\left( \Sigma_{\betabC} + \KbCC\right)^{-1} \map{\XbC}}\\
    &= \tr {\Ib_D } -\tr {\map{\XbC}^\top\left( \Sigma_{\betabC} + \KbCC\right)^{-1} \map{\XbC}}\\
    &=  \tr {\Ib_D } -\tr {\left( \Sigma_{\betabC} + \KbCC\right)^{-1} \map{\XbC}\map{\XbC}^\top}\\
    &= \tr {\Ib_D } -\tr {\left( \Sigma_{\betabC} + \KbCC\right)^{-1} \KbCC} \;,
\end{align}

and finally,

\begin{align}
    \log \mid \Sb_{\wb|\ybC} \mid &= \log \left|  \left( {\map{\XbC}^\prime}^\top{\map{\XbC}^\prime}+ \Ib_D\right)^{-1} \right| \\
    &= -\log \left|  \left( {\map{\XbC}^\prime}^\top{\map{\XbC}^\prime}+ \Ib_D\right)\right|\\
    &= -\log \left|  \left( {\map{\XbC}}^\top C{\map{\XbC}}+ \Ib_D\right)\right|\\
    &= -\log \left|\Sigma_{\betabC} + \map{\XbC}{\map{\XbC}}^\top \right| \left| \Sigma_{\betabC}^{-1} \right| \bcancel{\left| \Ib_D \right|}\\
    & \quad\text{ using \href{https://en.wikipedia.org/wiki/Matrix_determinant_lemma}{metrix determinant lemma} } \\
    &= - \log \left|\Sigma_{\betabC} + \KbCC \right| - \log \left| \Sigma_{\betabC}^{-1} \right| \;.
\end{align}

We put it all together for the analytical, weight-space variational lower-bound of CVGP,

\begin{align}
    \Loss_{CVGP} &=\sum_{i=1}^N \left( \log \N{y_i \mid m_{f_i|\ybC}, \sigma^2}\exp\left\{- \frac{1}{2}\sigma^{-2} k_{f_i|\ybC} \right\}\right) \nonumber \\
    & \qquad -\frac{1}{2} \left( +\ybC \left( \Sigma_{\betabC} +  \KbCC \right)^{-1} \KbCC\left( \Sigma_{\betabC} +  \KbCC \right)^{-1}\ybC \right. \nonumber  \\
    & \qquad \qquad + \bcancel{\tr {\Ib_D }} -\tr {\left( \Sigma_{\betabC} + \KbCC\right)^{-1} \KbCC} \nonumber \\
    & \qquad \qquad \left. + \log \left|\Sigma_{\betabC} + \KbCC \right| + \log \left| \Sigma_{\betabC}^{-1} \right| - \bcancel{\tr{\Ib_D}} \right) \\
    &=\sum_{i=1}^N \left(\log \N{y_i \mid m_{f_i|\ybC}, \sigma^2} - \frac{1}{2}\sigma^{-2} k_{f_i|\ybC} \right) \nonumber \\
    & \qquad -\frac{1}{2} \left( -\tr {\left( \KbCC + \Sigma_{\betabC} \right)^{-1} \KbCC} \right. \nonumber  \\
    & \qquad \qquad +\ybC \left( \Sigma_{\betabC} +  \KbCC \right)^{-1} \KbCC\left( \Sigma_{\betabC} +  \KbCC \right)^{-1}\ybC \nonumber \\
    & \qquad \qquad \left. + \log \left|\KbCC + \Sigma_{\betabC} \right| - \log \left| \Sigma_{\betabC} \right| \right)\\
    &=  \log \N{\yb | \mb_{\fb|\ybC}, \sigma^2 \Ib_D} - \frac{1}{2 \sigma^2}\tr{\Kb_{\fb|\ybC} } \nonumber\\
    & - \frac {1}{2} \left[ - \tr{\Ab \KbCC}  + \ybC^\top
        \Ab \KbCC \Ab \ybC  - \ln \left|\Ab\right| - \ln \left|\Sigmab_{\betabC} \right|  \right]\;,
\end{align}
where $\Ab = \left(\Sigmab_{\betabC} + \KbCC \right)^{-1}$ and
we have combined
($a$) the sum over $N$ scalar likelihoods
into a single, multivariate Gaussian with mean $\mb_{\fb|\ybC}$
(composed of $\mb_{f_i|\ybC}, \forall i)$ and diagonal unit covariance; and
($b$) all $\kb_{\fb_i|\ybC}$ terms into a diagonal matrix
${\Kb_{\fb|\ybC}}_{ii} = \kb_{\fb_i|\ybC}$.

\clearpage
\subsection{CVGP’S LOWER-BOUND AND ITS OPTIMUM}
\label{assec:cvgp_lower_bound_optimum}

Before expanding CVGP's lower-bound in Equation~\eqref{eq:cvtgp_loss_data_marginal_suffstats},
we defining some auxiliary quantities
\begin{align}
    \Ab &= \left( \KbCC + \SigmabetaC \right)^{-1} = \KbCC^{-1} - \SigmabCC^{-1}\\
    \text{ where } & \SigmabCC^{-1} = \KbCC^{-1} -  \left( \KbCC + \SigmabetaC \right)^{-1}
\end{align}
to write it explicitly in terms of its parameters:
\begin{align}
	\Loss_{CVGP} 
	&= -\frac{N}{2} \log (2\pi) + \frac{M}{2} - \frac{1}{2} \log\left|\sigma^2 \Ib_N \right| - \frac{1}{2} \log \left| \KbCC \right|-\frac{1}{2} \yb^\top \sigma^{-2} \yb  \nonumber \\
	& + \sigma^{-2} \mbtilde{\fbC}^\top \KbCC^{-1}  \KbCX \yb 
	-\frac{1}{2} \mbtilde{\fbC}^\top \KbCC^{-1} \SigmabCC \KbCC^{-1}  \mbtilde{\fbC} \nonumber \\
	& - \frac{1}{2 \sigma^{2}} \tr{\KbXX - \KbXC \KbCC^{-1} \KbCX} \nonumber \\
	& - \frac{1}{2 } \tr{ \KbCC^{-1} \SigmabCC \KbCC^{-1} \Kbtilde{\fbC}{\fbC} } 
	+ \frac{1}{2} \log \left| \Kbtilde{\fbC}{\fbC} \right| \\
	&= -\frac{N}{2} \log (2\pi) + \frac{M}{2} - \frac{1}{2} \log\left|\sigma^2 \Ib_N \right| - \frac{1}{2} \log \left| \KbCC \right|-\frac{1}{2} \yb^\top \sigma^{-2} \yb  \nonumber \\
	& + \sigma^{-2} \ybC^\top \left(\KbCC  + \SigmabetaC \right)^{-1} \KbCC 
	\KbCC^{-1}  \KbCX \yb \\
	& -\frac{1}{2} \ybC^\top \left(\KbCC  + \SigmabetaC \right)^{-1} \KbCC 
	\KbCC^{-1} \SigmabCC \KbCC^{-1} 
	 \KbCC \left( \KbCC + \SigmabetaC \right)^{-1} \ybC  \nonumber \\
	& - \frac{1}{2 \sigma^{2}} \tr{\KbXX - \KbXC \KbCC^{-1} \KbCX} \nonumber \\
	& - \frac{1}{2 } \tr{ \KbCC^{-1} \SigmabCC \KbCC^{-1} \left( \KbCC - \KbCC \left( \KbCC + \SigmabetaC \right)^{-1} \KbCC \right) } \\
	& + \frac{1}{2} \log \left| \KbCC - \KbCC \left( \KbCC + \SigmabetaC \right)^{-1} \KbCC \right| \\
	&= -\frac{N}{2} \log (2\pi) + \frac{M}{2} - \frac{1}{2} \log\left|\sigma^2 \Ib_N \right| - \frac{1}{2} \log \left| \KbCC \right|-\frac{1}{2} \yb^\top \sigma^{-2} \yb  \nonumber \\
	& + \sigma^{-2} \ybC^\top \left(\KbCC  + \SigmabetaC \right)^{-1} \KbCX \yb \\
	& -\frac{1}{2} \ybC^\top \left(\KbCC  + \SigmabetaC \right)^{-1} 
	\SigmabCC \left( \KbCC + \SigmabetaC \right)^{-1} \ybC  \nonumber \\
	& - \frac{1}{2 \sigma^{2}} \tr{\KbXX - \KbXC \KbCC^{-1} \KbCX} \nonumber \\
	& - \frac{1}{2 } \tr{ \KbCC^{-1} \SigmabCC } \\
	& + \frac{1}{2 } \tr{ \KbCC^{-1} \SigmabCC \left( \KbCC + \SigmabetaC \right)^{-1} \KbCC } \\
	& + \frac{1}{2} \log \left| \KbCC - \KbCC \left( \KbCC + \SigmabetaC \right)^{-1} \KbCC \right| \;.
\label{eq:cvtgp_loss_data_marginal_params}
\end{align}

We now compute the derivatives with respect to its free parameters
\begin{align}
	´\frac{\partial \Loss_{CVGP}}{\partial \ybC} &= 
		\sigma^{-2} \left(\KbCC  + \SigmabetaC \right)^{-1} \KbCX \yb 
		-\left(\KbCC  + \SigmabetaC \right)^{-1} \SigmabCC \left( \KbCC + \SigmabetaC \right)^{-1} \ybC  \; , \\
	\frac{\partial \Loss_{CVGP}}{\partial \Ab } &= 
	\sigma^{-2} \ybC^\top \KbCX \yb - \ybC^\top \SigmabCC A \ybC  \nonumber \\
	& \qquad + \frac{1}{2 } \tr{ \KbCC^{-1} \SigmabCC \KbCC } \nonumber \\
	& \qquad - \frac{1}{2} \tr{ \left(\KbCC - \KbCC \left( \KbCC + \SigmabetaC \right)^{-1} \KbCC \right)^{-1} \KbCC \KbCC} \; .
\end{align}

We can readily resolve that
\begin{align}
	\ybC^* &= \sigma^{-2} \left( \KbCC + \SigmabetaC \right) \SigmabCC^{-1} \KbCX \yb  \\
		&= \sigma^{-2} A^{-1} \SigmabCC^{-1} \KbCX \yb  \\
\end{align}
and replace it in the covariance expression
\begin{align}
	0 & = \sigma^{-2} \sigma^{-2} \yb^\top \KbXC \SigmabCC^{-1} A^{-1} \KbCX \yb \nonumber \\
	& \qquad -  \sigma^{-2} \yb^\top \KbXC \SigmabCC^{-1} A^{-1} \SigmabCC A \sigma^{-2} A^{-1} \SigmabCC^{-1} \KbCX \yb  \nonumber \\
	& \qquad + \frac{1}{2 } \tr{ \KbCC^{-1} \SigmabCC \KbCC } \nonumber \\
	& \qquad - \frac{1}{2} \tr{ \left(\KbCC - \KbCC \left( \KbCC + \SigmabetaC \right)^{-1} \KbCC \right)^{-1} \KbCC \KbCC} \\
	0 & = \sigma^{-2} \sigma^{-2} \yb^\top \KbXC \SigmabCC^{-1} A^{-1} \KbCX \yb \nonumber \\
	& \qquad -  \sigma^{-2} \sigma^{-2} \yb^\top \KbXC \SigmabCC^{-1} A^{-1} \KbCX \yb  \nonumber \\
	& \qquad + \frac{1}{2 } \tr{ \KbCC^{-1} \SigmabCC \KbCC } \nonumber \\
	& \qquad - \frac{1}{2} \tr{ \left(\KbCC - \KbCC \left( \KbCC + \SigmabetaC \right)^{-1} \KbCC \right)^{-1} \KbCC \KbCC} \\
	0 & = \frac{1}{2 } \tr{ \KbCC^{-1} \SigmabCC \KbCC } \nonumber \\
	& \qquad - \frac{1}{2} \tr{ \left(\KbCC - \KbCC \left( \KbCC + \SigmabetaC \right)^{-1} \KbCC \right)^{-1} \KbCC \KbCC}
\end{align}
Equating the matrices inside the traces, we have
\begin{align}
	\KbCC^{-1} \SigmabCC \KbCC  & = \left(\KbCC - \KbCC \left( \KbCC + \SigmabetaC \right)^{-1} \KbCC \right)^{-1} \KbCC \KbCC \\
	\KbCC^{-1} \SigmabCC & = \left(\KbCC - \KbCC \left( \KbCC + \SigmabetaC \right)^{-1} \KbCC \right)^{-1} \KbCC \\
	\KbCC^{-1} \SigmabCC \KbCC^{-1} & = \left(\KbCC - \KbCC \left( \KbCC + \SigmabetaC \right)^{-1} \KbCC \right)^{-1} \\
	\KbCC \SigmabCC^{-1} \KbCC & = \left(\KbCC - \KbCC \left( \KbCC + \SigmabetaC \right)^{-1} \KbCC \right) \\
	\KbCC \SigmabCC^{-1} \KbCC & = \KbCC \left( \KbCC^{-1}- \left( \KbCC + \SigmabetaC \right)^{-1} \right) \KbCC\\
	\SigmabCC^{-1} & = \left( \KbCC^{-1}- \left( \KbCC + \SigmabetaC \right)^{-1} \right) \\
	\left( \KbCC + \SigmabetaC \right)^{-1} & = \left( \KbCC^{-1} - \SigmabCC^{-1} \right) \\
        \KbCC^{-1} - \KbCC^{-1} \left(\SigmabetaC^{-1} + \KbCC^{-1}\right)^{-1}\KbCC^{-1} & = \left( \KbCC^{-1} - \SigmabCC^{-1} \right) \\
        \KbCC^{-1} \left(\SigmabetaC^{-1} + \KbCC^{-1}\right)^{-1}\KbCC^{-1} & = \SigmabCC^{-1} \\
        \KbCC \left(\SigmabetaC^{-1} + \KbCC^{-1}\right)\KbCC & = \SigmabCC \\
        \KbCC \SigmabetaC^{-1} \KbCC + \KbCC \KbCC^{-1} \KbCC & = \SigmabCC = \KbCC + \frac{1}{\sigma^2} \KbCX \KbXC \\
        \KbCC \SigmabetaC^{-1} \KbCC & = \frac{1}{\sigma^2} \KbCX \KbXC \\
        \SigmabetaC^{-1} & = \frac{1}{\sigma^2} \KbCC^{-1} \KbCX \KbXC \KbCC^{-1}\\
        \SigmabetaC^* & = \sigma^2 \KbCC \left(\KbCX \KbXC\right)^{-1} \KbCC
\end{align}

We now elaborate on the optimal values for CVGP's pseudo-coresets,
rewriting CVGP's optimal pseudo-observations as
\begin{align}
	\ybC^* &= \sigma^{-2} \left( \KbCC + \SigmabetaC \right)\SigmabCC^{-1} \KbCX \yb \\
	&= \sigma^{-2} \left( \KbCC + \SigmabetaC \right) \left[ \KbCC^{-1} -  \left( \KbCC + \SigmabetaC \right)^{-1} \right] \KbCX \yb \\
	&= \sigma^{-2} \left[ \left( \KbCC + \SigmabetaC \right) \KbCC^{-1} -  \Ib_M \right] \KbCX \yb \\
	&= \sigma^{-2} \left[ \KbCC \left[ \KbCC^{-1} + \sigma^{2} \left(\KbCX \KbXC\right)^{-1} \right] \KbCC \KbCC^{-1} -  \Ib_M \right] \KbCX \yb \\
	&= \sigma^{-2} \left[ \Ib_M  + \sigma^{2} \KbCC \left(\KbCX \KbXC\right)^{-1}  -  \Ib_M \right] \KbCX \yb \\
	&= \sigma^{-2} (\sigma^{2} \KbCC \left(\KbCX \KbXC\right)^{-1} \KbCX) \yb \\
        &= \sigma^{-2} \SigmabetaC^* \yb \\
        &= \KbCC \left(\KbCX \KbXC\right)^{-1} \KbCX \yb
\end{align}

\newpage

With this optimal values, we can now rewrite the lower-bound at its maxima
\begin{align}
	\Loss_{CVGP}(\ybC^*, \SigmabetaC^*)
	&= -\frac{N}{2} \log (2\pi) + \frac{M}{2} - \frac{1}{2} \log\left|\sigma^2 \Ib_N \right| - \frac{1}{2} \log \left| \KbCC \right|-\frac{1}{2} \yb^\top \sigma^{-2} \yb  \nonumber \\
	& + \sigma^{-2} \ybC^{*^{\top}} \left(\KbCC  + \SigmabetaC^* \right)^{-1} \KbCX \yb \\
	& -\frac{1}{2} \ybC^{*^{\top}} \left(\KbCC  + \SigmabetaC^* \right)^{-1} 
			\SigmabCC \left( \KbCC + \SigmabetaC^* \right)^{-1} \ybC^*  \nonumber \\
	& - \frac{1}{2 \sigma^{2}} \tr{\KbXX - \KbXC \KbCC^{-1} \KbCX} \nonumber \\
	& - \frac{1}{2 } \tr{ \KbCC^{-1} \SigmabCC } + \frac{1}{2 } \tr{ \KbCC^{-1} \SigmabCC \left( \KbCC + \SigmabetaC^* \right)^{-1} \KbCC } \\
	& + \frac{1}{2} \log \left| \KbCC - \KbCC \left( \KbCC + \SigmabetaC^* \right)^{-1} \KbCC \right| \\
	&= -\frac{N}{2} \log (2\pi) + \frac{M}{2} - \frac{1}{2} \log\left|\sigma^2 \Ib_N \right| - \frac{1}{2} \log \left| \KbCC \right|-\frac{1}{2} \yb^\top \sigma^{-2} \yb  \nonumber \\
	& + \sigma^{-2} \sigma^{-2} \yb^\top \KbXC \SigmabCC^{-1} \left( \KbCC + \SigmabetaC \right)  \left(\KbCC  + \SigmabetaC^* \right)^{-1} \KbCX \yb \\
	& -\frac{1}{2} \sigma^{-2} \yb^\top \KbXC \SigmabCC^{-1} \left( \KbCC + \SigmabetaC \right) \left(\KbCC  + \SigmabetaC^* \right)^{-1} 
	\SigmabCC \left( \KbCC \right.\\
    & \left.\quad \quad \quad \quad  \quad \quad + \SigmabetaC^* \right)^{-1} \sigma^{-2} \left( \KbCC + \SigmabetaC \right)\SigmabCC^{-1} \KbCX \yb \nonumber \\
	& - \frac{1}{2 \sigma^{2}} \tr{\KbXX - \KbXC \KbCC^{-1} \KbCX} \nonumber \\
	& - \frac{1}{2 } \tr{ \KbCC^{-1} \SigmabCC } + \frac{1}{2 } \tr{ \KbCC^{-1} \SigmabCC \left(\KbCC^{-1} - \SigmabCC^{-1} \right) \KbCC } \\
	& + \frac{1}{2} \log \left| \KbCC - \KbCC \left( \KbCC^{-1} - \SigmabCC^{-1} \right) \KbCC \right| \\
	&= -\frac{N}{2} \log (2\pi) + \frac{M}{2} - \frac{1}{2} \log\left|\sigma^2 \Ib_N \right| - \frac{1}{2} \log \left| \KbCC \right|-\frac{1}{2} \yb^\top \sigma^{-2} \yb  \nonumber \\
	& + \sigma^{-2} \sigma^{-2} \yb^\top \KbXC \SigmabCC^{-1} \KbCX \yb \\
	& -\frac{1}{2} \sigma^{-2} \yb^\top \KbXC \sigma^{-2} \SigmabCC^{-1} \KbCX \yb \nonumber \\
	& - \frac{1}{2 \sigma^{2}} \tr{\KbXX - \KbXC \KbCC^{-1} \KbCX} \nonumber \\
	& - \frac{1}{2 } \tr{ \KbCC^{-1} \SigmabCC } + \frac{1}{2 } \tr{ \KbCC^{-1} \SigmabCC} - \frac{1}{2 } \tr{ \KbCC^{-1} \KbCC } \\
	& + \frac{1}{2} \log \left| \KbCC \left( \KbCC^{-1} - \left( \KbCC^{-1} - \SigmabCC^{-1} \right)\right) \KbCC \right| \\
	&= -\frac{N}{2} \log (2\pi) + \frac{M}{2} - \frac{1}{2} \log\left|\sigma^2 \Ib_N \right| - \frac{1}{2} \log \left| \KbCC \right|-\frac{1}{2} \yb^\top \sigma^{-2} \yb  \nonumber \\
	& + \frac{1}{2} \sigma^{-2} \sigma^{-2} \yb^\top \KbXC \SigmabCC^{-1} \KbCX \yb \nonumber \\
	& - \frac{1}{2 \sigma^{2}} \tr{\KbXX - \KbXC \KbCC^{-1} \KbCX} - \frac{1}{2 } M \nonumber \\
	& + \frac{1}{2} \log \left| \KbCC \SigmabCC^{-1} \KbCC \right| \\
	&= -\frac{N}{2} \log (2\pi) \nonumber \\
	& -\frac{1}{2} \yb^\top \sigma^{-2} \left( \Ib_M + \sigma^{-2} \KbXC \SigmabCC^{-1} \KbCX \right) \yb \nonumber \\
	& - \frac{1}{2 \sigma^{2}} \tr{\KbXX - \KbXC \KbCC^{-1} \KbCX} \nonumber \\
	& - \frac{1}{2} \log\left|\sigma^2 \Ib_N \right| + \frac{1}{2} \log \left| \SigmabCC^{-1} \KbCC \right| \\
	&= \log \N{ \yb \mid \zerob, \sigma^2 \Ib_N + \Qb_{\fbC,\fbC} } - \frac{1}{2 \sigma^2}\tr{ \KbXX - \Qb_{\fbC,\fbC} } \label{eq:cvtgp_loss_data_marginal_optima}
\end{align}
which, for $M=Z$, and $\XbC = \Xb_Z$\footnote{Matching notations.}, corresponds with the same lower-bound as demonstrated by~\citet{titsias2009variational} for SparseGP. $\mathcal{L}_{CVGP} \leq \mathcal{L}_{SparseGP}$ and the bound is tight with equality when  $\KbCC \left(\KbCX \KbXC\right)^{-1} \KbCC$ is diagonal.

\clearpage
\subsection{COMPLEXITIES OF CVGP}
\label{assec:CVTGP_complexities}
CVGP maintains the time and space complexity of SVGP with less parameters.
This is because CVGP does not need to learn a free-form covariance matrix $\Sb$,
but only the coreset values $\XbC, \ybC$ and their weights $\betabC$, which is more space efficient ---note that these three are $C$-dimensional vectors.
We describe the complexities of benchmarks and their parameters in Table~\ref{table:bigo2} below.

\begin{table}[h]
    \rowcolors{5}{}{gray!10}
    \begin{tabular}{*5c}
        \toprule
        & \multicolumn{4}{c}{Complexities$\;\;\;\;\;\;\;\;\;\;\;\;\;\;\;\;\;\;\;\;\;\;\;\;$} \\
        \cmidrule(lr){2-4}
        Inference technique & Time  & Space & \# Parameter & Parameters\\    
        \midrule
        SparseGP \citep{titsias2009variational} &  $\;$$\;$$\;$$ \bigO{NM^2}$       &    $\;$$\;$$\;$$\bigO{NM^2}$    & $\bigO{M}$ & $\XbZ$ \\
        SVGP \citep{hensman2013gaussian} &     $\bigO{M^3}$   & $\bigO{M^2}$ &      $\;$$\;$$\bigO{M^2}$ & $\XbZ, \mb, \Sb$  \\
        CVGP &  $\bigO{M^3}$      & $\bigO{M^2}$ &  $\bigO{M}$ & $\XbC, \ybC, \betabC$ \\
        \bottomrule
    \end{tabular}
    \centering
    \caption{Computational analysis of CVGP and sparse $\gp$ alternatives:
    time and space complexities for obtaining an unbiased estimate of the log-marginal likelihood. CVGP enjoys same time and space complexity as SVGP, yet with a reduced variational parameter dimensionality. 
    Contrary to SVGP, CVGP does not learn a free-form covariance parameter $\Sb$, but only tempering-parameters $\betabC$ of same size as $\XbC, \ybC$.
    \label{table:bigo2}}
    \vspace{-1ex}
\end{table}

\clearpage
\section{EXPERIMENTS: SET-UP AND ADDITIONAL DETAILS}
\label{asec:experiments}

\vspace{-1ex}

\subsection{DATASETS}
\label{assec:datasets}

In this section, we describe the simulated and real-world datasets used in our experiments.
The generative processes of simulated data are explained below in Section~\ref{asssec:datasets_sim}.
We use UCI machine learning repository for real-world datasets \citep{UCIMachi51:online}, as described in Section~\ref{asssec:datasets_real}. For all datasets, $\Xb$ are normalized (0 centered and unit variance) before training. 
\subsubsection{Real-world Datasets}
\label{asssec:datasets_real}

\paragraph{Physicochemical properties of protein tertiary structure dataset (protein).} 
A physicochemical data collection containing the properties of protein tertiary structure, specifically sourced from CASP 5-9. The dataset includes 45730 data points and 9 features \citep{rana2013physicochemical}.

\paragraph{Bike sharing dataset (bike).} A bike sharing dataset comprised of 17 features and 17379 data points \citep{bike}.

\paragraph{Parkinsons telemonitoring dataset (parkinsons).} A biomedical voice measurements dataset obtained from 42 individuals in the early stages of Parkinson's disease. These individuals were enrolled in a six-month trial for remote symptom progression monitoring, using a telemonitoring device \citep{little2007exploiting}. There are 20 features and 5875 datapoints.

\paragraph{SkillCraft1 master table dataset (skillcraft).} A video gaming telemetry data collection consisting of 12 features and 3338 data points \citep{thompson2013video}.

\paragraph{Wine quality dataset (wine).} A collection of red wine samples with 11 features that are used to predict the wine's quality. In total, there are 1600 data points available for analysis \citep{cortez2009modeling}.

\subsubsection{Simulated Datasets}
\label{asssec:datasets_sim}

We generate $1000$ examples for each of the following synthetic datasets.
\paragraph{Synthetic 1.} A 1-dimensional dataset following the below generative process:

\begin{align}
    f &= \frac{2}{5}  \left(\sin{3x} \cos{2x} + \sin{\frac{x}{2}} + \cos{2x} + \exp\left\{-x^2\right\} + |x| \right), & x \sim U(-4,4) \;, \\
    y &= f + \epsilon \sin2\pi f \;, & \epsilon \sim \N{\epsilon \mid 0, 3 \times 10^{-1}} \; .
\end{align}

\paragraph{Synthetic 2.} A 1-dimensional dataset following the below generative process:

\begin{align}
    f &= \sin x^2 + \cos x^2 + \sin 3x + \cos 5x + \frac{\sqrt{|x|}}{2} \;, & x \sim U(-4,4) \;,\\
    y &= f + \epsilon \sin2\pi f \;, & \epsilon \sim \N{\epsilon \mid 0, 3 \times 10^{-1}} \;.
\end{align}

\paragraph{Synthetic 3.} A 1-dimensional dataset following the below generative process:

\begin{align}
    f  &= \cos 2\pi x \;, & x \sim U(0,2) \;, \\
    y &= f + \epsilon x^3 \;, & \epsilon \sim \N{\epsilon \mid 0, 1} \;.
\end{align}

\paragraph{Synthetic 4.} A 2-dimensional dataset following the below generative process:

\begin{align}
    \xb &\sim \texttt{MakeBlobs}(centers=3, std=0.4) \;,\\
    f_1  &= 4 \sin x_1 + 2 \sin 2 x_1 \;, \\
    f_2  &= 3 \cos 3 x_2 + 4 \sin 5 x_2 \;, \\
    f_{12} &=  \expp{-(x_1 + x_2)^2} \;, \\
    y &= f_1 + f_2 + f_{12} + \epsilon \;, & \epsilon \sim \N{\epsilon \mid 0, 2 \times 10^{-1}} \;.
\end{align}
where the function \texttt{MakeBlobs} is implemented as in \citet{scikit-learn}.

\paragraph{Synthetic 5.} A 2-dimensional dataset following the below generative process:

\begin{align}
    \xb &\sim \texttt{MakeMoons}( noise=0.05)\\
    f_1  &= \frac{x_1}{2} + \sin 2 x_1\\
    f_2  &= \frac{x_2}{2}+ \cos 5 x_2 \\
    f_{12} &=   \frac{\expp{
    -(x_1 + x_2)^2}}{2}\\
    y &= f_1 + f_2 + f_{12} + \epsilon \;, & \epsilon \sim \N{\epsilon \mid 0, 2 \times 10^{-1}}
\end{align}
where the function 
\texttt{MakeMoons} is implemented by \citet{scikit-learn}.

\clearpage
\subsection{BASELINES}
\label{assec:baselines}

We use the GPytorch \citep{gardner2018gpytorch} implementation of SparseGP, SGVP, and PPGPR.
For ExactGP, we simply use the derivation of \citet{rasmussen2006gaussian} implemented using the \texttt{MultivariateNormal} method of Pytorch \citep{paszke2019pytorch}.

\paragraph{SparseGP.} Introduced by \citep{titsias2009variational}, SparseGP offers a variational solution to inducing point methods. In particular, SparseGP minimizes the KL divergence between an approximate and true posterior distribution. The loss function is derived by finding and plugging the optimal posterior variational distribution, which can be derived in terms of the $\gp$ kernel parameters.

\paragraph{SVGP.} SVGP minimizes the KL divergence between an approximate and true posterior distribution where the posterior distribution is explicitly defined \citep{hensman2013gaussian}. The parameters of the posterior and model are learned jointly.
SVGP is an stochastic approximation to SparseGP, which allows computationally efficient learning.

\paragraph{PPGPR.} PPGPR is a variational predictive method for $\gp$s that, instead of lower-bounding the prior-predictive distribution as the methods above,
proposed to optimize a lower-bound over the posterior predictive \citep{jankowiak2020parametric}.
This method provides predictive uncertainty estimates that model the variance of the observed data more accurately.
PPGPR results in~\citet{jankowiak2020parametric} were based on 400 epochs, which we found insufficient for convergence to optimal RMSE in our experiments. Although longer training helps with better predictive RMSE performance, it also causes severe overfitting on noise ---a behavior not observed in other $\gp$ algorithms.

\paragraph{ExactGP.} The exact learning of Gaussian processes as described by \citet{rasmussen2006gaussian}. We compute the marginal log-likelihood (prior predictive) by integrating the likelihood over the latent function-space (with respect to prior distribution) to learn model hyperparameters, at a computational complexity of $\bigO{N^3}$.

\subsection{EXPERIMENT DETAILS FOR REPRODUCIBILITY}
\label{assec:reproducibility}
We employ 5-fold cross-validation
to compute and report each variational technique's lower-bound objective $\mathcal{L}$ in inference,
and their predictive root-mean-squared error (RMSE) over held out test splits.

We do not scale the KL-divergence terms in each model's objective, for them to be valid lower-bounds.
We use a fixed random seed over all datasets to ensure that the folds (with 70\%-30\% train and validations splits) for different models are the same. 

We use Adam optimizer with a learning rate of  $10^{-3}$ for all methods and single precision floating point \citep{kingma2014adam}.
For the techniques amenable to stochastic optimization (SVGP, PPGPR, and CVGP), we use a batch size of 512.
Each model is run on a single NVIDIA$^{\text{\textregistered}}$ GeForce$^{\text{\textregistered}}$ RTX 20 series graphics card.

To leverage full model capacity and achieve full optimization performance,
we train for $10^5$ epochs maximum, and stop training only if there are no RMSE improvements for $3\times 10^3$ consecutive epochs over the validation set.
We early stop with respect to the best \emph{held-out} validation set RMSE metric attained.

\newpage
\section{ADDITIONAL EXPERIMENTS}

We showcase different performance findings. For box plots, unless otherwise stated, the best performing stochastic model -- in comparison to other stochastic sub-sampling methods -- is showcased in bold. 

\label{assec:app_experiments}
\subsection{EVOLUTION OF PERFORMANCE METRICS BY TRAINING EPOCHS}\label{asec:epoch_results}

\begin{figure*}[!ht]
    \centering
\includegraphics[width=\textwidth]{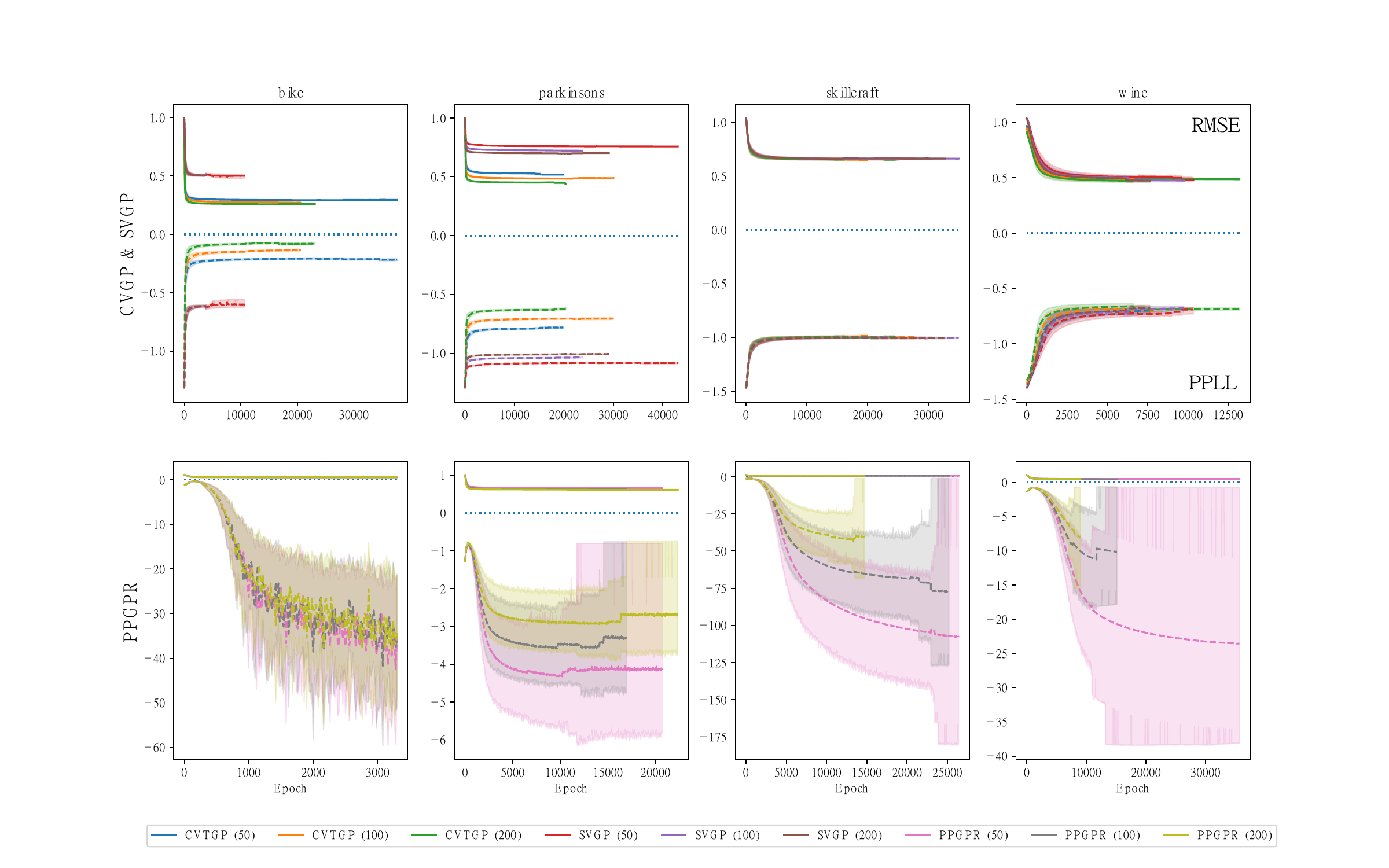}
        \caption{
Evolution of RMSE and PPLL across training epochs. For CVGP and SVGP, validation RMSE and PPLL consistently decrease ---showing no critical indication of overfitting. In contrast, PPGPR’s RMSE improves, but PPLL worsens, indicating overfitting of noise and cross-correlation, leading to suboptimal PPLL. Training stops only when RMSE no longer improves. Large negative PPLL values prevent reporting PPGPR’s PPLL in Figure \ref{fig:exp_predictive_real_all}. 
}
\label{fig:epoch_results}
\end{figure*}

\newpage
\subsection{MODEL LEARNING AND INFERENCE GAPS}\label{asec:gap_results}
\begin{figure*}[!ht]
        \includegraphics[width=\textwidth]{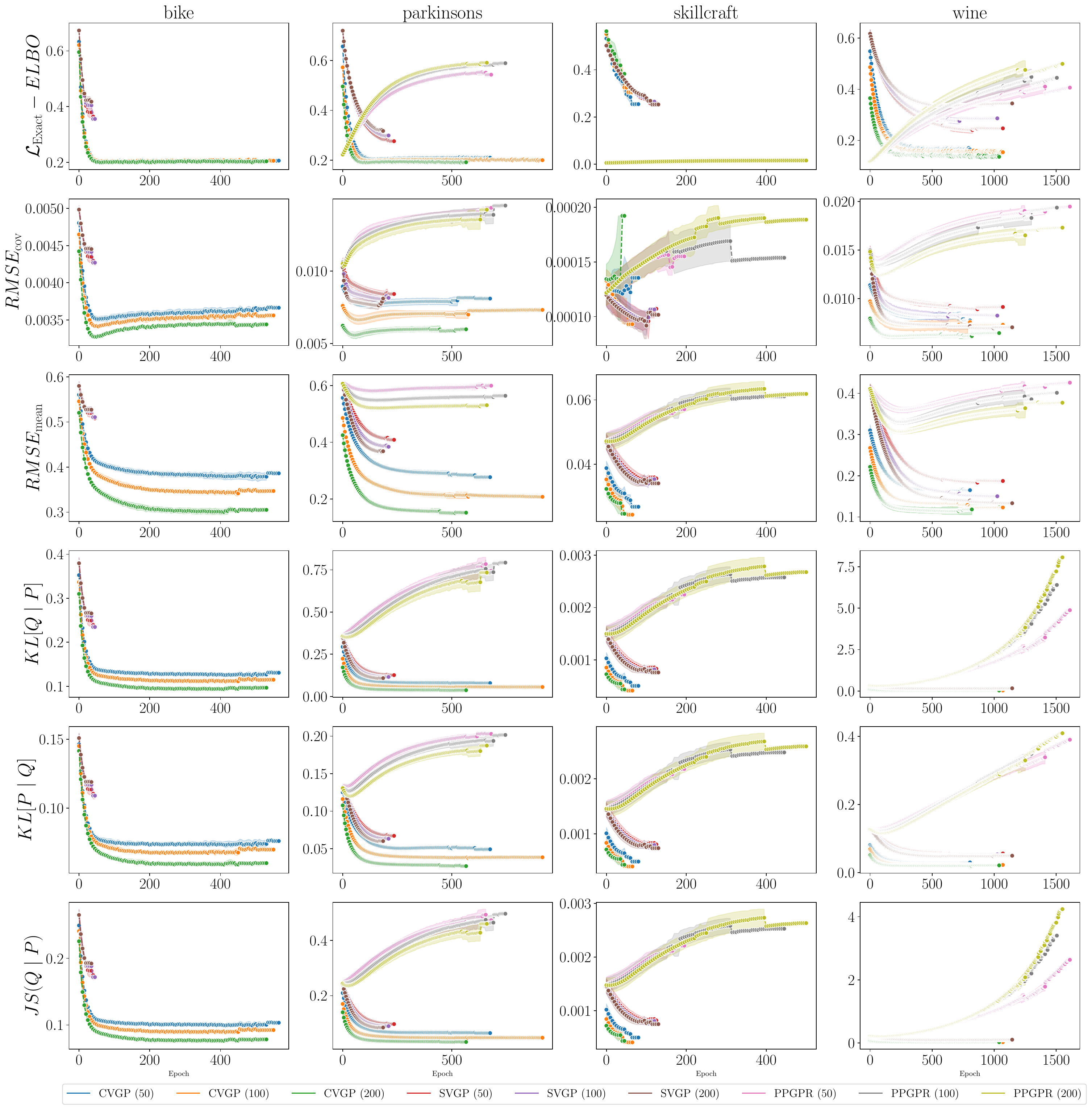}
 \label{fig:gap_results_all}
    \caption{
Divergence from the true posterior $\cp{f^\star}{\xb^\star, \Xb, \yb}$ and lower-bound tightness relative to the exact solution (\ie the difference in log-marginal and ELBO, always positive). PPGPR does not lower-bound ExactGP and diverges, as shown in the figure. By directly fitting noisy observations, PPGPR overfits early, while SVGP and CVGP filter noise.
    }
    \label{fig:sim}
\end{figure*}

\clearpage

\subsection{ROBUSTNESS TO INITIALIZATION}
\label{asssec:app_exp_robustness}

Below, we demonstrate CVGP's robustness to random initialization. We observe that RandomCVGP (CVGP initialized with white Gaussian noise) performs on par with CVGP in almost all cases and metrics. For very big datasets with many input-features (\eg Song), a random initialization over high-dimensional input-output spaces is a clear disadvantage.
Hence, we recommend, in general, to initialize CVGP with k-means.


\begin{figure*}[!ht]
    \includegraphics[width=\textwidth]{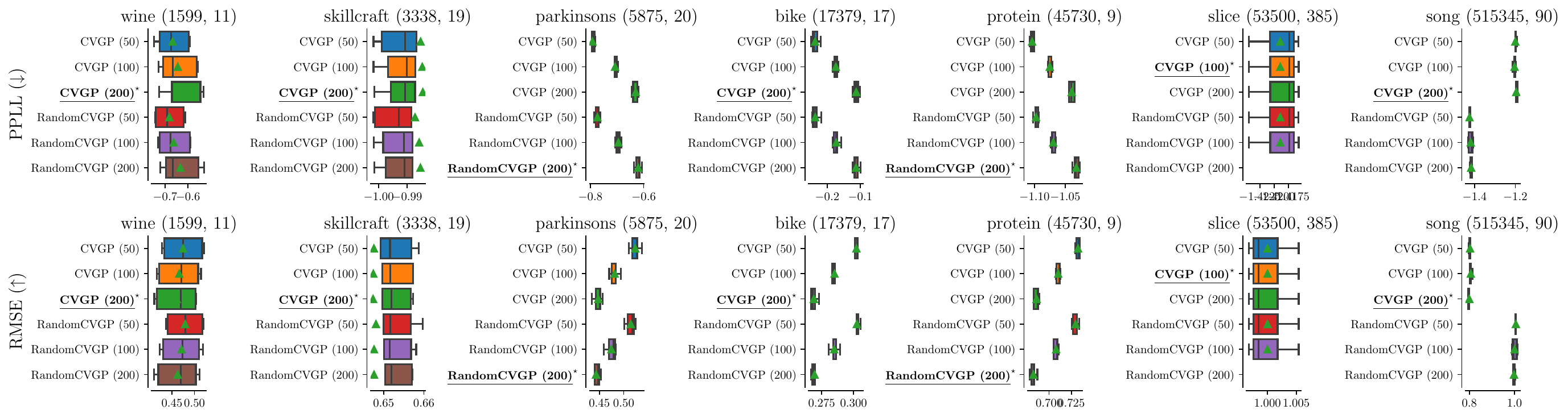}
    \label{fig:sim_stochastic}
    \vspace{-3ex}
    \caption{\small{
    Predictive performance comparison for CVGP when initialized with random points (RandomCVGP) or k-means over the observed (real) datasets.
    CVGP is robust to random initializations.  The best performing initiation mean statistic (\textcolor{darkgreen}{\scalebox{1}{$\blacktriangle$}}) is $\textbf{\underline{emphasized}}^\star$.
    }}
    \label{fig:exp_predictive_cvgp_randomcvgp_real}
\end{figure*}

\newpage

\subsection{POSTERIOR-PRIOR INTERPOLATION: NOISY REAL-WORLD DATA}
\label{assec:app_exp_noisy}

We discuss all sparse $\gp$ method's posterior ability to interpolate between the model prior and the information provided by observations.
For CVGP, as the observation noise increases ($\sigma^2 \rightarrow \infty$),
its posterior mean $\mb_{\fbC | \ybC}$ converges to $\mathbf{0}$,
and its posterior covariance $\Kb_{\fbC | \ybC}$ converges to the prior covariance $\KbZZ$;
\ie the observations are noninformative and CVGP's posterior reverts to the $\gp$ prior.
Conversely, for noiseless data ($\sigma^2 \rightarrow 0$),
CVGP's posterior mean approaches $\ybC$,
and its posterior covariance diminishes to $\mathbf{0}$ (see Equation \ref{eq:cvtgp_q}).
On the contrary, 
SVGP's posterior statistics ($\mathbf{m}, \mathbf{S}$)
have no explicit model dependencies, and therefore,
are adjusted based purely on variational parameter optimization.

We run an empirical experiment below, where
we take a real-world dataset and progressively add noise to the true regression values,
before training SVGP, PPGPR, and CVGP on these extra-noisy versions of the datasets.
As in any Bayesian model,
we expect that for low noise regimes,
the posterior should diverge from the prior to capture the information provided by observations;
while for high noise regimes (uninformative data), the posterior should remain similar to the prior.
Below, we notice that
CVGP effectively resorts to the prior under uninformative data, a behavior exhibited by ExactGP,
while PPGPR does not recover the prior --- fitting the noisy data.


\begin{figure}[!h]
    \centering
    \begin{subfigure}{0.45\textwidth}
        \centering
        \includegraphics[width=\linewidth]{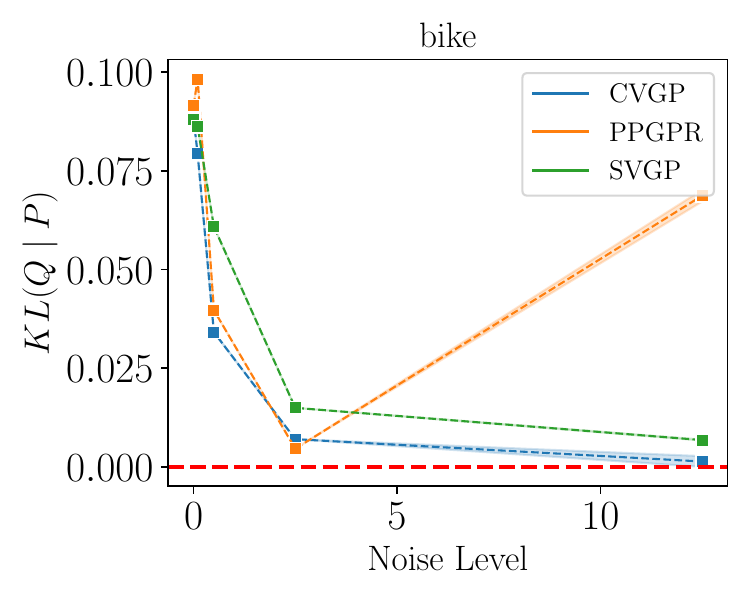}
    \end{subfigure}
    \begin{subfigure}{0.45\textwidth}
        \centering
        \includegraphics[width=\linewidth]{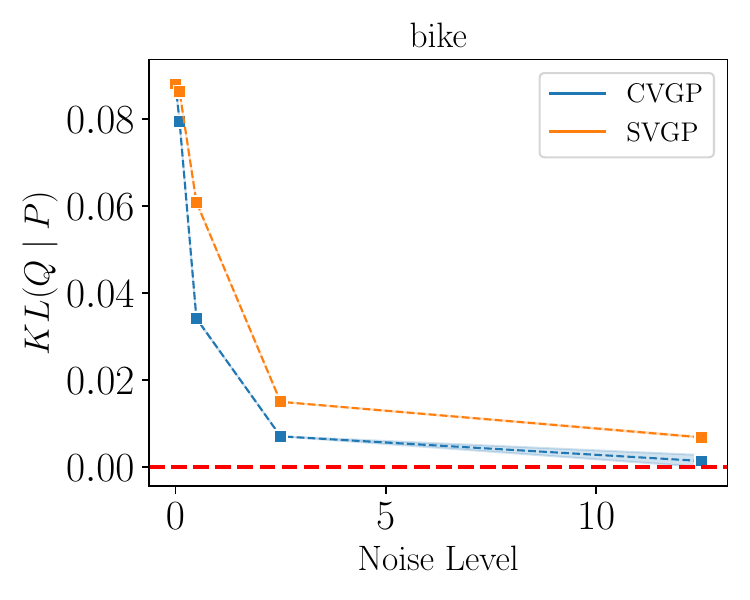}
    \end{subfigure}
    \caption{Study of the difference between sparse $\gp$ approximate posteriors and model prior for the Bike dataset, as measured by the KL-divergence between prior and approximate variational posterior ($\kl{q(\fbC)}{p(\fbC)}$) across different observation noise regimes. Left: CVGP, PPGPR, and SVGP, right: CVGP and SVGP. We see that PPGPR diverges from prior vastly while SVGP and CVGP retains the Gaussian prior-likelihood conjugacy (\ie as noise increase they do not diverge from the prior vastly).} \label{fig:exp_inference_prior_to_posterior2}
\end{figure}

\newpage

\section{QUALITATIVE STUDY}\label{asec:qual_study}

\subsection{QUALITATIVE EVALUATION OF POSTERIOR PREDICTIVE}
\label{asssec:app_exp_posterior_predictive}
\begin{figure}[h!]
    \centering
    \begin{subfigure}[c]{0.85\textwidth}
        \includegraphics[width=\textwidth]{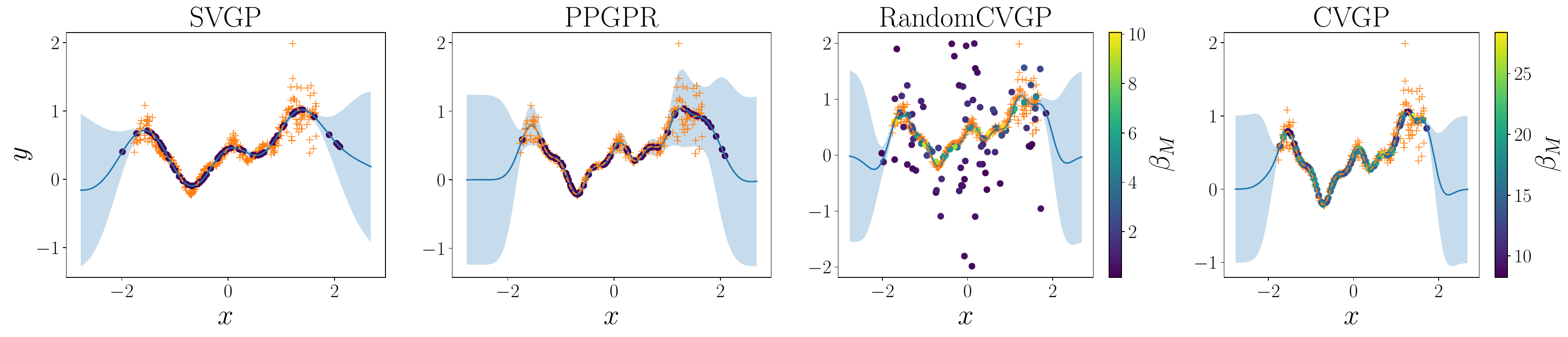}
        \vspace*{-5.2ex}
    \end{subfigure}
    \begin{subfigure}[c]{0.85\textwidth}
        \includegraphics[width=\textwidth]{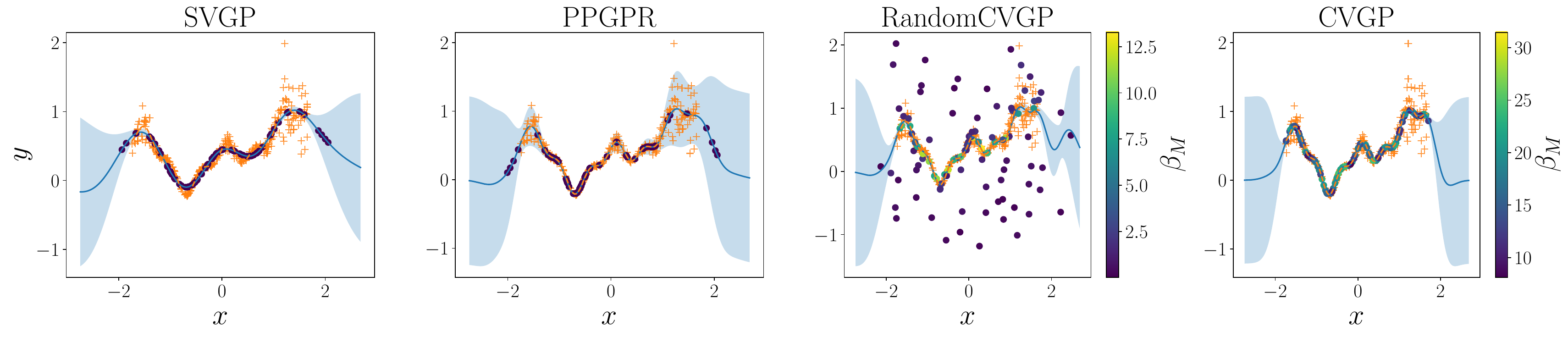}
        \vspace*{-5.2ex}
    \end{subfigure}
    \begin{subfigure}[c]{0.85\textwidth}
        \includegraphics[width=\textwidth]{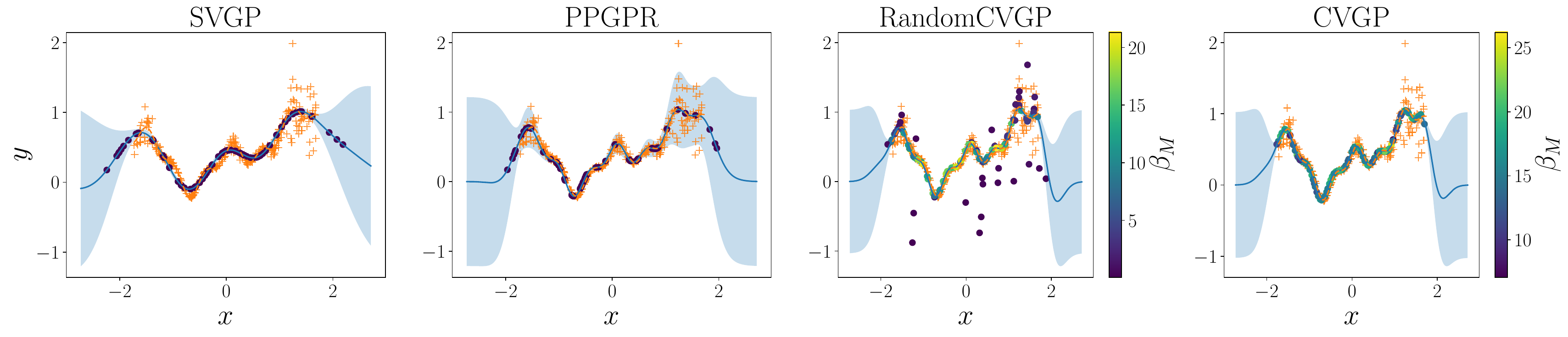}
        \vspace*{-5ex}
    \end{subfigure}
    \begin{subfigure}[c]{0.85\textwidth}
        \includegraphics[width=\textwidth]{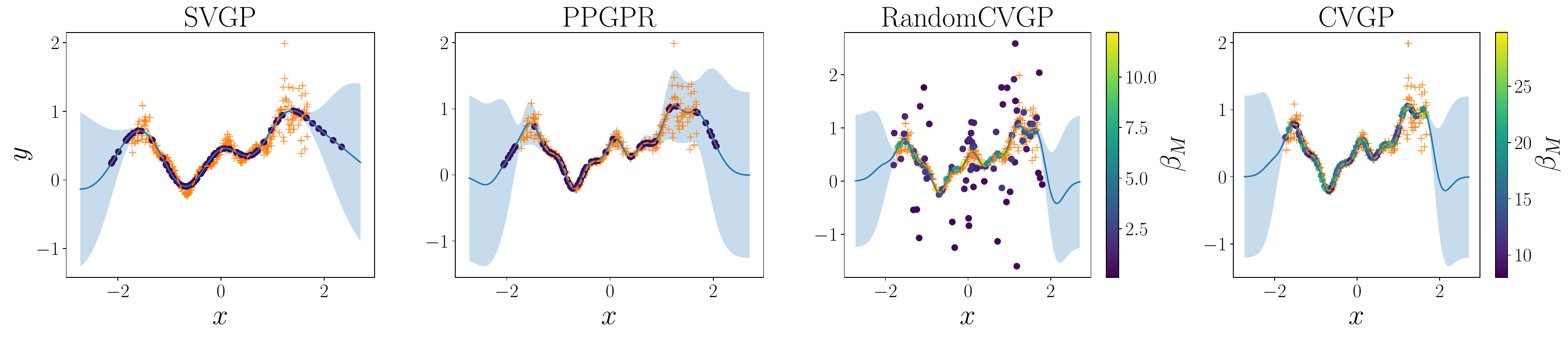}
        \vspace*{-5.2ex}
    \end{subfigure}
    \begin{subfigure}[c]{0.85\textwidth}
        \includegraphics[width=\textwidth]{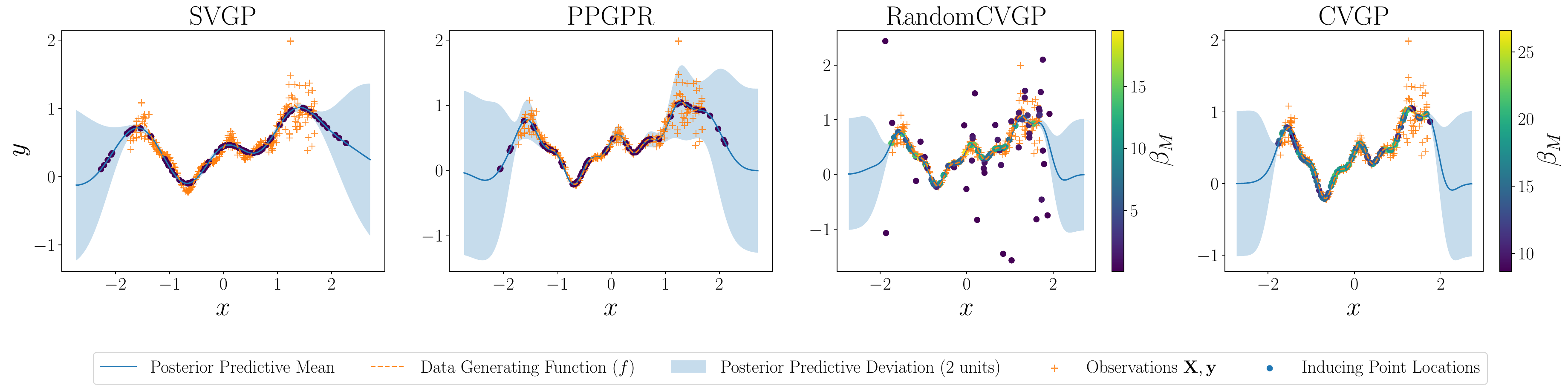}
    \end{subfigure}
    \caption{Posterior predictive distribution for the synthetic 1 dataset across different 5 folds, with 100 inducing points. Note that RandomCVGP is initialized with Gaussian white noise and faces a significantly more challenging task in fitting the data compared to SVGP, PPGPR, and CVGP. In fact, some of its learned inducing points/coresets fall off-the-grid and appear unrelated to the data. In these cases, the corresponding coreset weights are low (indicated in purple). Conversely, points that effectively capture the $y|x$ relationship are shown in yellow-green, while those that do not are depicted in purple—and are consequently disregarded during posterior inference. SVGP and PPGPR do not have this advantage. Hence, we use a single color for their coresets. This design endows CVGP with significant flexibility: if an inducing point proves unhelpful for predictions, its influence can be driven to 0 (with the help of its corresponding $\beta_m$). In contrast, both PPGPR and SVGP must select inducing points that consistently capture the data structure. }
    \label{fig:posterior_predictive_synthetic1}
\end{figure}

\begin{figure}[h!]
    \centering
    \begin{subfigure}[c]{0.85\textwidth}
        \includegraphics[width=\textwidth]{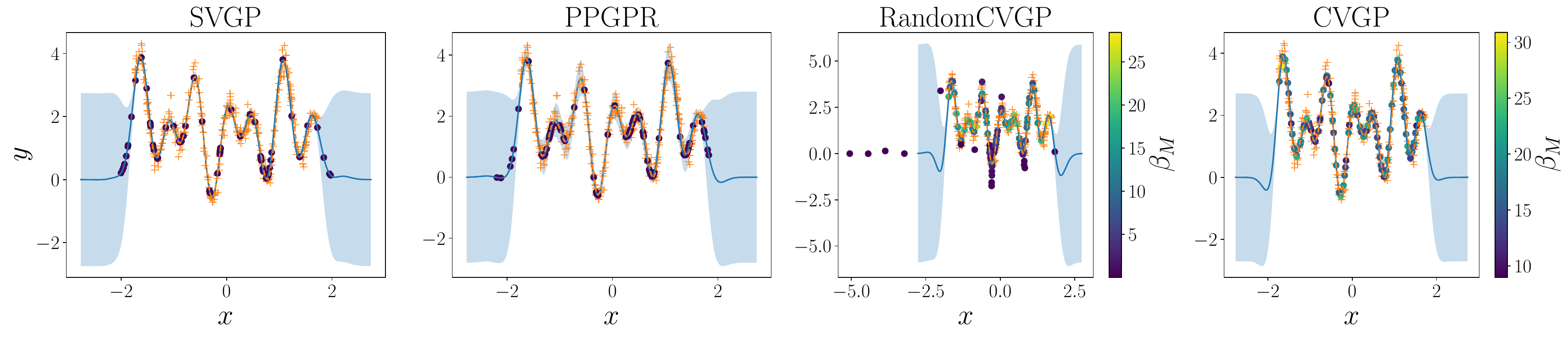}
        \vspace*{-5.2ex}
    \end{subfigure}
    \begin{subfigure}[c]{0.85\textwidth}
        \includegraphics[width=\textwidth]{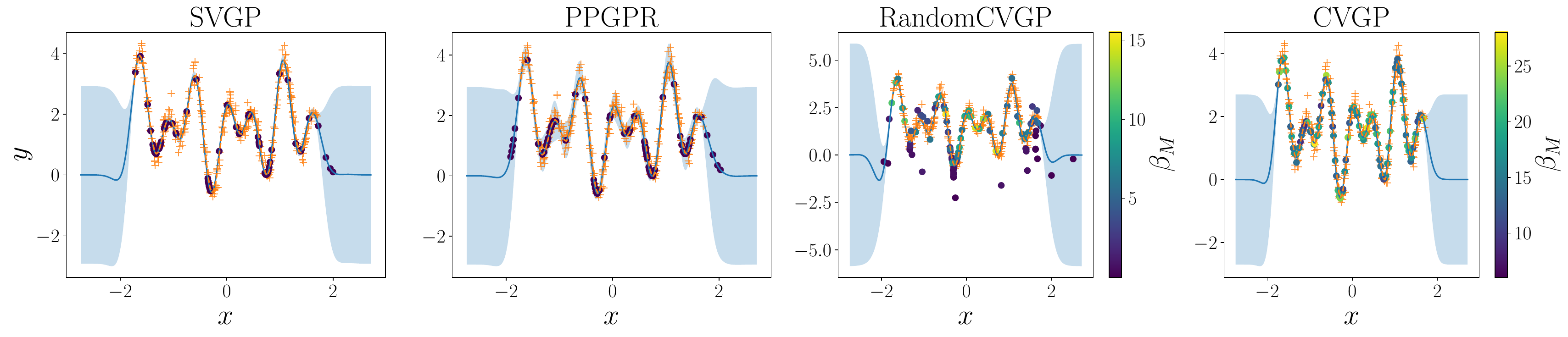}
        \vspace*{-5.2ex}
    \end{subfigure}
    \begin{subfigure}[c]{0.85\textwidth}
        \includegraphics[width=\textwidth]{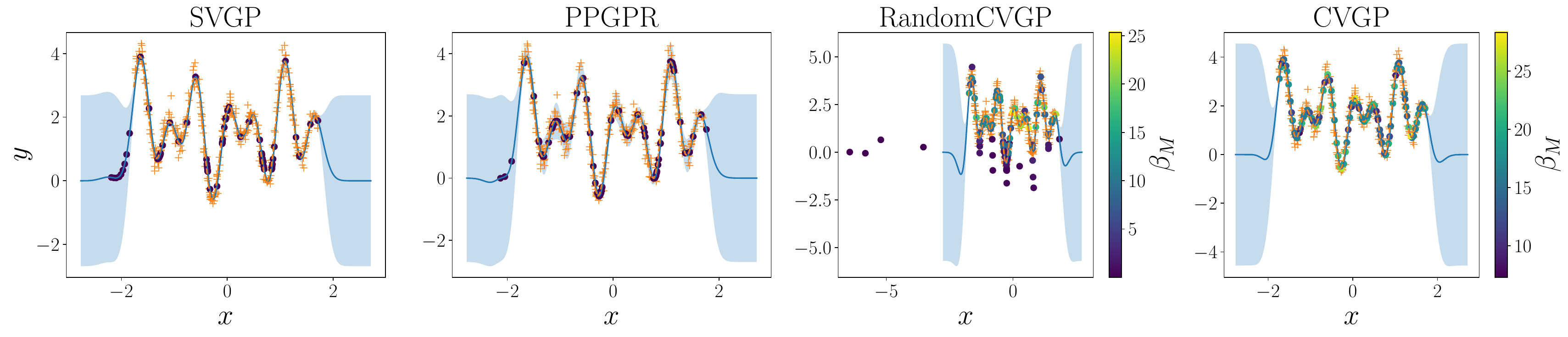}
        \vspace*{-5.2ex}
    \end{subfigure}
    \begin{subfigure}[c]{0.85\textwidth}
        \includegraphics[width=\textwidth]{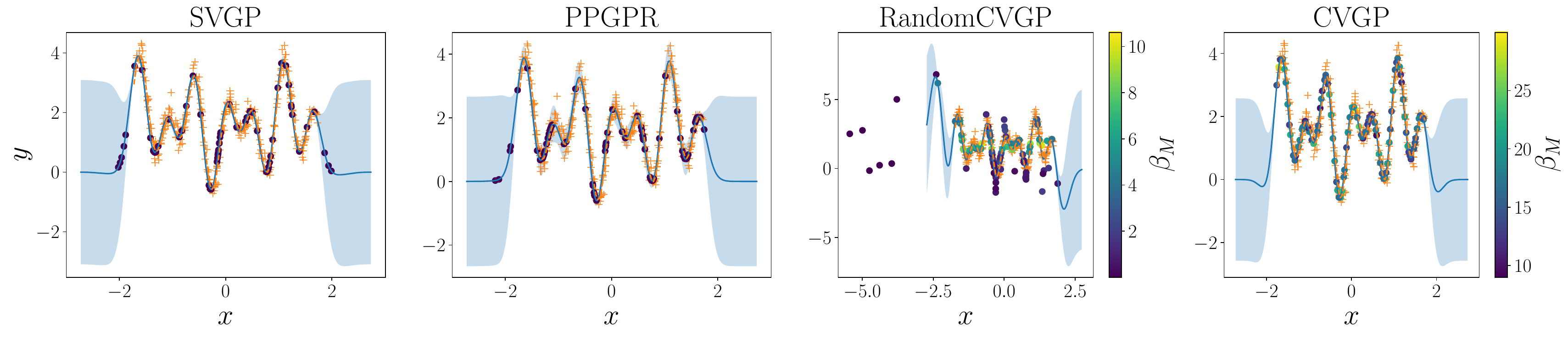}
        \vspace*{-5.2ex}
    \end{subfigure}
    \begin{subfigure}[c]{0.85\textwidth}
        \includegraphics[width=\textwidth]{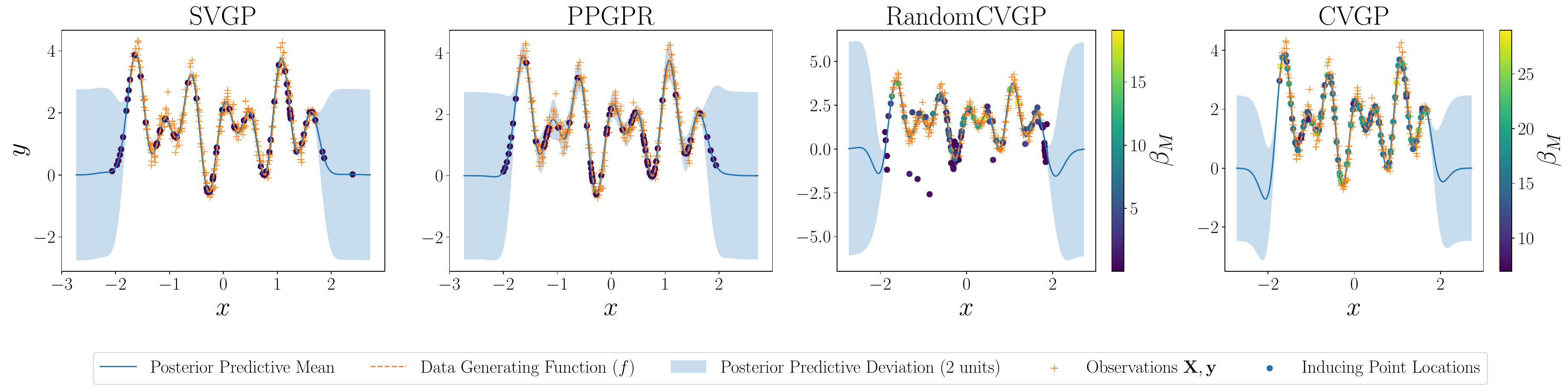}
    \end{subfigure}
    \caption{Posterior predictive distribution for the synthetic 2 dataset across different 5 folds, with 100 inducing points. Note that RandomCVGP is initialized with Gaussian white noise and faces a significantly more challenging task in fitting the data compared to SVGP, PPGPR, and CVGP. In fact, some of its learned inducing points/coresets fall off-the-grid and appear unrelated to the data. In these cases, the corresponding coreset weights are low (indicated in purple). Conversely, points that effectively capture the $y|x$ relationship are shown in yellow-green, while those that do not are depicted in purple—and are consequently disregarded during posterior inference. SVGP and PPGPR do not have this advantage. Hence, we use a single color for their coresets. This design endows CVGP with significant flexibility: if an inducing point proves unhelpful for predictions, its influence can be driven to 0 (with the help of its corresponding $\beta_m$). In contrast, both PPGPR and SVGP must select inducing points that consistently capture the data structure. }
    \label{fig:posterior_predictive_synthetic2}
\end{figure}

\begin{figure}[h!]
    \centering
    \begin{subfigure}[c]{0.85\textwidth}
        \includegraphics[width=\textwidth]{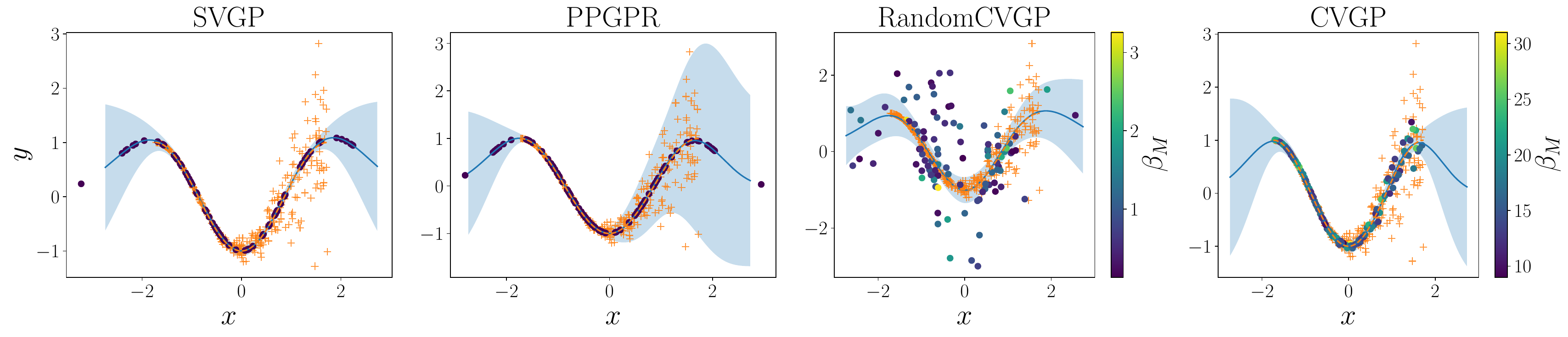}
        \vspace*{-5.2ex}
    \end{subfigure}
    \begin{subfigure}[c]{0.85\textwidth}
        \includegraphics[width=\textwidth]{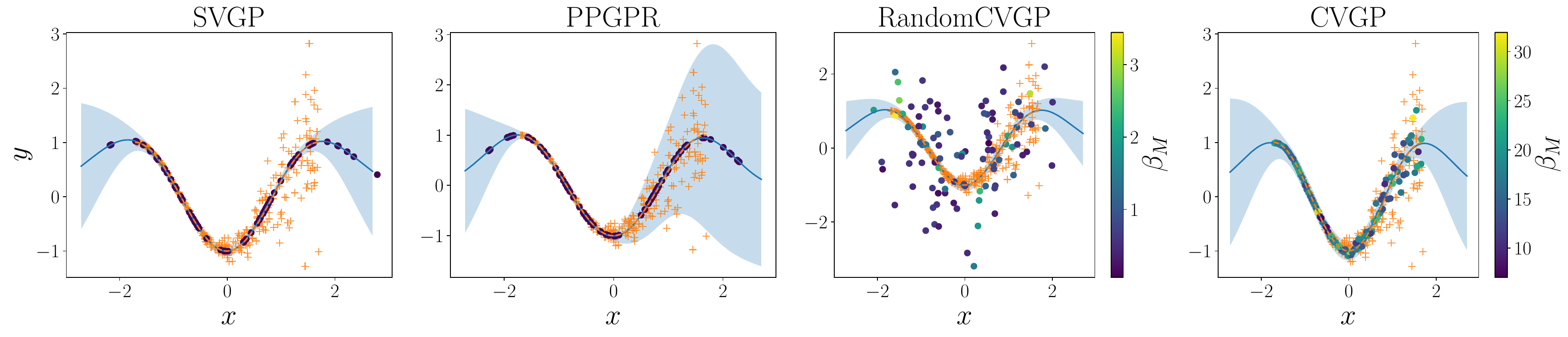}
        \vspace*{-5.2ex}
    \end{subfigure}
    \begin{subfigure}[c]{0.85\textwidth}
        \includegraphics[width=\textwidth]{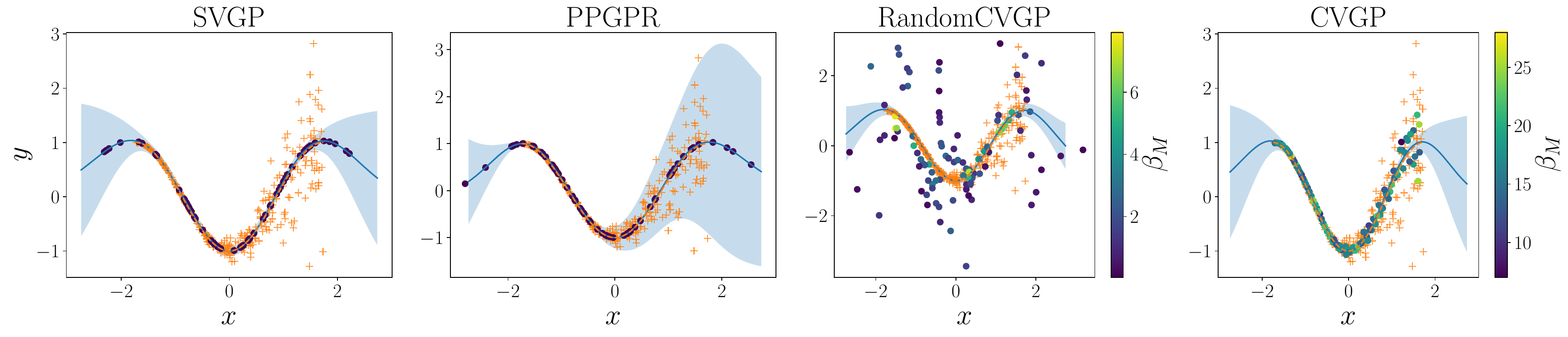}
        \vspace*{-5.2ex}
    \end{subfigure}
    \begin{subfigure}[c]{0.85\textwidth}
        \includegraphics[width=\textwidth]{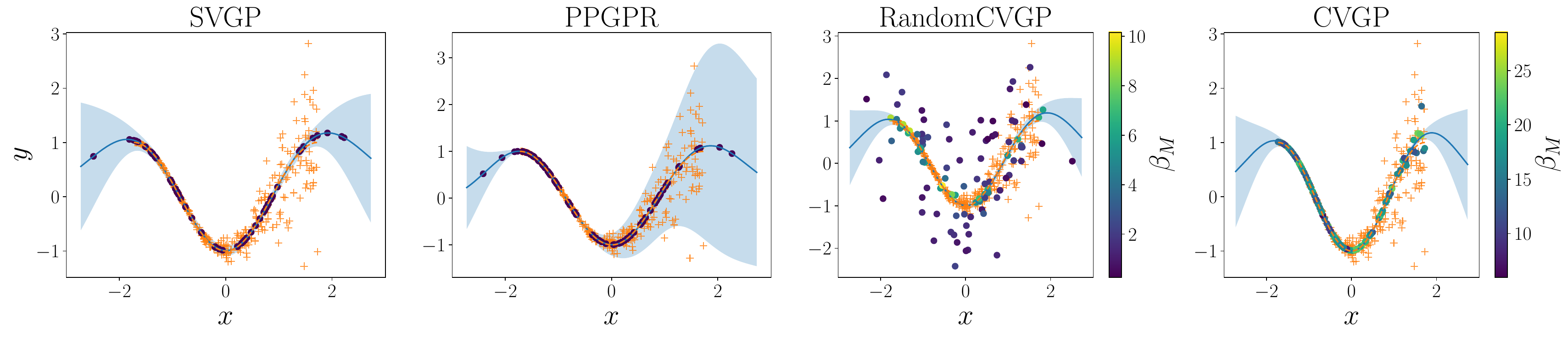}
        \vspace*{-5.2ex}
    \end{subfigure}
    \begin{subfigure}[c]{0.85\textwidth}
        \includegraphics[width=\textwidth]{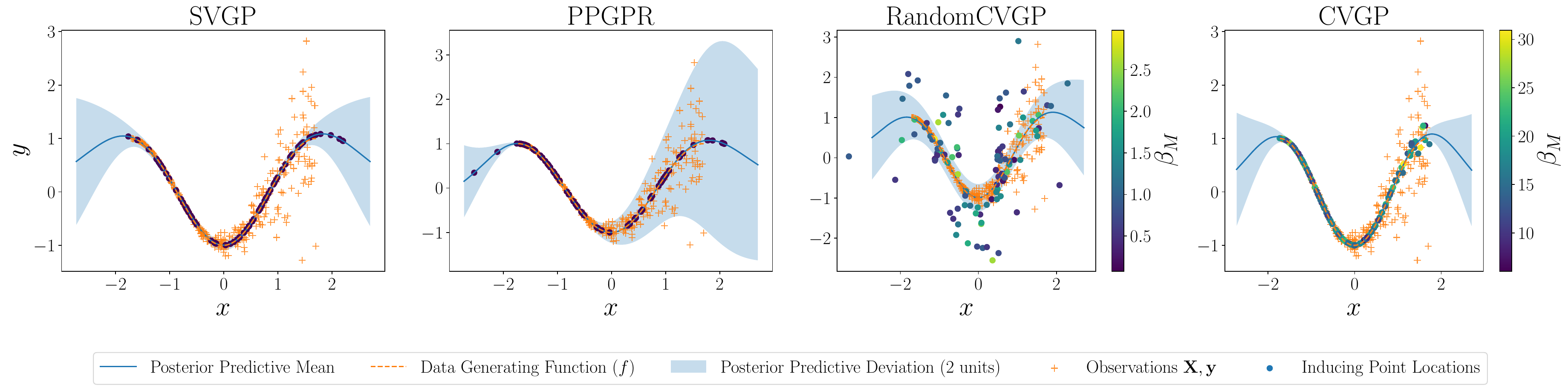}
    \end{subfigure}
    \caption{Posterior predictive distribution for the synthetic 3 dataset across different 5 folds, with 100 inducing points. Note that RandomCVGP is initialized with Gaussian white noise and faces a significantly more challenging task in fitting the data compared to SVGP, PPGPR, and CVGP. In fact, some of its learned inducing points/coresets fall off-the-grid and appear unrelated to the data. In these cases, the corresponding coreset weights are low (indicated in purple). Conversely, points that effectively capture the $y|x$ relationship are shown in yellow-green, while those that do not are depicted in purple—and are consequently disregarded during posterior inference. SVGP and PPGPR do not have this advantage. Hence, we use a single color for their coresets. We observe that PPGPR captures heteroscholastic uncertainty. \emph{Although this seems like a plausible property the noise in $y$ sometimes can purely be noise and could lead PPGPR to overfit as we discussed and demonstrated in Figures \ref{fig:exp_predictive_training_parkinsons} and \ref{fig:epoch_results}}.}
    \label{fig:posterior_predictive_synthetic3}
\end{figure}

Below, we showcase the predictive distributions of each trained $\gp$ model, for the synthetic 1D datasets,
\ie synthetic 1 in Figure~\ref{fig:posterior_predictive_synthetic1},
synthetic 2 in Figure~\ref{fig:posterior_predictive_synthetic2},
and synthetic 3 in Figure~\ref{fig:posterior_predictive_synthetic3}.

\clearpage
\newpage

\subsection{STUDY OF INDUCING POINTS ($\XbZ$)}
\label{asssec:app_exp_coresets}

\begin{figure}[h]
    \centering
    \begin{subfigure}[c]{0.85\textwidth}
        \includegraphics[width=\textwidth]{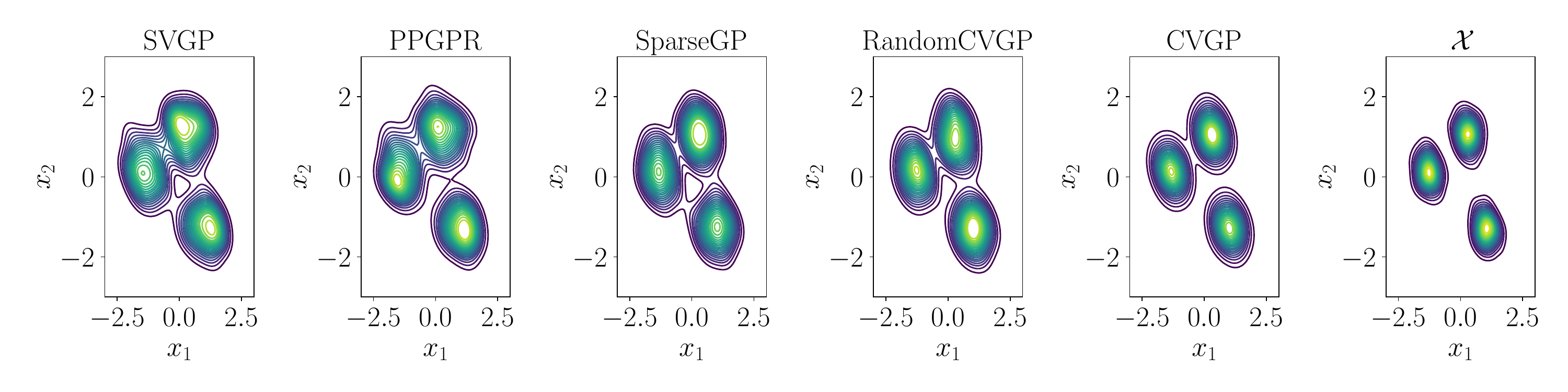}
        \vspace*{-5.5ex}
    \end{subfigure}
    \begin{subfigure}[c]{0.85\textwidth}
        \includegraphics[width=\textwidth]{figures/coreset_kde_synthetic4_fold_1.pdf}
        \vspace*{-5.5ex}
    \end{subfigure}
    \begin{subfigure}[c]{0.85\textwidth}
        \includegraphics[width=\textwidth]{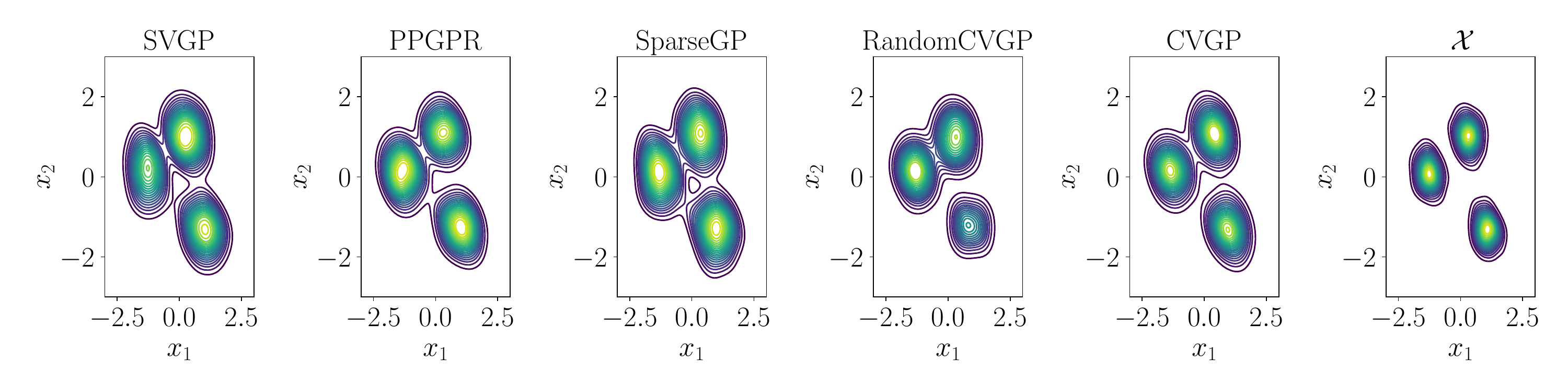}
        \vspace*{-5.5ex}
    \end{subfigure}
    \begin{subfigure}[c]{0.85\textwidth}
        \includegraphics[width=\textwidth]{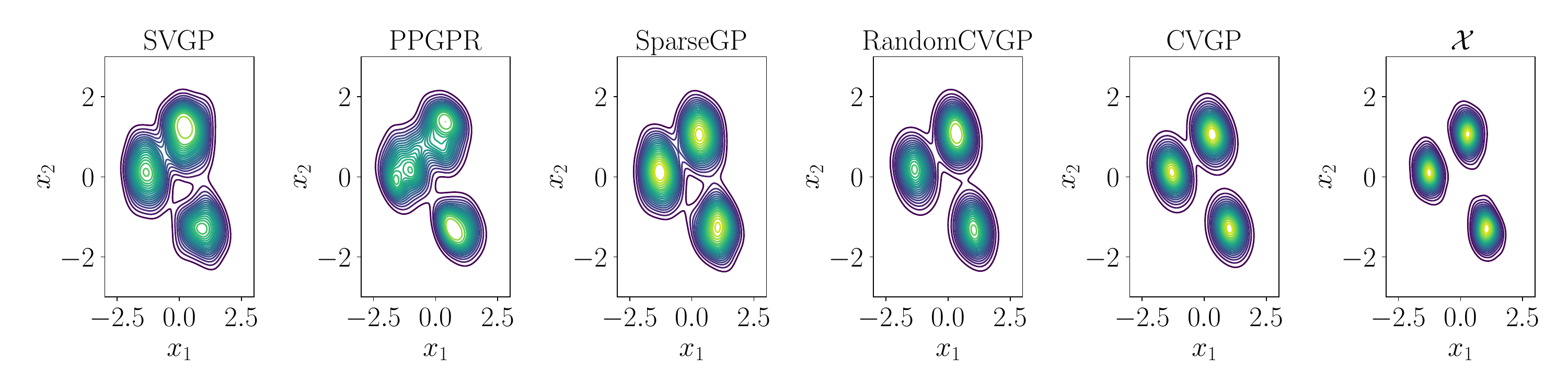}
        \vspace*{-5.5ex}
    \end{subfigure}
    \begin{subfigure}[c]{0.85\textwidth}
        \includegraphics[width=\textwidth]{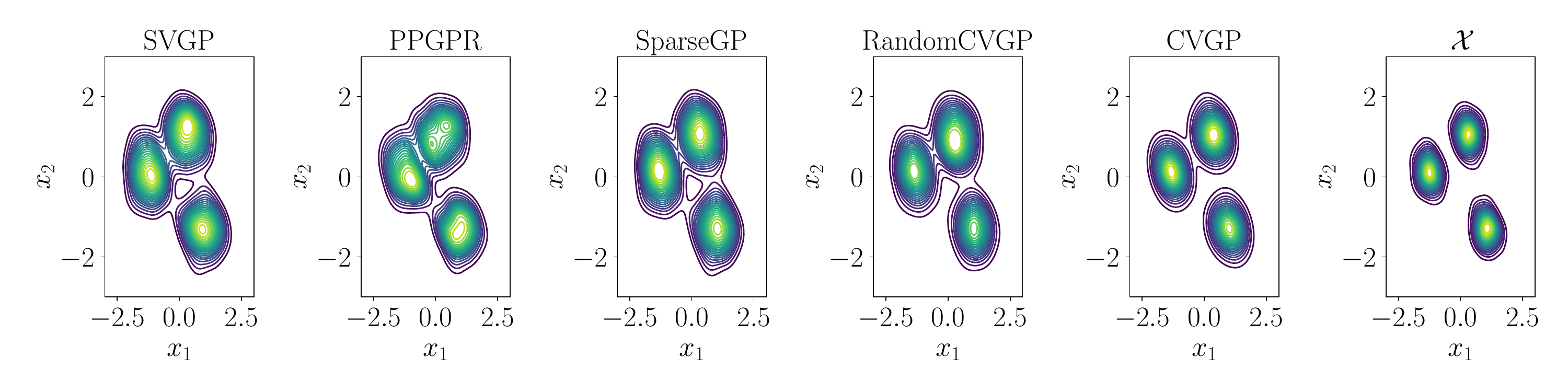}
        \vspace*{-5ex}
    \end{subfigure}
    \caption{Learned representations for synthetic 4 dataset over 5 different folds with 100 inducing points. CVGP learns meaningful representations over different folds. RandomCVGP is more noisy than CVGP as it is initialized with Gaussian white noise. }
    \label{fig:coreset_kde_synthetic4_folds}
\end{figure}

\begin{figure}[h]
    \centering
    \begin{subfigure}[c]{0.85\textwidth}
        \includegraphics[width=\textwidth]{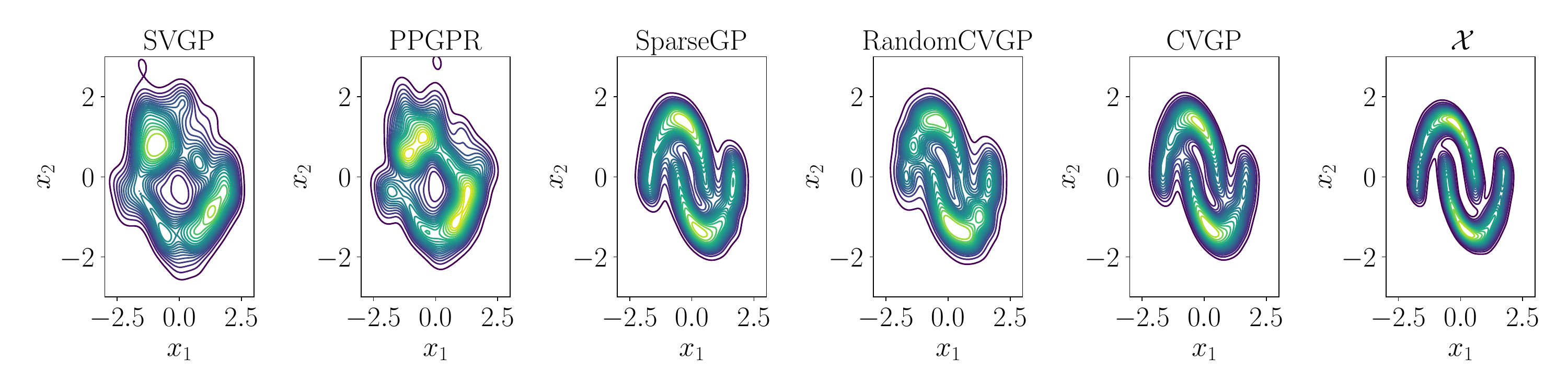}
        \vspace*{-5.5ex}
    \end{subfigure}
    \begin{subfigure}[c]{0.85\textwidth}
        \includegraphics[width=\textwidth]{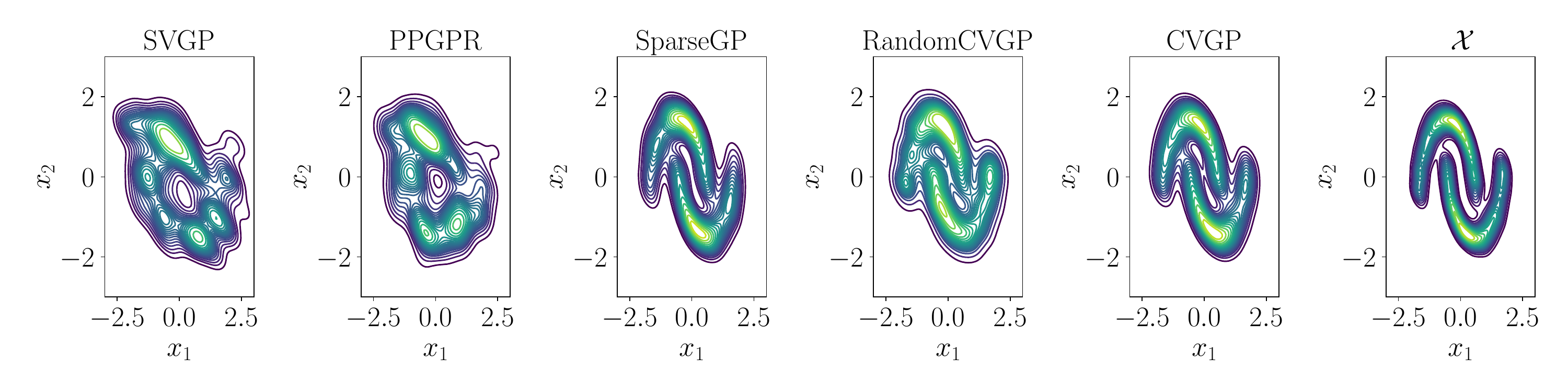}
        \vspace*{-5.5ex}
    \end{subfigure}
    \begin{subfigure}[c]{0.85\textwidth}
        \includegraphics[width=\textwidth]{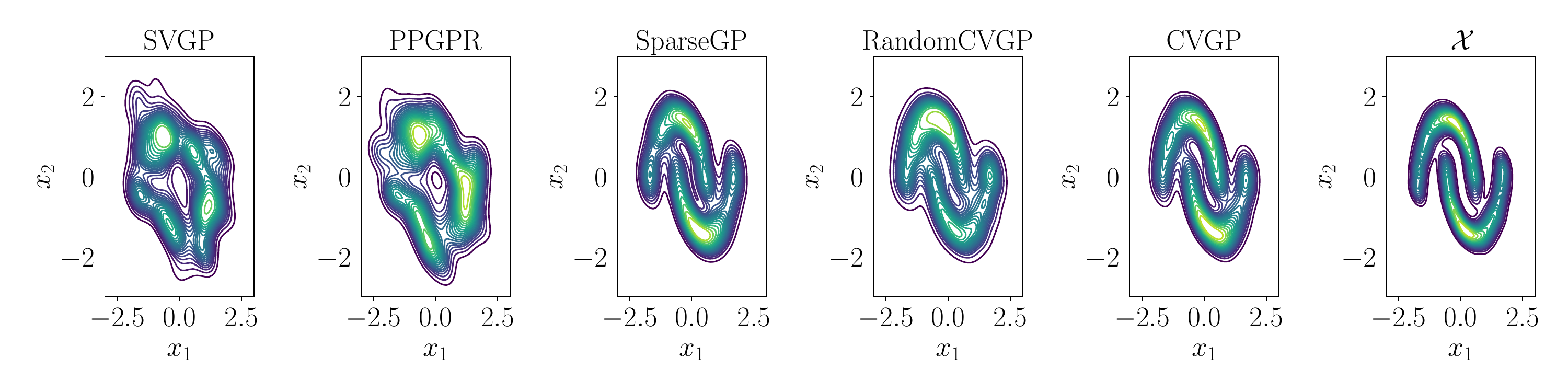}
        \vspace*{-5.5ex}
    \end{subfigure}
    \begin{subfigure}[c]{0.85\textwidth}
        \includegraphics[width=\textwidth]{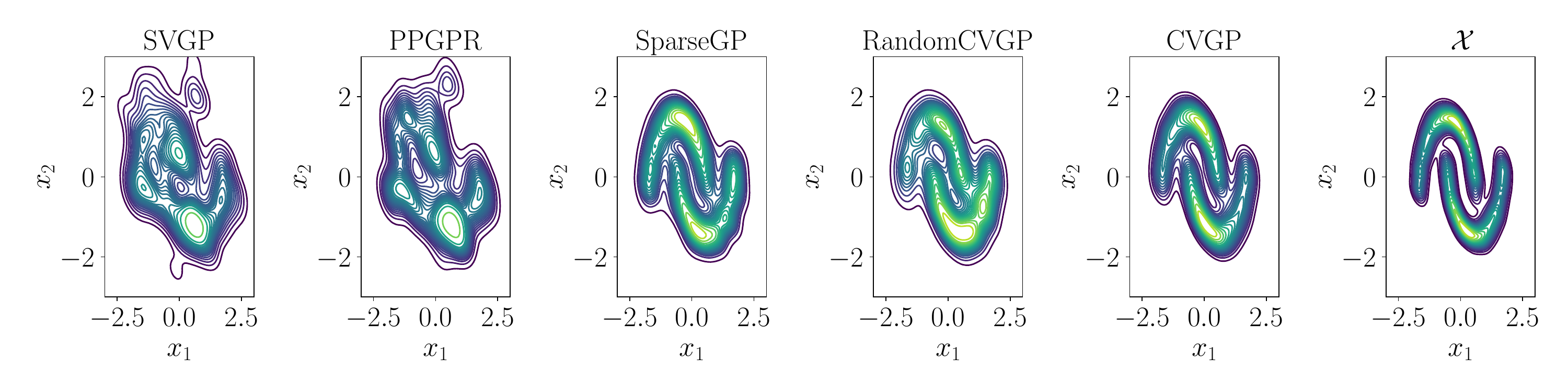}
        \vspace*{-5.5ex}
    \end{subfigure}
    \begin{subfigure}[c]{0.85\textwidth}
        \includegraphics[width=\textwidth]{figures/coreset_kde_synthetic5_fold_4.pdf}
        \vspace*{-5ex}
    \end{subfigure}
    \caption{Learned representations for synthetic 5 dataset over 5 different folds.
    CVGP learns meaningful representations over the different folds while other models, except for SparseGP,
    struggle to capture the empirical distribution.  RandomCVGP is more noisy than CVGP as it is initialized with Gaussian white noise.}
    \label{fig:coreset_kde_synthetic5_folds}
\end{figure}

We showcase the density of $\XbC$s learned by CVGP (weighted by $\betabC$),
and the $\XbZ$ points learned by other sparse $\gp$ methods
on the 2-dimensional synthetic \texttt{Blobs} (Figure~\ref{fig:coreset_kde_synthetic4_folds}) and \texttt{TwoMoons} (Figure~\ref{fig:coreset_kde_synthetic5_folds}) datasets,
across different folds of the training data.
Notice how CVGP consistently learns meaningful data representations over all folds.

\begin{figure*}[!ht]
    \centering
    \begin{subfigure}[c]{0.95\textwidth}
        \includegraphics[width=\textwidth]{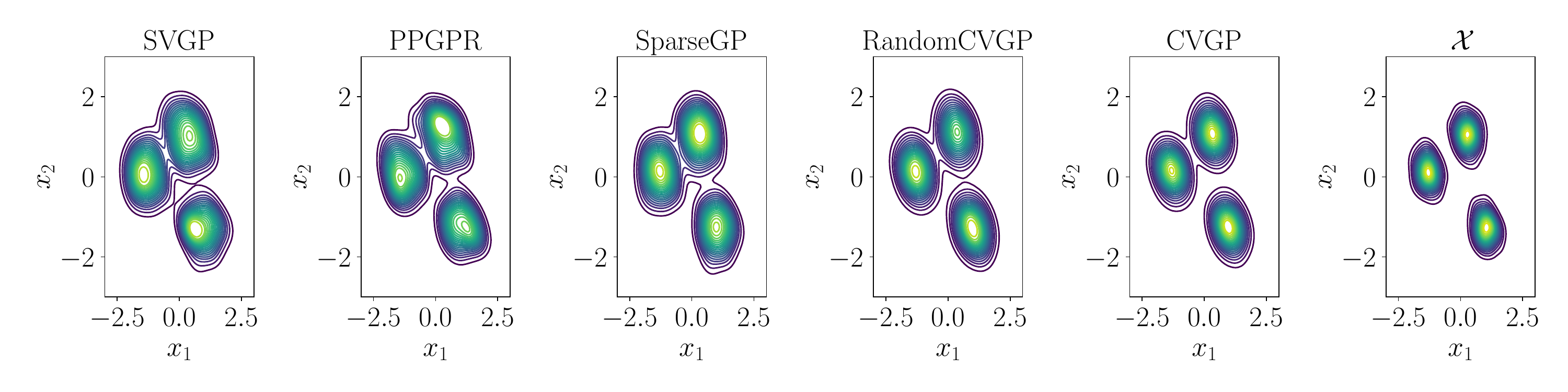}

                \caption{Synthetic 4 dataset where $y = f(\xb) + \epsilon$, and $\xb \sim \texttt{MakeBlobs}(.)$}
        \label{fig:coreset_syn4}
    \end{subfigure}

    \begin{subfigure}[c]{0.95\textwidth}
        \includegraphics[width=\textwidth]{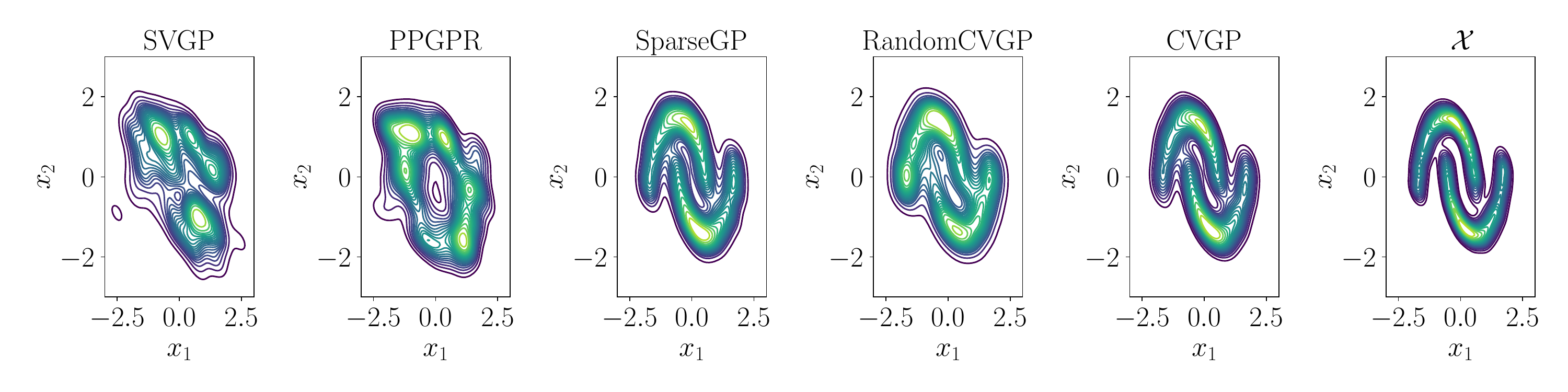}

        \caption{Synthetic 5 dataset where $y = f(\xb) + \epsilon$, and $\xb \sim \texttt{MakeMoons}(.)$}
        \label{fig:coreset_syn5}
    \end{subfigure}

        \caption{
        Kernel density estimation (KDE) plots for $\XbC$ learned by CVGP and $\XbZ$ for sparse baselines,
        on (a) synthetic 4 and (b) synthetic 5 datasets.
        For CVGP we use $\betabC$-weighted KDE plots, not possible for alternatives.
        All methods capture the clustered \texttt{Blobs} empirical distribution in Figure~\ref{fig:coreset_syn4},
        yet CVGP models the bi-modal nature of data more clearly.
        CVGP, RandomCVGP, and SparseGP adeptly capture the distinctive \texttt{TwoMoons} shape exhibited by the empirical data distribution in Figure~\ref{fig:coreset_syn5},
        in contrast to other stochastic sparse inference alternatives.
    }
    \label{fig:coreset}
\end{figure*}

\clearpage
\newpage

\subsection{LEARNED CORESET WEIGHT DISTRIBUTION FOR K-MEANS AND RANDOM INITIALIZATIONS}
\label{asssec:app_exp_coreset_weights}

We show below the histogram of CVGP's learned coreset's weights across all synthetic datasets.
Note that, all learned coresets have nonzero weights $\beta_m >0, \; \forall m$,
with very different histograms depending on the dataset:
for some datasets, some pseudo input-output points $\{\XbC, \ybC\}$ are considerably up-weighted. We observe that RandomCVGP drives the weight of unplausible inducing points to $0$.

\begin{figure}[h!]
\centering
\includegraphics[width=\linewidth]{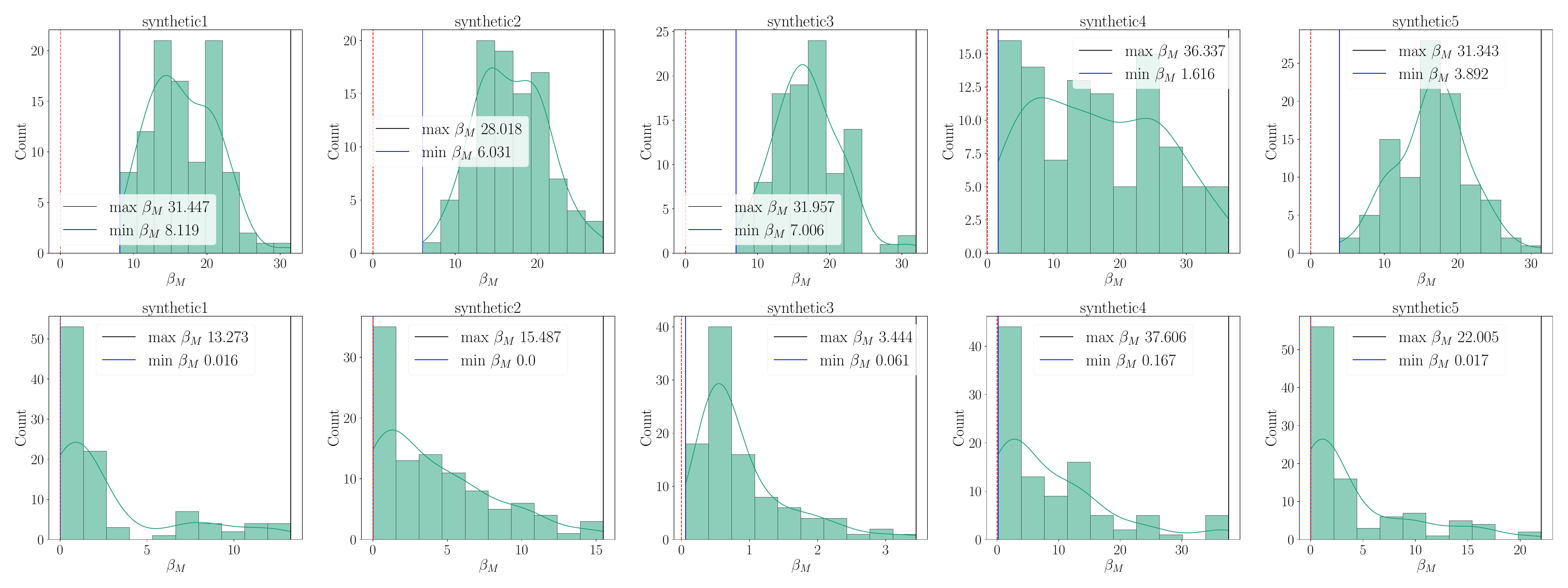}
\caption{Histogram of learned CVGP coreset weights $\betabC$. Top CVGP, bottom RandomCVGP (\ie initialization with white noise). We see that some of weights of RandomCVGP go to $0$ while almost all weight values of CVGP are non-zero (\ie no coreset tuple $\{\XbC, \ybC\}$ is discarded ($\beta_m >0, \; \forall m$)).
}
\label{fig:beta_hist}
\end{figure}

\vfill

\end{document}